\documentclass[runningheads]{llncs}
\usepackage{graphicx}
\usepackage{amssymb} 
\usepackage{amsmath}
\usepackage{color}
\usepackage[ruled,vlined]{algorithm2e}
\usepackage[tight]{subfigure}
\usepackage{multirow}
\usepackage{tabularx}
\usepackage{dsfont}
\usepackage{url}
\usepackage{float}
\usepackage{wrapfig}
\usepackage[toc,page]{appendix}

\usepackage{todonotes}

\def\argmin{\operatornamewithlimits{arg\,min}}

\definecolor{darkgreen}{rgb}{0,0.694,0.298}
\definecolor{purple}{rgb}{0.4,0.176,0.569}

\usepackage{xspace}
\makeatletter
\DeclareRobustCommand\onedot{\futurelet\@let@token\@onedot}
\def\@onedot{\ifx\@let@token.\else.\null\fi\xspace}
\def\eg{\emph{e.g}\onedot} 
\def\ie{\emph{i.e}\onedot}

\def\wrt{w.r.t\onedot} 
\def\etal{\emph{et al}\onedot}
\makeatother

\begin{document}

\title{RankGAN: A Maximum Margin Ranking GAN for Generating Faces} 
\titlerunning{RankGAN: A Maximum Margin Ranking GAN for Generating Faces} 


\author{Felix Juefei-Xu\thanks{These authors contribute equally and should be considered co-first authors.}\inst{1}\orcidID{0000-0002-0857-8611} \and
Rahul Dey$^{\star}$\inst{2}\orcidID{0000-0002-3594-5122} \and
Vishnu Naresh Boddeti\inst{2}\orcidID{0000-0002-8918-9385} \and
Marios Savvides\inst{1}}
%

\authorrunning{F. Juefei-Xu et al.} 


\institute{Carnegie Mellon University, Pittsburgh, PA 15213, USA \and
Michigan State University, East Lansing, MI 48824, USA}

\maketitle

\begin{abstract}
    We present a new stage-wise learning paradigm for training generative adversarial networks (GANs). The goal of our work is to progressively strengthen the discriminator and thus, the generators, with each subsequent stage without changing the network architecture. We call this proposed method the RankGAN. We first propose a margin-based loss for the GAN discriminator. We then extend it to a margin-based ranking loss to train the multiple stages of RankGAN. We focus on face images from the CelebA dataset in our work and show visual as well as quantitative improvements in face generation and completion tasks over other GAN approaches, including WGAN and LSGAN.
\end{abstract}

\section{Introduction}
Generative modeling approaches can learn from the tremendous amount of data around us to obtain a compact descriptions of the data distribution. Generative models can provide meaningful insight about the physical world that human beings can perceive, insight that can be valuable for machine learning systems. Take visual perception for instance, in order to generate new instances, the generative models must search for intrinsic patterns in the vast amount of visual data and distill its essence. Such systems in turn can be leveraged by machines to improve their ability to understand, describe, and model the visual world.
\begin{figure}
  \centering
  \includegraphics[width=\linewidth]{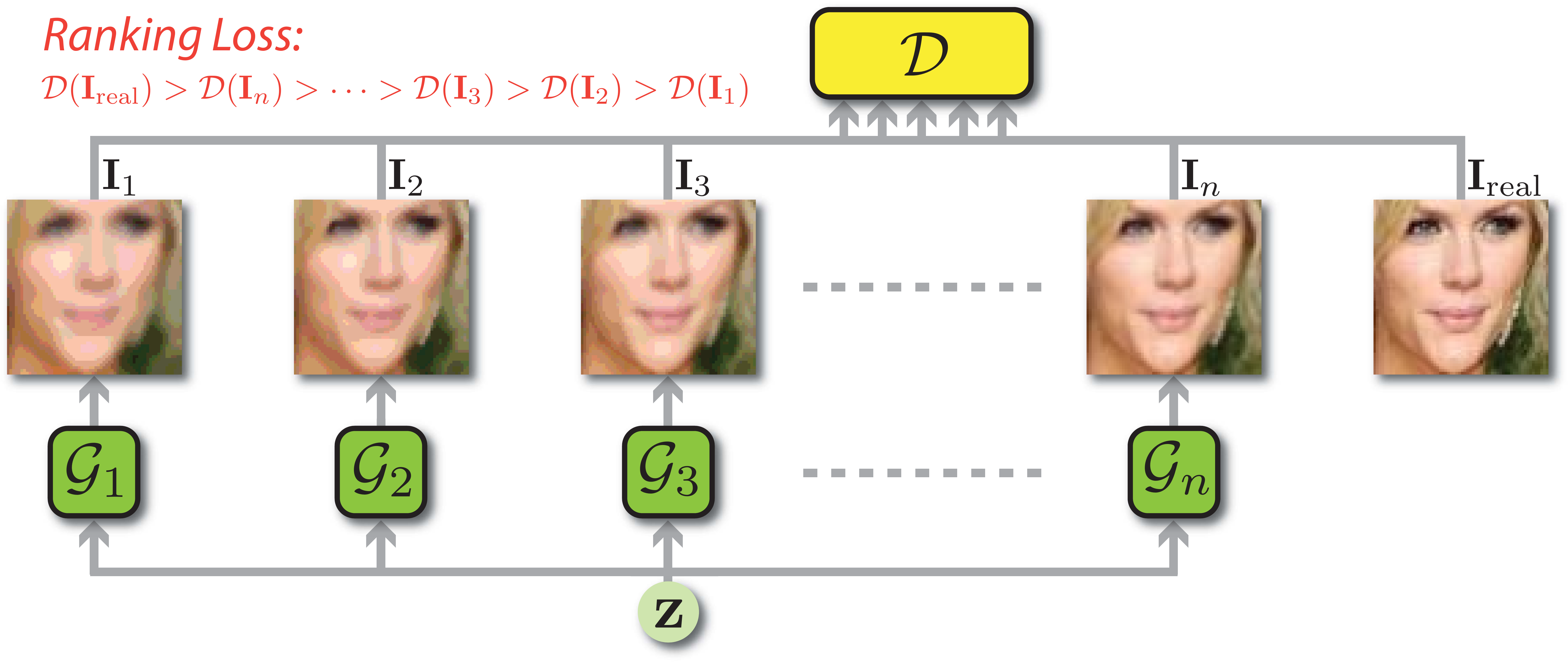}
  \caption{The RankGAN framework consists of a discriminator that ranks the quality of the generated images from several stages of generators. The ranker guides the generators to learn the subtle nuances in the training data and progressively improve with each stage.}
  \label{fig:basic_idea}
\end{figure}

Recently, three classes of algorithms have emerged as successful generative approaches to model the visual data in an unsupervised manner. \emph{Variational autoencoders} (VAEs) \cite{vae} formalize the generative problem as a maximum log-likelihood based learning objective in the framework of probabilistic graphical models with latent variables. The learned latent space allows for efficient reconstruction of new instances. The VAEs are straightforward to train but at the cost of introducing potentially restrictive assumptions about the approximate posterior distribution. Also, their generated samples tend to be slightly blurry.
\emph{Autoregressive} models such as PixelRNN \cite{pixelrnn} and PixelCNN \cite{pixelcnn} get rid of the latent variables and instead directly model the conditional distribution of every individual pixel given the previous starting pixels. PixelRNN/CNN have a stable training process via softmax loss and currently give the best log likelihoods on the generated data, indicating high plausibility. However, they lack a latent code and are relatively inefficient during sampling.

\emph{Generative adversarial networks} bypass maximum-likelihood learning by training a generator using adversarial feedback from a discriminator. Using a latent code, the generator tries to generate realistic-looking data in order to fool the discriminator, while the discriminator learns to classify them apart from the real training instances. This two-player minimax game is played until the Nash equilibrium where the discriminator is no longer able to distinguish real data from the fake ones. The GAN loss is based on a measure of distance between the two distributions as observed by the discriminator. GANs are known to generate highest quality of visual data by far in terms of sharpness and semantics.

Because of the nature of GAN training, the strength (or quality) of the generator, which is the desired end-product, depends directly on the strength of the discriminator. The stronger the discriminator is, the better the generator has to become in generating realistic looking images, and vice-versa. Although a lot of GAN variants have been proposed that try to achieve this by exploring different divergence measures between the real and fake distributions, there has not been much work dedicated to self-improvement of GAN, \ie, progressively improving the GAN based on self-play with the previous versions of itself. One way to achieve this is by making the discriminator not just compare the real and fake samples, but also rank fake samples from various stages of the GAN, thus forcing it to get better in attending to the finer details of images. In this work, we propose a progressive training paradigm to train a GAN based on a maximum margin ranking criterion that improves GANs at later stages keeping the network capacity same. Thus, our proposed approach is orthogonal to other progressive paradigms such as \cite{progressive} which increase the network capacity to improve the GAN and the resolution of generated images in a stage-wise manner. We call our proposed method RankGAN.

Our contributions include (1) a margin-based loss function for training the discriminator in a GAN; (2) a self-improving training paradigm where GANs at later stages improve upon their earlier versions using a maximum-margin ranking loss (see Figure~\ref{fig:basic_idea}); and (3) a new way of measuring GAN quality based on image completion tasks.

\subsection{Related Work}\label{related_work}

Since the introduction of Generative Adversarial Networks (GANs) \cite{gan}, numerous variants of GAN have been proposed to improve upon it. The original GAN formulation suffers from practical problems such as vanishing gradients, mode collapse and training instability. To strive for a more stable GAN training, Zhao \etal{} proposed an energy-based GAN (EBGAN) \cite{ebgan} which views the discriminator as an energy function that assigns low energy to the regions near the data manifold and higher energy to other regions. The authors have shown one instantiation of EBGAN using an autoencoder architecture, with the energy being the reconstruction error. The boundary-seeking GAN (BGAN) \cite{bgan} extended GANs for discrete data while improving training stability for continuous data. BGAN aims at generating samples that lie on the decision boundary of a current discriminator in training at each update. The hope is that a generator can be trained in this way to match a target distribution at the limit of a perfect discriminator.
Nowozin \etal{} \cite{fgan} showed that the generative-adversarial approach in GAN is a special case of an existing more general variational divergence estimation approach, and that any $f$-divergence can be used for training generative neural samplers. On these lines, least squares GAN (LSGAN) \cite{lsgan} adopts a least squares loss function for the discriminator, which is equivalent to minimizing the Pearson $\chi^2$ divergence between the real and fake distributions, thus providing smoother gradients to the generator.

Perhaps the most seminal GAN-related work since the inception of the original GAN \cite{gan} idea is the Wasserstein GAN (WGAN) \cite{wgan}. Efforts have been made to fully understand the training dynamics of GANs through theoretical analysis in \cite{wgan-pre} and \cite{wgan}, which leads to the creation of WGAN. By incorporating the smoother Wasserstein distance metric as the objective, as opposed to the KL or JS divergences, WGAN is able to overcome the problems of vanishing gradient and mode collapse. WGAN also made it possible to first train the discriminator till optimality and then gradually improve the generator making the training and balancing between the generator and the discriminator much easier. Moreover, the new loss function also correlates well with the visual quality of generated images, thus providing a good indicator for training progression.


On the other hand, numerous efforts have been made to improve the training and performance of GANs architecturally. Radford \etal{} proposed the DCGAN \cite{dcgan} architecture that utilized strided convolution and transposed-convolution to improve the training stability and performance of GANs. The Laplacian GAN (LAPGAN) \cite{lapgan} is a sequential variant of the GAN model that generates images in a coarse-to-fine manner by generating and upsampling in multiple steps. Built upon the idea of sequential generation of images, the recurrent adversarial networks \cite{gam} has been proposed to let the recurrent network learn the optimal generation procedure by itself, as opposed to imposing a coarse-to-fine structure on the procedure. The stacked GAN \cite{sgan} consists of a top-down stack of GANs, each trained to generate plausible lower-level representations, conditioned on higher-level representations. Discriminators are attached to each feature hierarchy to provide intermediate supervision. Each GAN of the stack is first trained independently, and then the stack is trained end-to-end. The generative multi-adversarial networks (GMAN) \cite{gman} extends the GANs to multiple discriminators that collectively scrutinize a fixed generator, thus forcing the generator to generate high fidelity samples. Layered recursive generative adversarial networks (LR-GAN) \cite{lrgan} generates images in a recursive fashion. First a background is generated, conditioned on which, the foreground is generated, along with a mask and an affine transformation that together define how the background and foreground should be composed to obtain a complete image.

The introspective adversarial networks (IAN) \cite{ian} proposes to hybridize the VAE and the GAN by leveraging the power of the adversarial objective while maintaining the efficient inference mechanism of the VAE.

Among the latest progress in GANs, Karras \etal~\cite{progressive} has the most impressive image generation results in terms of resolution and image quality. The key idea is to grow both the generator and discriminator progressively: starting from a low resolution, new layers that model increasingly fine details are added as the training progresses. This both speeds the training up and greatly stabilizes it, allowing us to produce images of unprecedented quality. On the contrary, we focus on improving the performance of GANs without increasing model capacity, making our work orthogonal to \cite{progressive}. In the following sections, we will first discuss the background and motivation behind our work, followed by details of the proposed approach.

\section{Background}
We first provide a brief background of a few variants of GAN to motivate the maximum margin ranking based GAN proposed in this paper.
\subsection{GAN and WGAN}
The GAN framework \cite{gan} consists of two components, a Generator $\mathcal{G}_\theta(\mathbf{z}):\mathbf{z}\rightarrow \mathbf{x}$ that maps a latent vector $\mathbf{z}$ drawn from a known prior $p_{\mathbf{z}}(\mathbf{z})$ to the data space and a Discriminator $\mathcal{D}_\omega(\mathbf{x}): \mathbf{x}\rightarrow [0,1]$ that maps a data sample (real or generated) to a likelihood value in $[0, 1]$. The generator $\mathcal{G}$ and the discriminator $\mathcal{D}$ play adversary to each other in a two-player minimax game while optimizing the following GAN objective:
\begin{align}
\min \limits_{\mathcal{G}} \max \limits_{\mathcal{D}} V(\mathcal{G},\mathcal{D}) = \mathbb{E}_{\mathbf{x} \sim p_\mathrm{data}(\mathbf{\mathbf{x}})} [\log(\mathcal{D}(\mathbf{x})) ] + \mathbb{E}_{\mathbf{z}\sim p_\mathbf{z}(\mathbf{z})} [\log(1-\mathcal{D}(\mathcal{G}(\mathbf{z}))]
\end{align}
where $\mathbf{x}$ is a sample from the data distribution $p_{\mathrm{data}}$. This objective function is designed to learn a generator $\mathcal{G}$ that minimizes the Jensen-Shannon divergence between the real and generated data distributions.

Many of the variants of GAN described in Section~\ref{related_work} differ in the objective function that is optimized to minimize the divergence between the real and generated data distributions. Wasserstein GAN \cite{wgan,wgan-pre} has been proposed with the goal of addressing the problems of vanishing gradients and mode collapse in the original GAN. Instead of minimizing the cross-entropy loss, the discriminator in WGAN is optimized to minimize the Wasserstein-1 (Earth Movers') distance $W(\mathds{P}_r,\mathds{P}_g)$ between the real and generated distributions.
\begin{align}
W(\mathds{P}_r,\mathds{P}_g) = \inf_{\gamma\in\Gamma(\mathds{P}_r,\mathds{P}_g)}\mathds{E}_{(x,y)\sim \gamma}\big[\|x-y\|\big]
\label{was_inf}
\end{align}
where $\Gamma(\mathds{P}_r,\mathds{P}_g)$ is the set of all joint distributions $\gamma(x,y)$ whose marginals are $\mathds{P}_r$ and $\mathds{P}_g$ respectively. Given the intractability of finding the infimum in Eqn.~\eqref{was_inf}, WGAN optimizes the dual objective given by the Kantorovich-Rubinstein duality \cite{wgan-book} instead, which also constraints the discriminator to be a 1-Lipshichtz function.


\subsection{Limitations with GANs and its Variants}

\begin{figure}
  \centering
  \includegraphics[width=\textwidth]{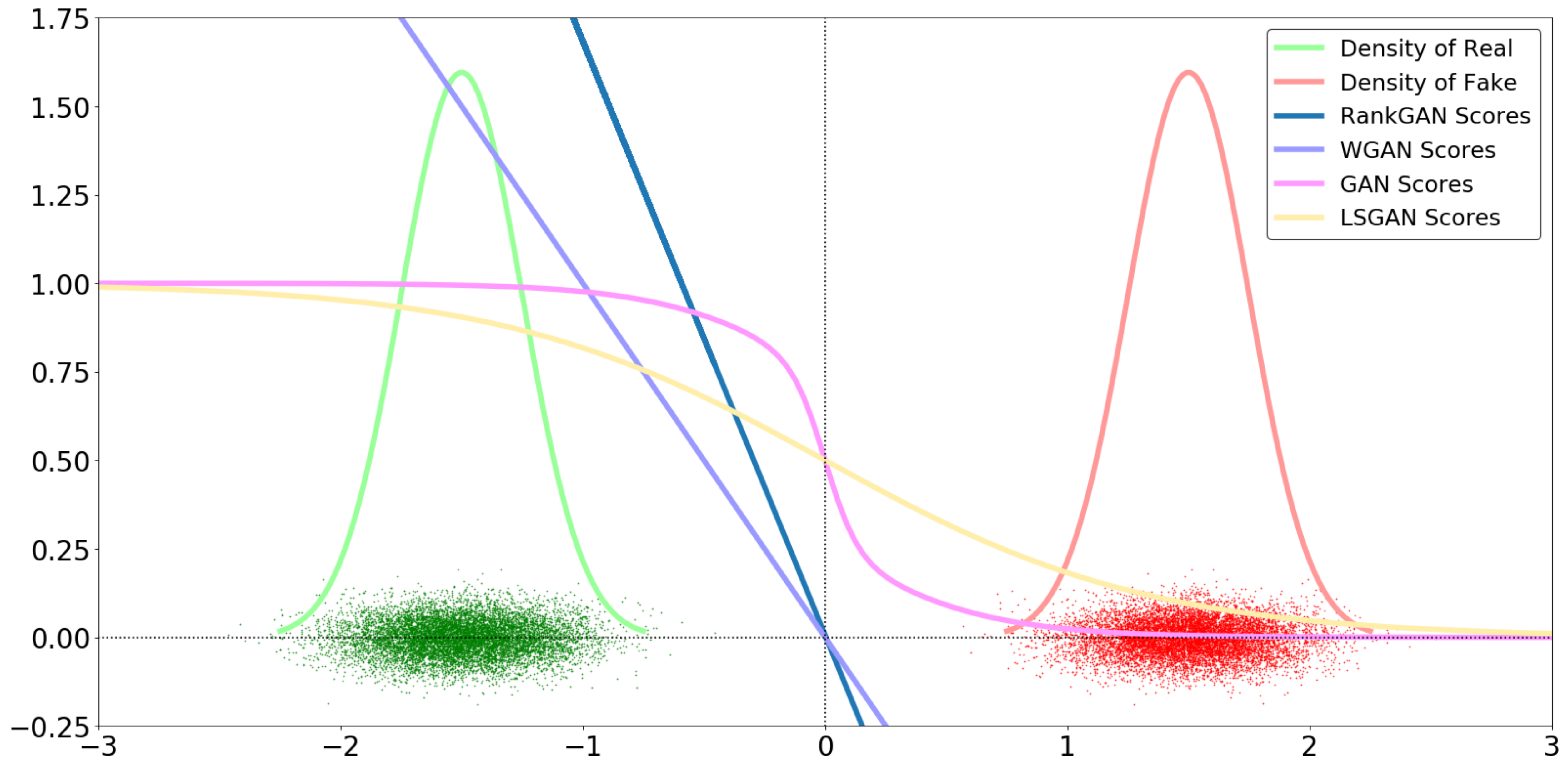}
  \caption{Scores of the optimal discriminator for GAN, WGAN, LSGAN and RankGAN when learning to differentiate between two normal distributions. The GAN scores are saturated and hence results in vanishing gradients. The WGAN and RankGAN models do not suffer from this problem. See text for more details.}
  \label{fig:gan_scores}
\end{figure}

An essential part of the adversarial game being played in a GAN is the discriminator, which is modeled as a two-class classifier. Thus, intuitively, the stronger the discriminator, the stronger (better) should be the generator. In the original GAN, stronger discriminator led to problems like vanishing gradients \cite{wgan-pre}.
Variants like WGAN and LSGAN attempt to solve this problem by proposing new loss functions that represent different divergence measures. We illustrate this effect in Figure~\ref{fig:gan_scores}. The scores of the standard GAN model saturate and thus provide no useful gradients to the discriminator. The WGAN model has a constant gradient of one while RankGAN model (described in the next section) has a gradient that depends on the slope of the linear decision boundary. Therefore, from a classification loss perspective, RankGAN generalizes the loss of the WGAN critic.
In practice, these variants don't easily reach convergence, partially because of limited network capacity and finite sample size of datasets. Loss functions for optimizing the discriminator are typically averaged over the entire dataset or a mini-batch of samples. As a result, the discriminator often keeps on increasing the margin between well-separated real and fake samples while struggling to classify the more difficult cases.
Furthermore, we argue that a margin-based loss, as in the case of support vector machines, enables the discriminator to focus on the difficult cases once the easier ones have been well classified, making it a more effective classifier. Going one step further, by ranking several versions of the generator, the discriminator would more effectively learn the subtle nuances in the training data. The supervision from such a strong discriminator would progressively improve the generators. This intuition forms the basic motivation behind our proposed approach.

\section{Proposed Method: RankGAN}\label{sec-gogan:gogan}
\begin{figure}
  \centering
  \includegraphics[width=\linewidth]{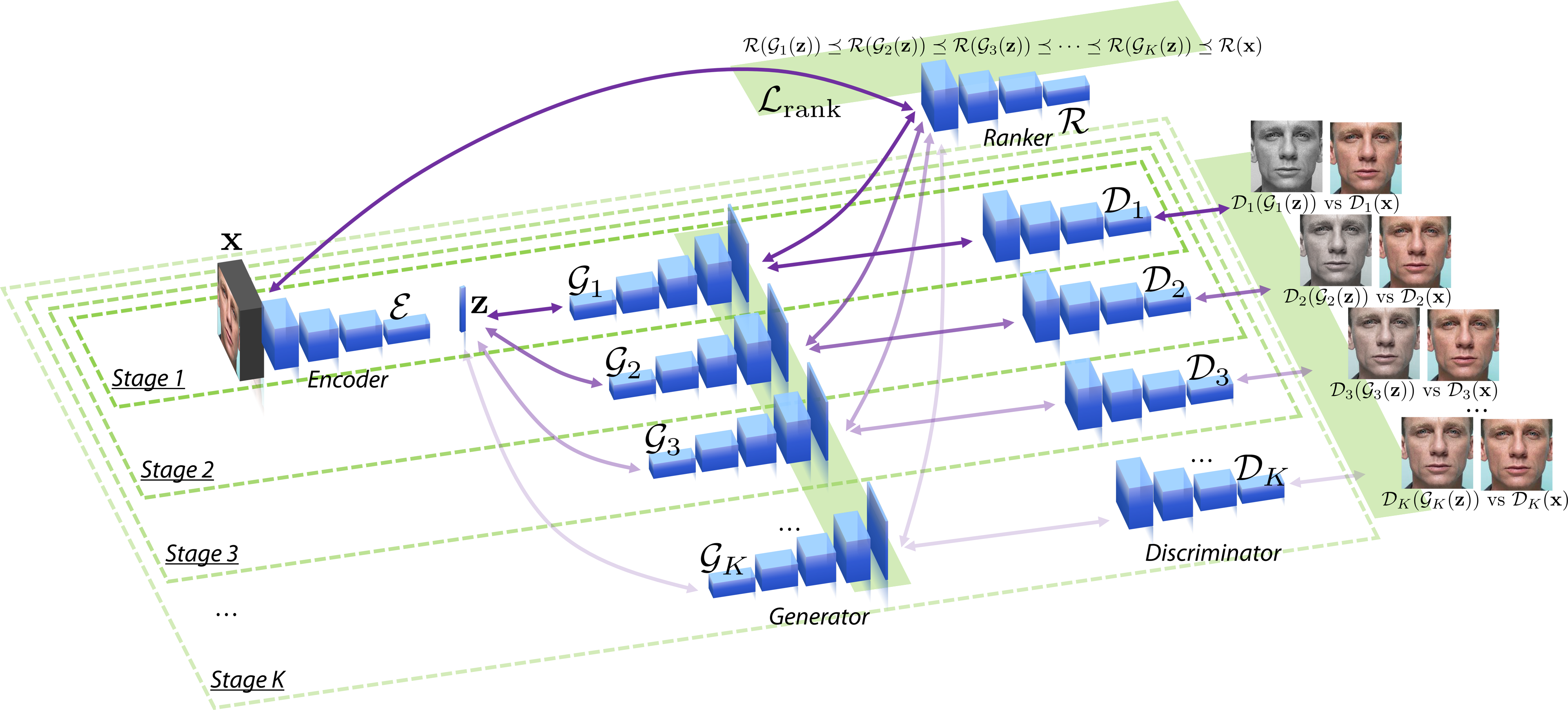}  
  \caption{Overall flowchart of the proposed RankGAN method. Our model consists of (1) An encoder that maps an image to a latent representation. (2) A series of generators that are learned in a stage-wise manner. (3) A series of discriminators that are learned to differentiate between the real and the generated data. (4) A ranker that ranks the real face image and the corresponding generated face images at each stage. In practice the discriminator and the ranker are combined into a single model.}
  \label{fig:flowchart}
\end{figure}

In this section, we describe our proposed GAN training framework - RankGAN. This model is designed to address some of the limitations of traditional GAN variants. RankGAN is a stage-wise GAN training paradigm which aims at improving the GAN convergence at each stage by ranking one version of GAN against previous versions without changing the network architecture (see Figure~\ref{fig:flowchart}). The two basic aspects of our proposed approach are the following:
\begin{itemize}
\item  We first adopt a margin based loss for the discriminator of the GAN, as opposed to the cross-entropy loss of the original GAN and the WGAN loss. We refer to this model as MarginGAN. 
\item  We extend the margin-based loss into a margin-based ranking loss. This enables the discriminator to rank multiple stages of generators by comparing the scores of the generated samples to those of the real samples (see Figure~\ref{fig:gogan_training} for an illustration).
By applying certain constraints on the discriminator, which we will describe later, we can use this mechanism to steadily improve the discriminator at each stage, thereby improving the quality of generated samples.
\end{itemize}
The complete RankGAN training flow is shown in Algorithm~\ref{algo:gogan}. We now describe the various novelties in our approach.

\subsection{Margin Loss}
The intuition behind the MarginGAN loss is as follows. WGAN loss treats a gap of 10 or 1 equally and it tries to increase the gap even further. The MarginGAN loss will focus on increasing separation of examples with gap 1 and leave the samples with separation 10, which ensures a better discriminator, hence a better generator. The $\epsilon$-margin loss is given by:
\begin{align}
\mathcal{L}_{\mathrm{margin}} = [\mathcal{D}_w( \mathcal{G}_\theta(\mathbf{z})) + \epsilon - \mathcal{D}_w(\mathbf{x})]_+
\label{eq:l_margin}
\end{align}
where $[x]_+ = \max(0,x)$ is the hinge loss. The margin loss becomes equal to the WGAN loss when the margin $\epsilon \rightarrow \infty$, hence the generalization.

\subsection{Ranking Loss}
\begin{figure}
  \centering
  \includegraphics[width=0.7\linewidth]{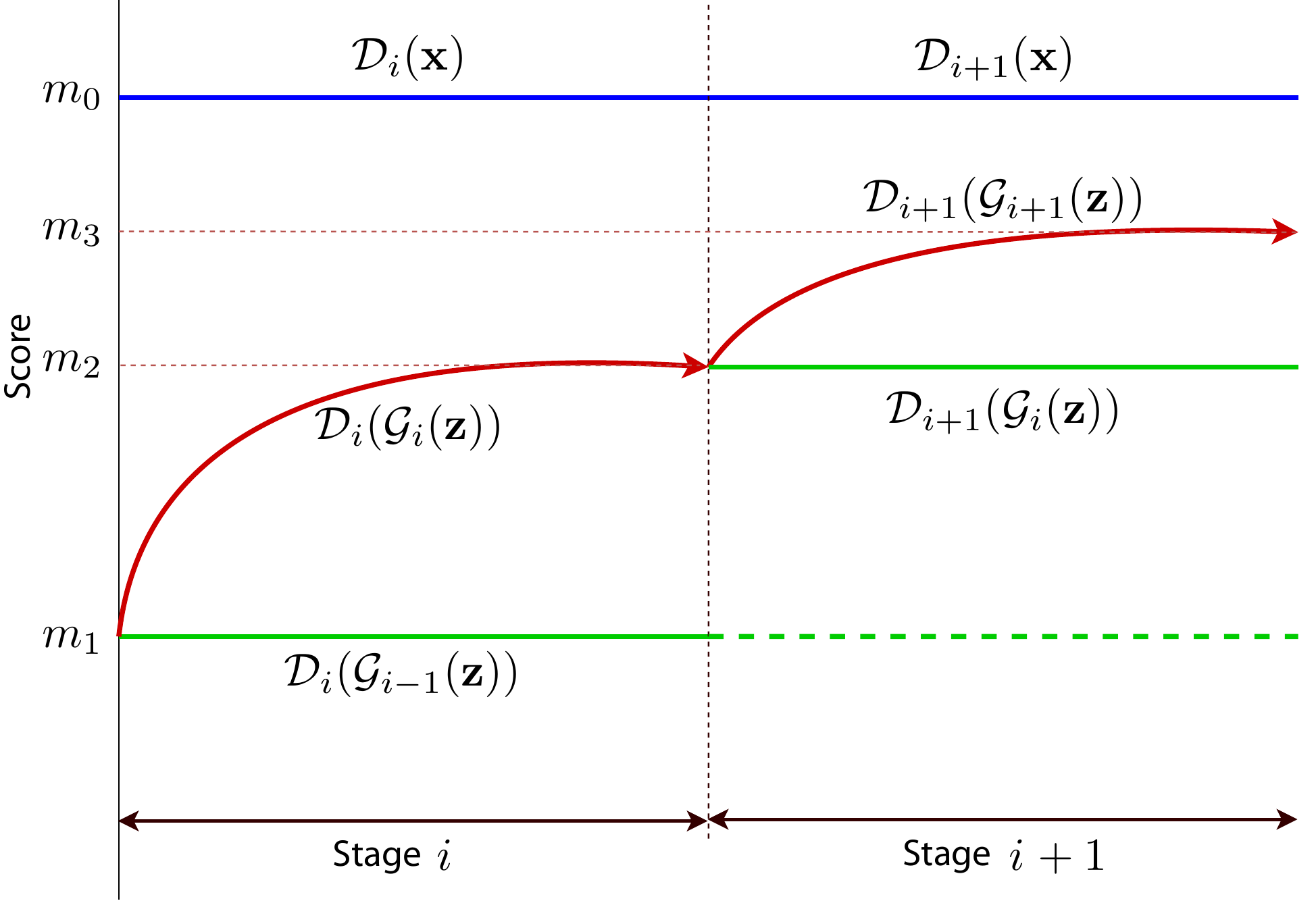}
  \caption{RankGAN stage-wise training progression following $\mathcal{D}_i(\mathbf{x}) > \mathcal{D}_i(\mathcal{G}_i(\mathbf{z})) > \mathcal{D}_i(\mathcal{G}_{i-1}(\mathbf{z}))$. At stage $i$, $\mathcal{D}_i(\mathbf{x})$ and $\mathcal{D}_i(\mathcal{G}_{i-1}(\mathbf{z}))$ are clamped at the initial margins $m_0$ and $m_1$, respectively while $\mathcal{D}_i(\mathcal{G}_i(\mathbf{z}))$ slowly increases from $m_1$ to $m_2$ (point of Nash equilibrium) at the end of stage $i$. The same is repeated at stage $i+1$, where $\mathcal{D}_{i+1}(\mathbf{x})$ and $\mathcal{D}_{i+1}(\mathcal{G}_i(\mathbf{z}))$ are clamped at margins $m_0$ and $m_2$ respectively while $\mathcal{D}_{i+1}(\mathcal{G}_{i+1}(\mathbf{z}))$ slowly increases from $m_2$ to $m_3$ till convergence.}\label{fig:gogan_training}
\end{figure}

The ranking loss uses margin loss to train the generator of our GAN by ranking it against previous version of itself. For stage $i$ discriminator $\mathcal{D}_i$ and generator $\mathcal{G}_i$, the ranking loss is given by:
\begin{equation}
    \begin{aligned}
    \label{eqn:ranking_loss}
    \mathbf{\mathcal{L}_{\mathrm{disc\_rank}}} &= \left[ \mathcal{D}_i\left( \mathcal{G}_i (\mathbf{z}) \right) - \mathcal{D}_i\left( \mathcal{G}_{i-1} (\mathbf{z}) \right) \right]_+ \\
    \mathbf{\mathcal{L}_{\mathrm{gen\_rank}}} &= \left[ \mathbf{\mathcal{D}}_i\left( \mathbf{x} \right) - \mathbf{\mathcal{D}}_i\left( \mathbf{\mathcal{G}}_{i} (\mathbf{z}) \right) \right]_+
\end{aligned}
\end{equation}
The ranking losses for the discriminator and the generator are thus zero margin loss functions ($\epsilon \rightarrow 0$) where the discriminator $\mathcal{D}_i$ is trying to have a zero margin between $\mathcal{D}_i(\mathcal{G}_{i}(\mathbf{z}))$ and $\mathcal{D}_i(\mathcal{G}_{i-1}(\mathbf{z}))$, while the generator is trying to have zero margin between $\mathcal{D}_i(\mathcal{G}_{i}(\mathbf{z}))$ and $\mathcal{D}_i(\mathbf{x})$ (see Figure~\ref{fig:gogan_training}). The discriminator is trying to push $\mathcal{D}_i(\mathcal{G}_i(\mathbf{z}))$ down to $\mathcal{D}_i(\mathcal{G}_{i-1}(\mathbf{z}))$ so that it gives the same score to the fake samples generated by stage $i$ generator as those generated by stage $i-1$ generator. In other words, the discriminator is trying to become as good in detecting fake samples from $\mathcal{G}_i$ as it is in detecting fake samples from $\mathcal{G}_{i-1}$. This forces the generator to `work harder' to fool the discriminator and give the same score to the fake samples $\mathcal{G}_i(\mathbf{z})$ as to the real samples. This adversarial game leads to the self-improvement of GAN with subsequent stages.

\subsection{Encoder $\mathcal{E}$}
Although RankGAN works even without an encoder, in practice, we have observed that adding an encoder improves the performance and training convergence of RankGAN considerably. This is because adding an encoder allows the discriminator to rank generated and real samples based on image quality and realisticity rather than identity. To obtain the encoder, we first train a VAE~\cite{vae} in the zeroth stage. After the VAE is trained, the encoder is frozen and forms the first component of the RankGAN architecture (see Figure~\ref{fig:flowchart}). During RankGAN training, the encoder takes the real image  $\mathbf{x}$ and outputs a mean $\mathbf{\mu(x)}$ and variance $\mathbf{\Sigma(x)}$ to sample the latent vector as $\mathbf{z} \sim \mathbf{\mathcal{N}(\mu(x), \Sigma(x))}$ which is used by the subsequent stage generators to generate fake samples for training. The VAE decoder can also be used as the zeroth stage generator.

\subsection{Discriminator Penalties}
We enforce Lipschitz constrain on the discriminator using gradient penalty (GP) as proposed by Gulrajani \etal~\cite{wgan-gp}. GP penalizes the norm of the gradient of the discriminator \wrt its input $\hat{\mathbf{x}}\sim \mathds{P}_{\hat{\mathbf{x}}}$, which enforces a soft version of the constraint. The GP loss is given by:
\begin{align}
\label{eqn:wgan_gp}
\mathbf{\mathcal{L}}_{\mathrm{gp}} = \mathds{E}_{\hat{\mathbf{x}}\sim\mathds{P}_{\hat{\mathbf{x}}}}\big[ (\| \nabla_{\hat{\mathbf{x}}} \mathcal{D}(\hat{\mathbf{x}}) \|_2 - 1 )^2 \big]
\end{align}
In addition, Eqn.~\eqref{eqn:ranking_loss} does not prevent the discriminator from cheating by letting $\mathcal{D}_i(\mathbf{x})$ and $\mathcal{D}_i(\mathcal{G}_{i-1}(\mathbf{z}))$ to simultaneously converge to the level of $\mathcal{D}_i(\mathcal{G}_{i}(\mathbf{z}))$ (blue and green curves converging towards the red curve in Figure~\ref{fig:gogan_training}), thereby defeating the purpose of training. To prevent this, we add a penalty term to the overall ranking loss given by:
\begin{align}
    \label{eqn:clamp}
    \mathbf{\mathcal{L}}_{\mathrm{clamp}} = \big[m_i^{\mathrm{high}} - \mathbf{\mathcal{D}}_{i}(\mathbf{x})\big]_{+} + \big[\mathbf{\mathcal{D}}_i(\mathbf{\mathcal{G}}_{i-1}(\mathbf{z})) - m_i^{\mathrm{low}}\big]_{+}
\end{align}
where $m_i^{\mathrm{high}}$ and $m_i^{\mathrm{low}}$ are the high and low margins for stage-$i$ RankGAN respectively. Thus, the clamping loss constraints the discriminator so as not to let $\mathcal{D}_i(\mathbf{x})$ go below $m_i^{\mathrm{high}}$ and $\mathcal{D}_{i-1}(\mathcal{G}_{i}(\mathbf{z}))$ go above $m_i^{\mathrm{low}}$. We call this \textbf{Discriminator Clamping}. The overall discriminator loss thus becomes:
\begin{align}
    \label{eqn:disc_loss}
    \mathbf{\mathcal{L}_{\mathrm{disc}}} = \mathbf{\mathcal{L}_{\mathrm{disc\_rank}}} + \lambda_{\mathrm{gp}} \mathbf{\mathcal{L}_{\mathrm{gp}}} + \lambda_{\mathrm{clamp}} \mathbf{\mathcal{L}_{\mathrm{clamp}}}
\end{align}
In our experiments, we find $\lambda_{\mathrm{gp}} = 10$ and $\lambda_{\mathrm{clamp}}=1000$ to give good results.

\begin{algorithm}[!ht]
\footnotesize
\SetAlgoLined
 $\alpha_\mathcal{D}, \alpha_\mathcal{G} \leftarrow 5e-5, \alpha_\mathcal{E} \leftarrow 1e-4$\;
 \For{$i = 1\text{ }...\text{ } nstages$}{
  \eIf{$i = 1$}{
  train \textbf{VAE} with \textbf{Encoder} $\mathbf{\mathcal{E}}$ and \textbf{Decoder} $\mathbf{\mathcal{G}}_1$\;
  train \textbf{Discriminator} $\mathbf{\mathcal{D}}_1$ for 1 epoch using WGAN loss of Eqn.~\ref{eqn:wgan_gp}\;
  }{
  $j, k \leftarrow 0, 0$\;
  initialize $\mathbf{\mathcal{D}}_{i} \leftarrow \mathbf{\mathcal{D}}_{i-1}$ and $\mathbf{\mathcal{G}}_{i} \leftarrow \mathbf{\mathcal{G}}_{i-1}$\;
  freeze $\mathbf{\mathcal{D}}_{i-1}$ and $\mathbf{\mathcal{G}}_{i-1}$\;
  compute $m_i^{\mathrm{high}} = \mathds{E}[\mathbf{\mathcal{D}}_{i-1}(\mathbf{x}_{\mathrm{val}})]$ and $m_i^{\mathrm{low}} = \mathds{E}[\mathbf{\mathcal{D}}_{i-1}(\mathbf{\mathcal{G}}_{i-1}(\mathbf{z}))]$\;
  \While{$j < nepochs$}{
    \While{$k < 5$}{
     obtain real samples $\mathbf{x}$ and latent vectors $\mathbf{z} \sim \mathbf{\mathcal{E}(x)}$\;
     compute $\mathbf{\mathcal{L}}_{\mathrm{disc}}$ using Eqn.~\ref{eqn:disc_loss}\;
     optimize $\mathbf{\mathcal{D}}_i$ using $AdamOptimizer(\alpha_\mathcal{D}, \beta_1=0, \beta_2=0.99)$\;
     $j \leftarrow j+1$, $k \leftarrow k+1$}
     compute $\mathbf{\mathcal{L}}_{\mathrm{gen}}$ using Eqn.~\ref{eqn:ranking_loss}\;
     optimize $\mathbf{\mathcal{G}}_i$ using $AdamOptimizer(\alpha_\mathcal{G}, \beta_1=0, \beta_2=0.99)$\;
     $k \leftarrow 0$
  }
  }
 }
 \caption{RankGAN Training}\label{algo:gogan}
\end{algorithm}

\section{Experiments}
In this section, we describe our experiments evaluating the effectiveness of the RankGAN against traditional GAN variants \ie, WGAN and LSGAN. For this purpose, we trained the RankGAN, WGAN and LSGAN models on face images and evaluated their performance on face generation and face completion tasks. Due to space limit, we will omit some implementation details in the paper. Full implementation details will be made publicly available.

\subsection{Database and Metrics}
We use the \textbf{CelebA} dataset \cite{celebA} which is a large-scale face attributes dataset with more than 200K celebrity images covering large pose variations and background clutter. The face images are pre-processed and aligned into an image size of $64\times64$ while keeping a 90-10 training-testing split.

To compare the performance of RankGAN and other GAN variants quantitatively, we computed several metrices including Inception Score~\cite{inception-score} and Fr\'{e}chet Inception distance (FID)~\cite{fid}. Although, Inception score has rarely been used to evaluate face generation models before, we argue that since it is based on sample entropy, it will favor sharper and more feature-full images. The FID, on the other hand, captures the similarity of the generated images to the real ones, thus capturing their realisticity and fidelity.

\subsection{Evaluations on Face Generation Tasks}
For all the experiments presented in this paper, we use the same network architecture based on the one used in \cite{progressive}. Both the discriminators and generators are optimized using the Adam optimizer \cite{adam} with $\beta_1 = 0.0$ and $\beta_2 = 0.99$ and a learning rate of $5e-5$. The criterion to end a stage is based on the convergence of that particular stage and is determined empirically. In practice, we terminate a stage when either the discriminator gap stabilizes for 10-20 epochs or at least 200 stage-epochs are finished, whichever is earlier. Lastly, no data augmentation was used for any of our experiments.

\begin{figure*}[!ht]
    \centering
    \subfigure[Stage 1, Open Set.]{\includegraphics[width=0.32\linewidth]{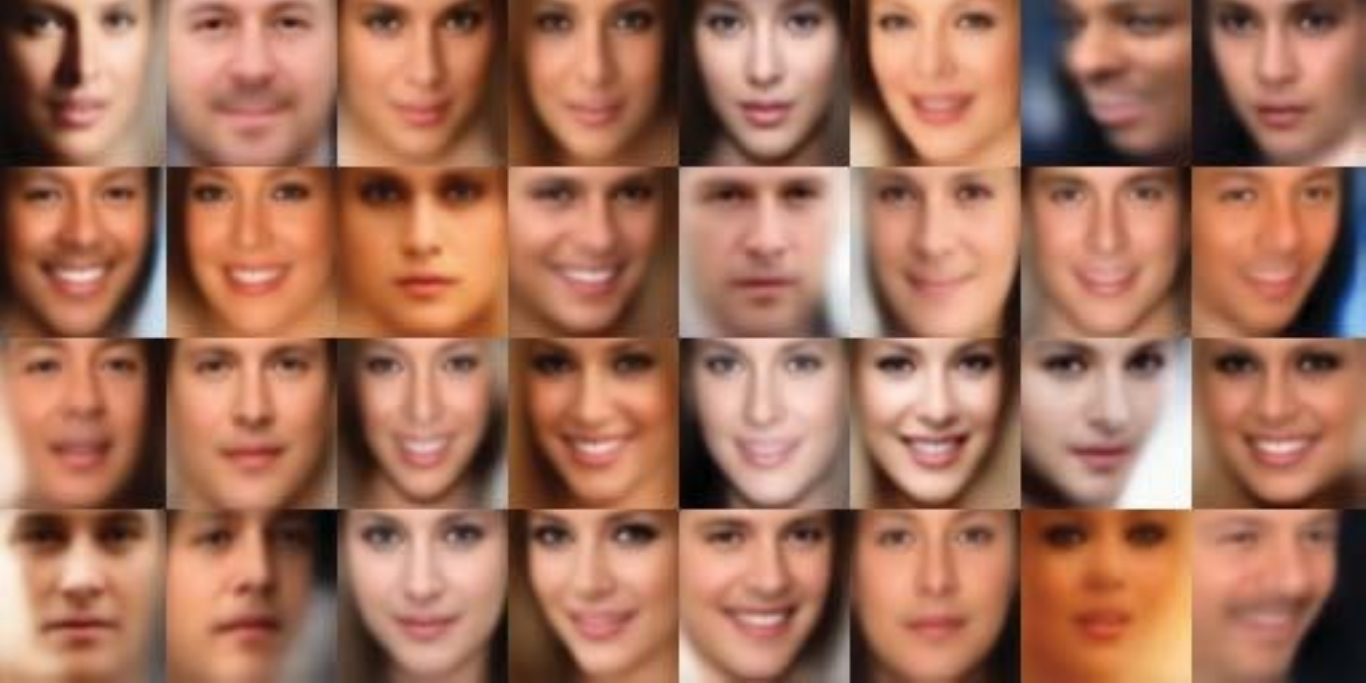}}
    \subfigure[Stage 2, Open Set.]{\includegraphics[width=0.32\linewidth]{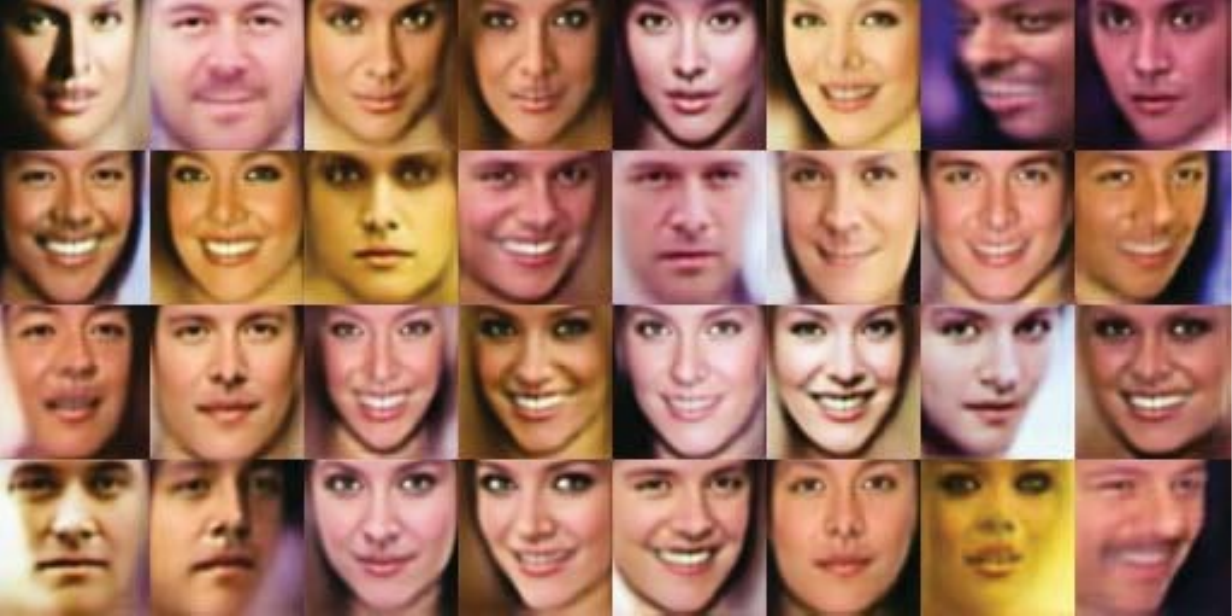}}
    \subfigure[Stage 3, Open Set.]{\includegraphics[width=0.32\linewidth]{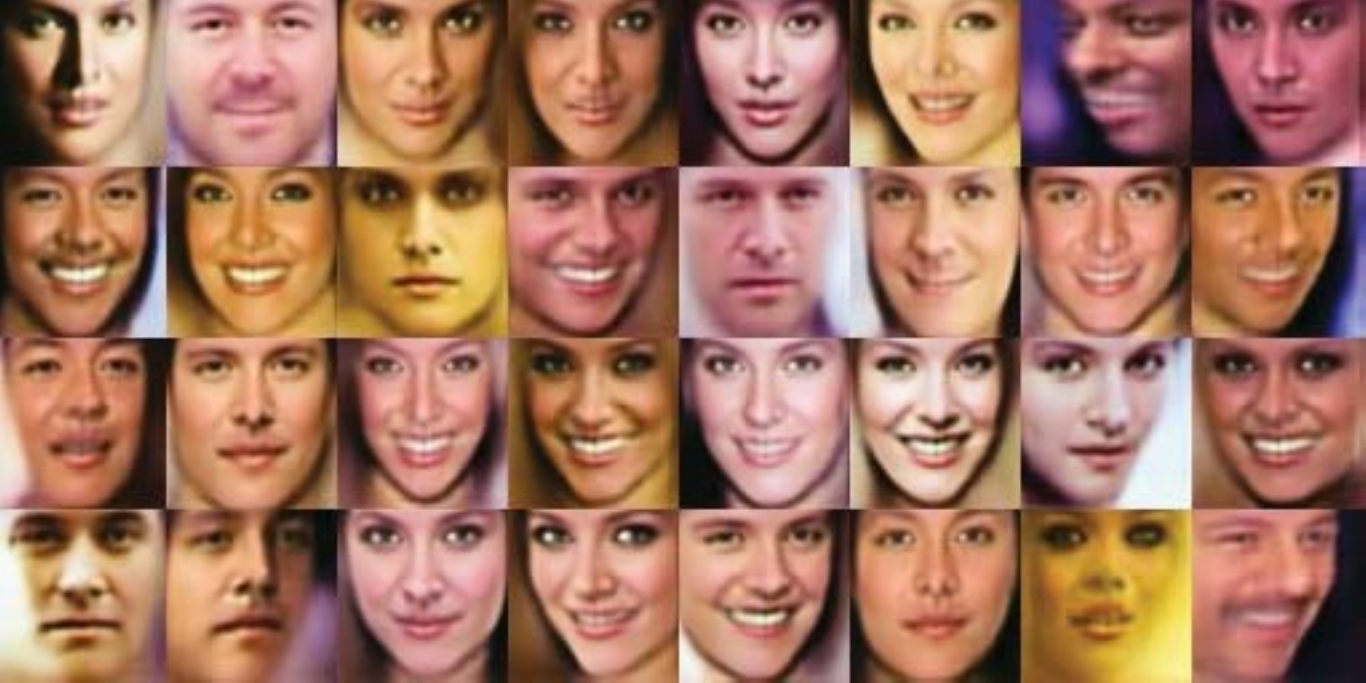}}
\caption{Face generation with RankGAN. Latent vectors $\mathbf{z}$'s are randomly generated w/o encoder $\mathcal{E}$.}\label{fig:gen_normal_z}
\end{figure*}

Figure~\ref{fig:gen_encoder_z} shows the visual progression of Open-Set face generation results from various stages in RankGAN when the latent vector $\mathbf{z}$ is obtained by passing the input faces through the encoder ${\mathcal{E}}$. Figure~\ref{fig:gen_normal_z} shows the visual progression of face generation results when the latent vectors $\mathbf{z}$'s are randomly generated without the encoder ${\mathcal{E}}$. In both the cases, we can clearly see that as the stage progresses, RankGAN is able to generate sharper face images which are visually more appealing.

Quantitative results are consolidated in Table~\ref{tab:gen_results} with FID (the lower the better) and Inception score (the higher the better). As can be seen, as the training progresses from stage-1 to stage-3, the trend conforms with the visual results where stage-3 yields the highest Inception score and the lowest FID.
\begin{table}[t]
\centering
\scriptsize
\caption{Quantitative results for face image generation with and without the encoder.}
\begin{tabular}{|c|c|c|c|c|}
  \hline
  & \multicolumn{2} {c|} {With Encoder} &\multicolumn{2}{c|} {Without Encoder} \\\hline
  ~       & FID  & Inception Score & FID  & Inception Score\\\hline
  Real    & N/A    & 2.51 & N/A    & 2.51 \\
  Stage-1 & 122.17 & 1.54 & 140.45 & 1.54 \\
  Stage-2 & 60.45  & 1.78 & 75.53  & 1.75 \\
  Stage-3 & \textbf{46.01}  & \textbf{1.89} & \textbf{63.34}  & \textbf{1.91} \\ 
  \hline
\end{tabular}
\label{tab:gen_results}
\end{table}
\begin{figure*}[!ht]
    \centering
    \subfigure[Input faces.]{\includegraphics[width=0.49\linewidth]{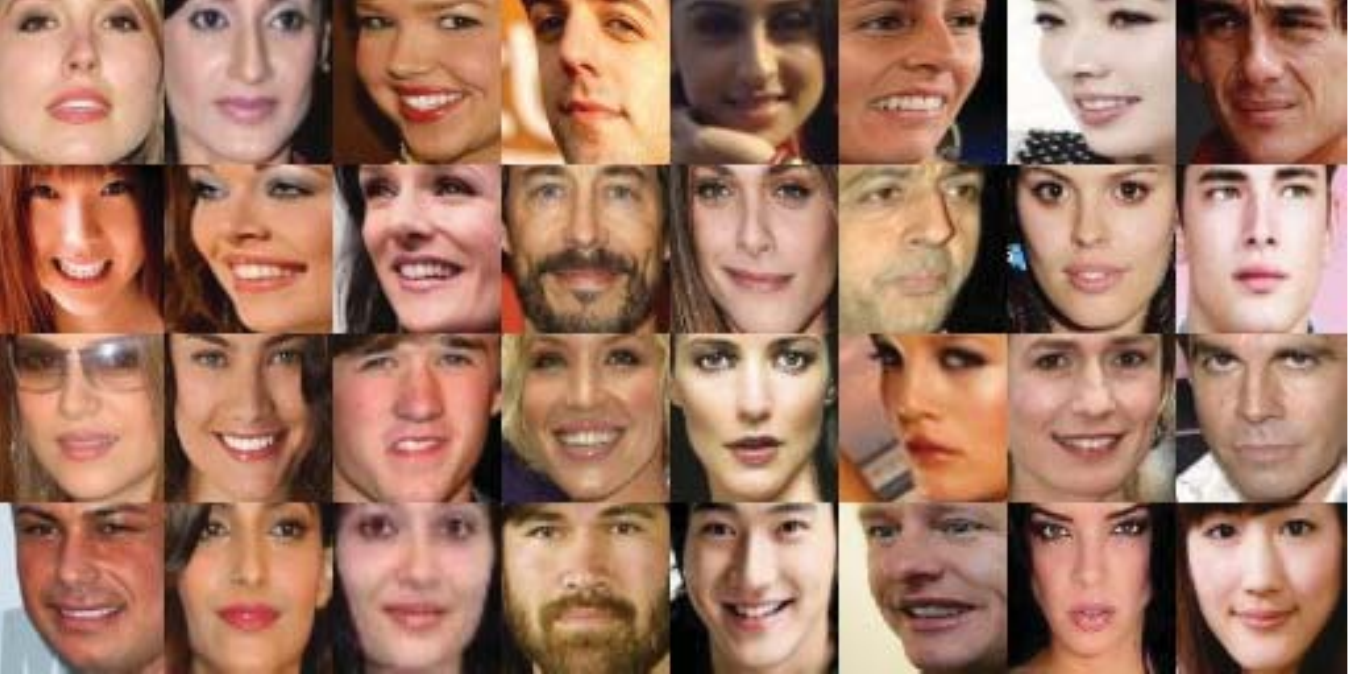}}
    \subfigure[Stage 1 generated faces, Open Set.]{\includegraphics[width=0.49\linewidth]{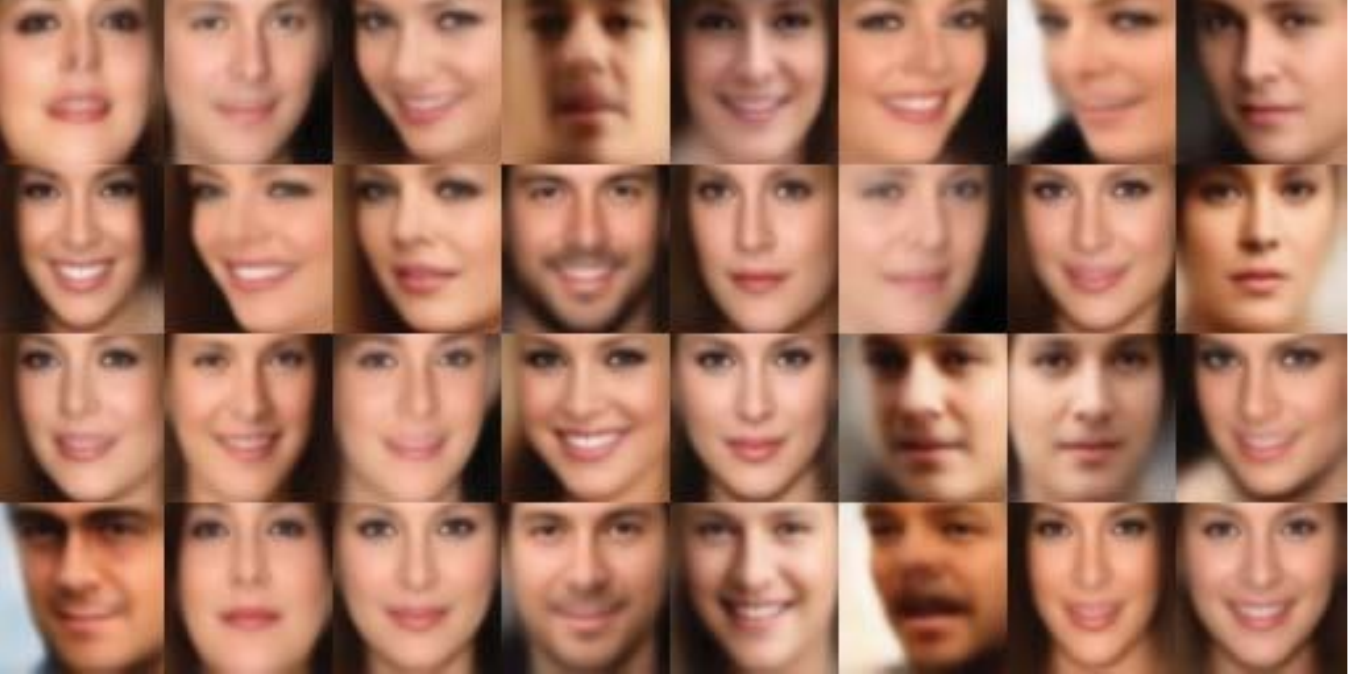}}
    \subfigure[Stage 2 generated faces, Open Set.]{\includegraphics[width=0.49\linewidth]{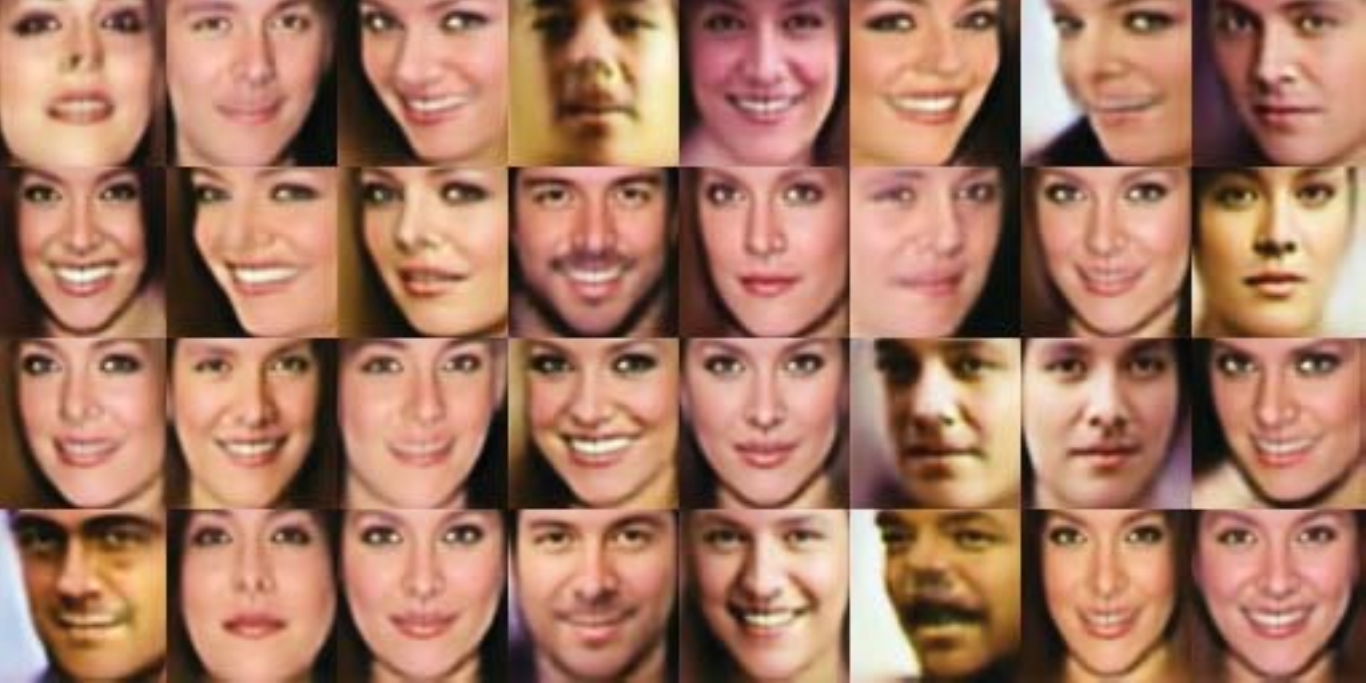}}
    \subfigure[Stage 3 generated faces, Open Set.]{\includegraphics[width=0.49\linewidth]{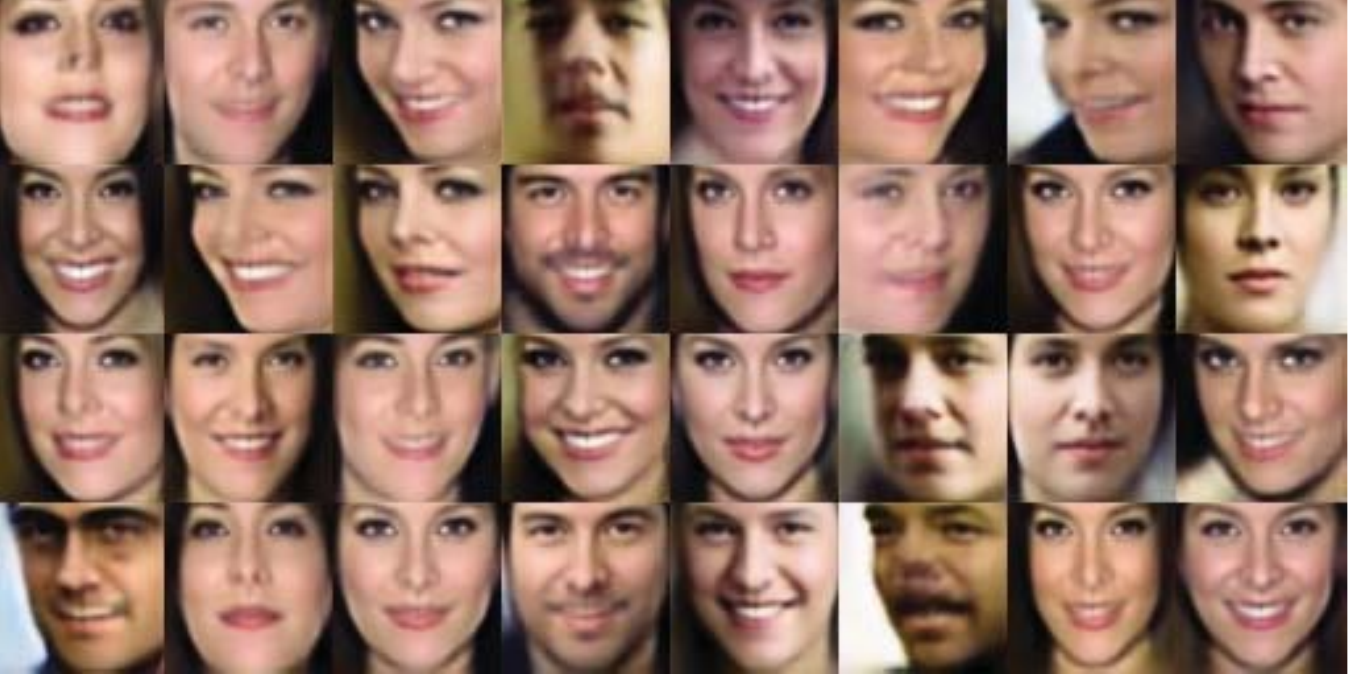}}
\caption{Face generation with RankGAN. Latent vectors $\mathbf{z}$'s are obtained by passing the input faces through the encoder $\mathcal{E}$.}\label{fig:gen_encoder_z}
\end{figure*}
\begin{figure*}[!ht]
\centering
   \includegraphics[width=0.32\linewidth]{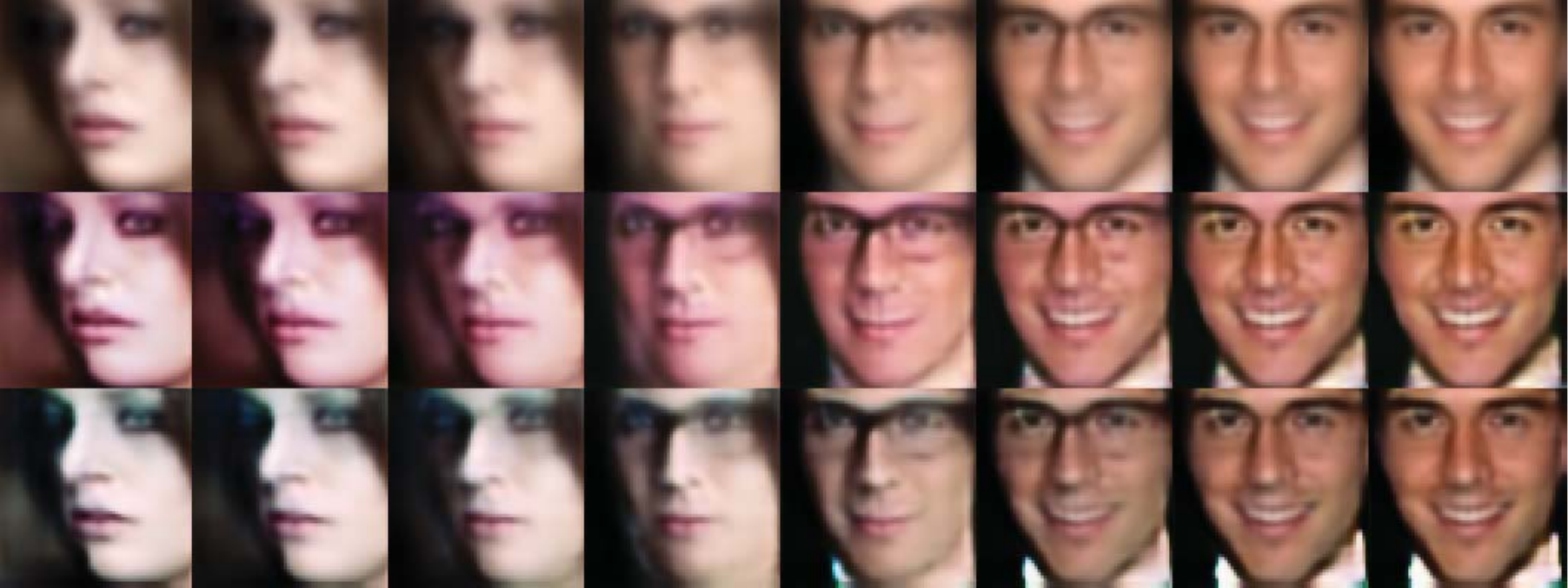}
   \includegraphics[width=0.32\linewidth]{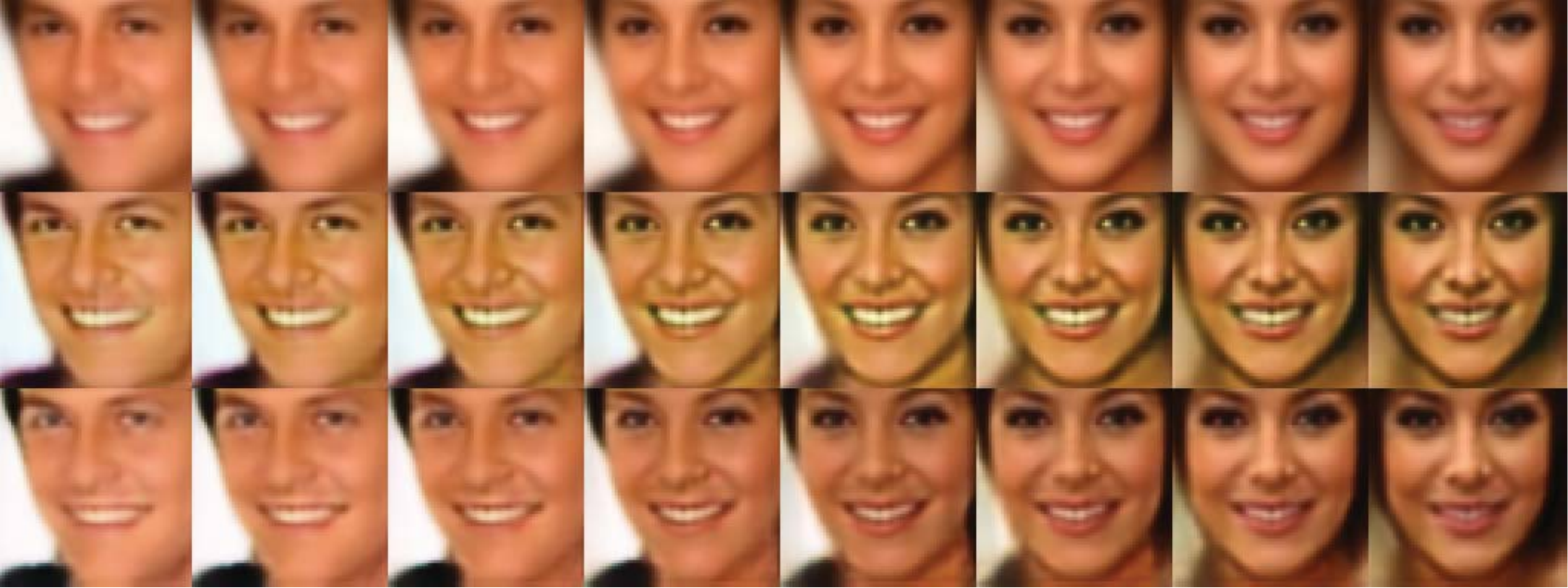}
   \includegraphics[width=0.32\linewidth]{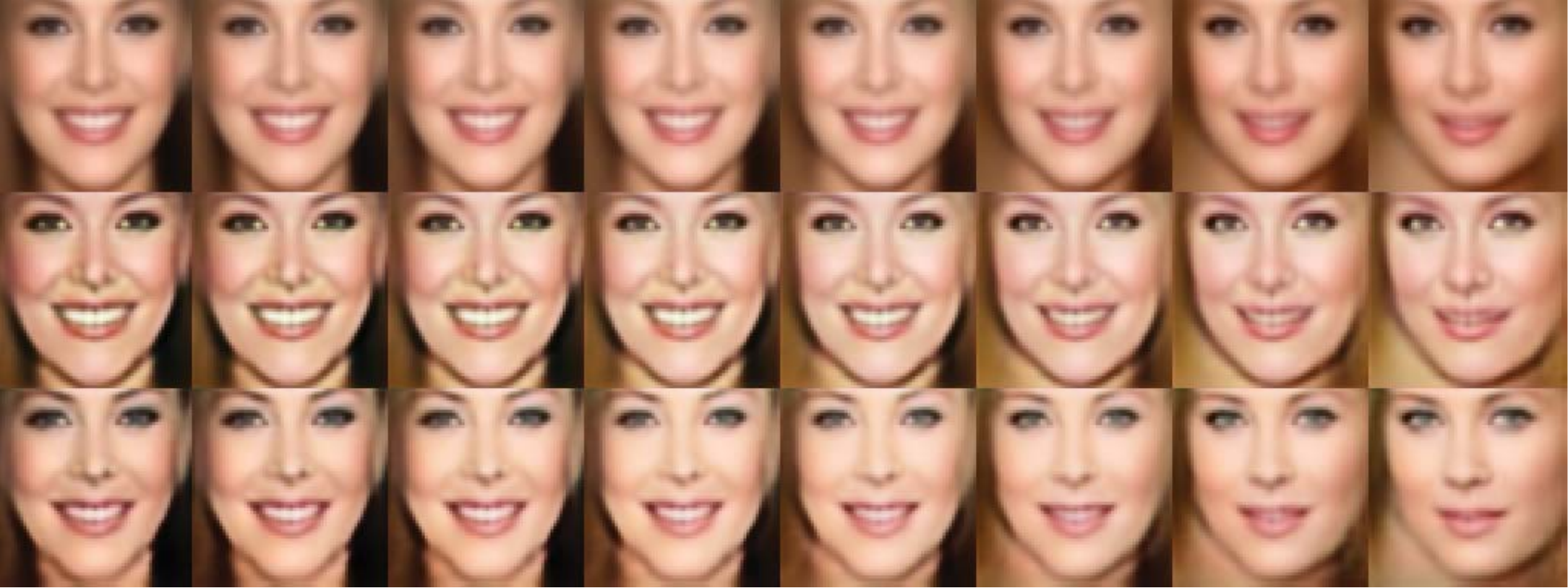}
\caption{Interpolation between two latent vectors which are obtained by passing the input faces through the encoder $\mathcal{E}$. The 3 rows within each montage correspond to Stage 1, 2, and 3 in RankGAN.}\label{fig:interp_encoder_z}
\end{figure*}
\begin{figure*}[!ht]
    \centering
    \includegraphics[width=0.32\linewidth]{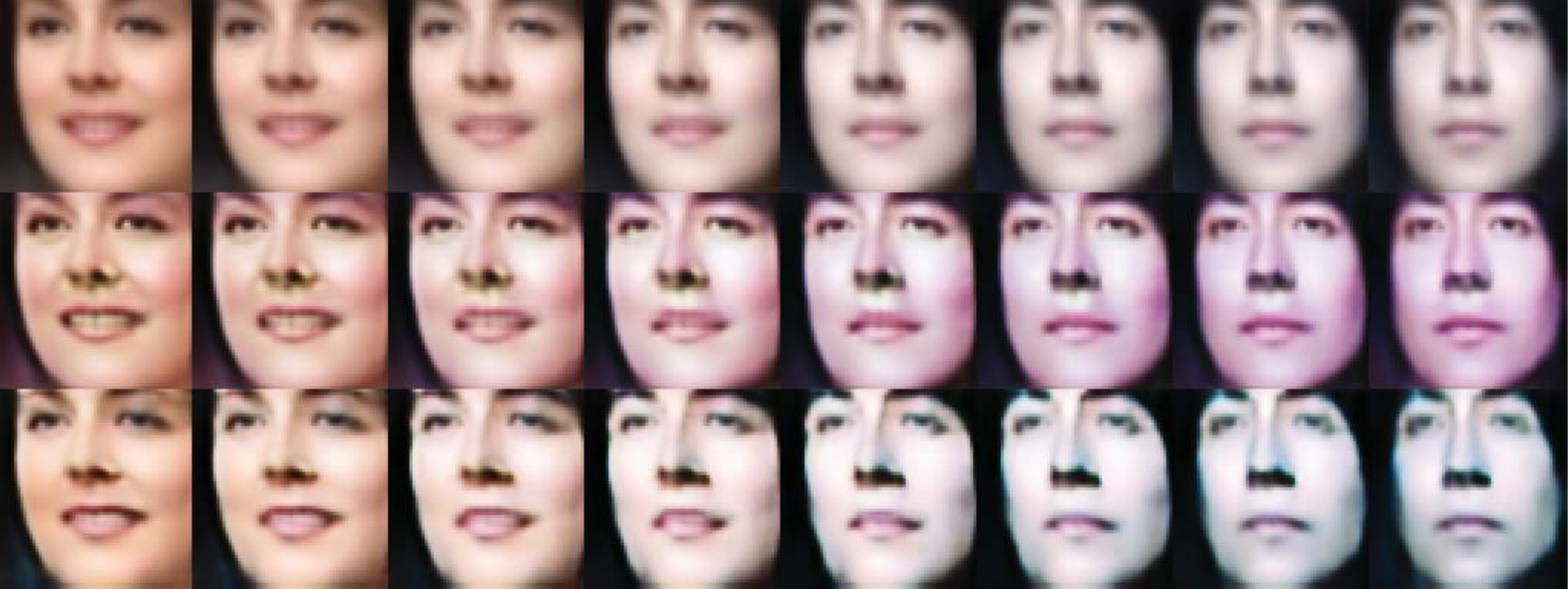}
    \includegraphics[width=0.32\linewidth]{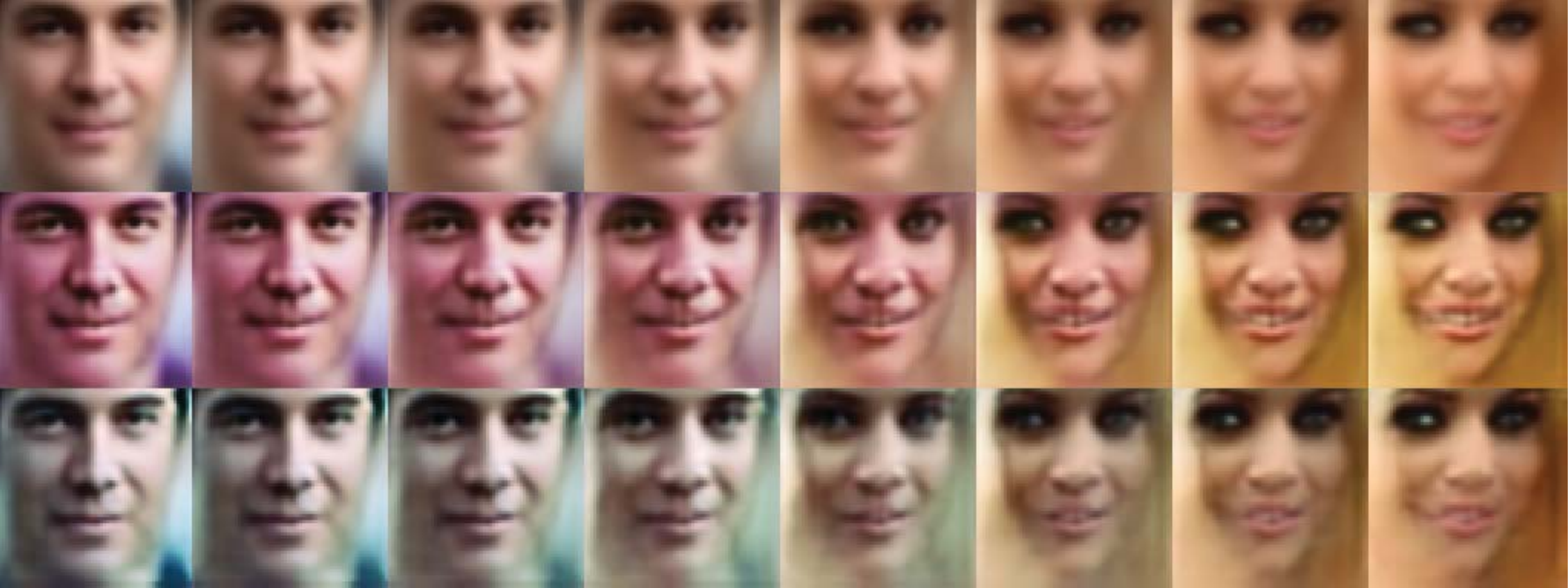}
    \includegraphics[width=0.32\linewidth]{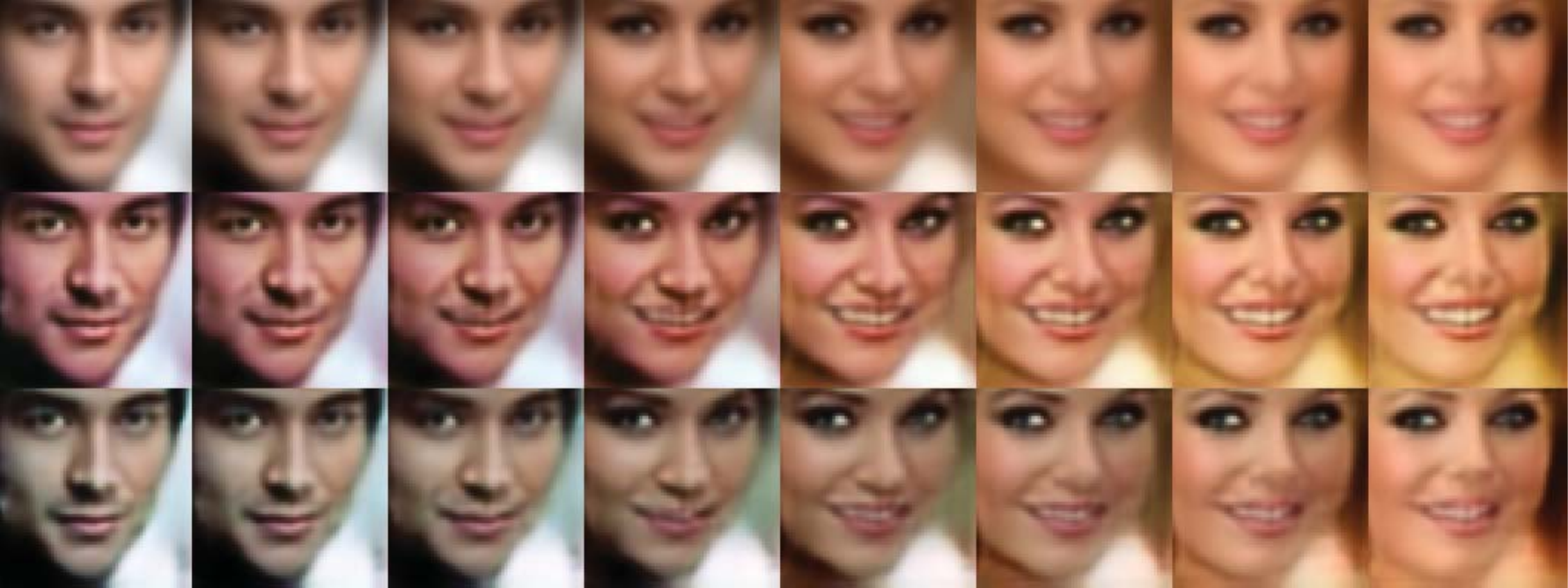}
\caption{Interpolation between two latent vectors that are randomly selected (without the encoder $\mathcal{E}$) from a unit normal distribution. The 3 rows within each montage correspond to Stage 1, 2, and 3 in RankGAN.}\label{fig:interp_normal_z}
\end{figure*}

\subsection{Evaluations on Face Completion Tasks}
A good generative model should perform well on missing data problems. Motivated by this argument, we propose to use image completion as a quality measure for GAN models. In short, the quality of the GAN models can be quantitatively measured by the image completion fidelity, in terms of PSNR, SSIM and other metrics. Traditional shallow methods \cite{felix_cvprw14_hallucinate,felix_btas16_fastfood} have shown some promising results but still struggle when dealing with face variations. Deep learning methods based on GANs are expected to handle image variations much more effectively. To take on the image completion task, we need to utilize both the $\mathcal{G}$ and $\mathcal{D}$ from the RankGAN and the baselines WGAN and LSGAN, pre-trained with uncorrupted data. After training, $\mathcal{G}$ is able to embed the images from $p_\mathrm{data}$ onto some non-linear manifold of $\mathbf{z}$. An image that is not from $p_\mathrm{data}$ (\eg, with missing pixels) should not lie on the learned manifold. We seek to recover the image $\mathbf{\hat{y}}$ on the manifold ``closest'' to the corrupted image $\mathbf{y}$ as the image completion result. To quantify the ``closest'' mapping from $\mathbf{y}$ to the reconstruction, we define a function consisting of contextual and perceptual losses \cite{dcgan_semantic}.
%
The \textbf{contextual loss} measures the fidelity between the reconstructed image portion and the uncorrupted image portion, and is defined as:
\begin{align}
\mathcal{L}_\mathrm{contextual}(\mathbf{z}) = \| \mathbf{M} \odot  \mathcal{G}(\mathbf{z})-\mathbf{M} \odot  \mathbf{y}\|_1
\label{eq:occlusion_mask}
\end{align}
where $\mathbf{M}$ is the binary mask of the uncorrupted region and $\odot$ denotes the Hadamard product.
%
The \textbf{perceptual loss} encourages the reconstructed image to be similar to the samples drawn from the training set (true distribution $p_\mathrm{data}$). This is achieved by updating $\mathbf{z}$ to fool $\mathcal{D}$, or equivalently by maximizing $\mathcal{D}(\mathcal{G}(\mathbf{z}))$. As a result, $\mathcal{D}$ will predict $\mathcal{G}(\mathbf{z})$ to be from the real data with a high probability.
\begin{align}
\mathcal{L}_\mathrm{perceptual}(\mathbf{z}) = - \mathcal{D}( \mathcal{G} (\mathbf{z}))
\end{align}
%
Thus, $\mathbf{z}$ can be updated, using backpropagation, to lie closest to the corrupted image in the latent representation space by optimizing the objective function:
\begin{align}
\hat{\mathbf{z}} = \argmin_{\mathbf{z}}(\mathcal{L}_\mathrm{contextual}(\mathbf{z}) + \lambda \mathcal{L}_\mathrm{perceptual}(\mathbf{z}))
\end{align}
where $\lambda$ (set to 10 in our experiments) is a weighting parameter. After finding the optimal solution $\hat{\mathbf{z}}$, the reconstructed image $\mathbf{y}_\mathrm{completed}$ can be obtained by:
\begin{align}
\mathbf{y}_\mathrm{completed} = \mathbf{M} \odot \mathbf{y} + (1 - \mathbf{M}) \odot \mathcal{G}(\hat{\mathbf{z}})
\end{align}
\subsubsection{Metrics:}In addition to the FID and Inception Score, we used metrics such as PSNR \cite{felix_cvprw15_nir}, SSIM, OpenFace \cite{openface} feature distance under normalized cosine similarity (NCS) \cite{felix_tip15_spartans} and PittPatt face matching score \cite{felix_cvprw15_cots} to measure fidelity between the original and reconstructed face images. The last two are off-the-shelf face matchers that can be used to examine the similarity between pairs of face images. For these two matchers, we also obtain the area under the ROC curves (AUC) score as an auxiliary metric.

\subsubsection{Occlusion Masks:}We carried out face completion experiments on four types of facial masks, which we termed as: `Center Small', `Center Large', `Periocular Small', and `Periocular Large'.

\subsubsection{Open-Set:}It is important to note that all of our experiments are carried out in an Open-Set fashion, \ie, none of the images and subjects were seen during training. This is of course a more challenging setting than Closed-Set and reflects the generalization performance of these models.

\subsubsection{Discussion:}
\begin{figure*}[!ht]
    \centering
    \subfigure[Original faces.]{\includegraphics[width=0.455\linewidth]{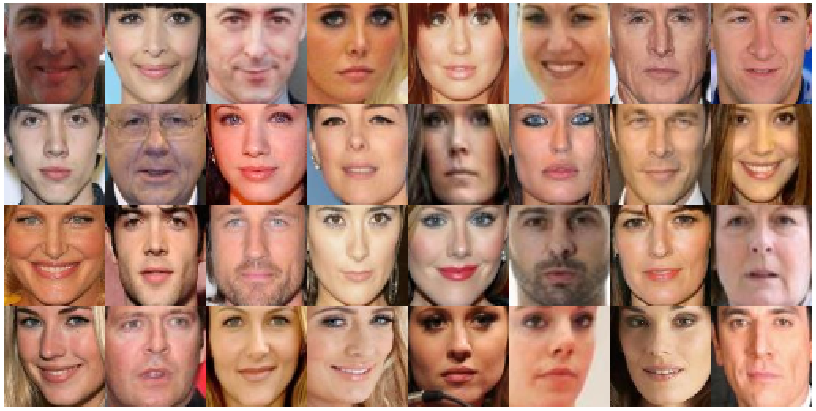}}
    \subfigure[Masked faces. (`Center Large')]{\includegraphics[width=0.455\linewidth]{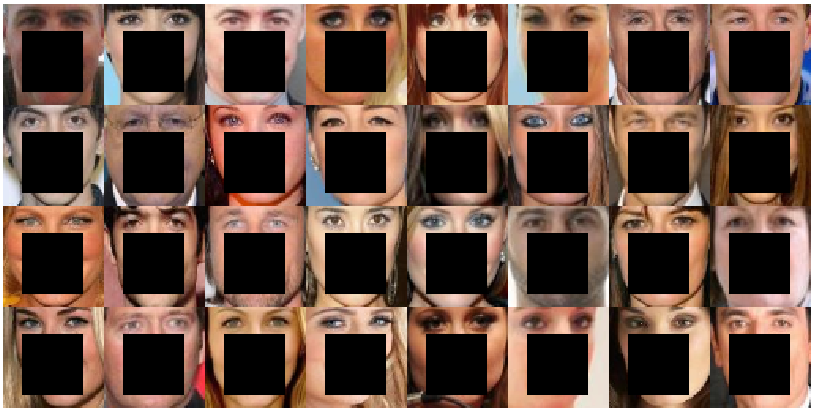}}
    \subfigure[WGAN completion, Open Set.]{\includegraphics[width=0.455\linewidth]{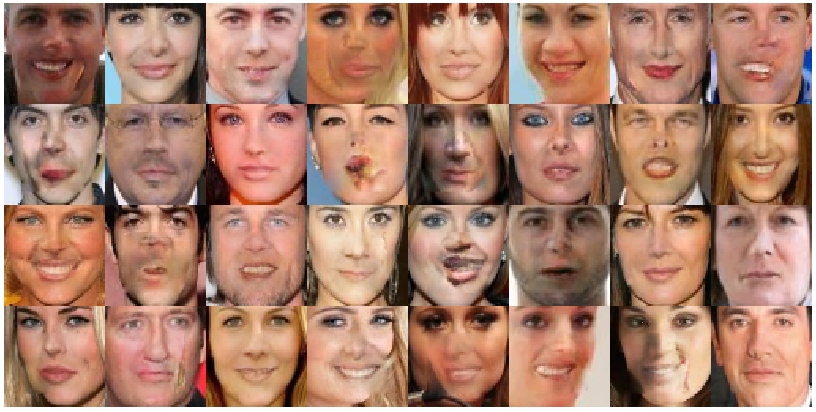}}
    \subfigure[LSGAN completion, Open Set.]{\includegraphics[width=0.455\linewidth]{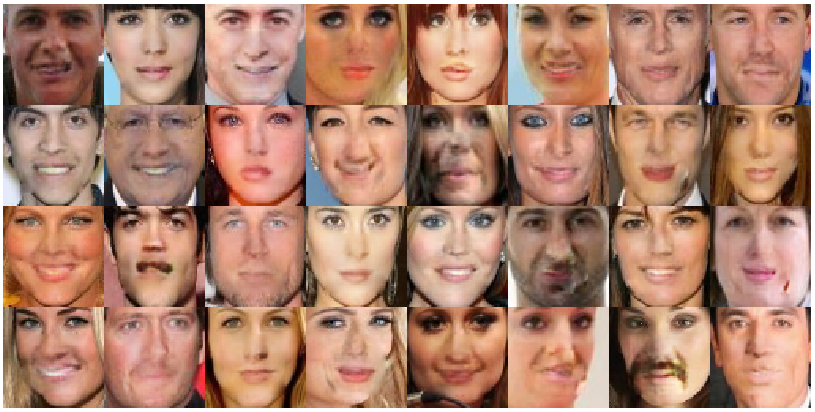}}
    \subfigure[Stage 1 completion, Open Set.]{\includegraphics[width=0.455\linewidth]{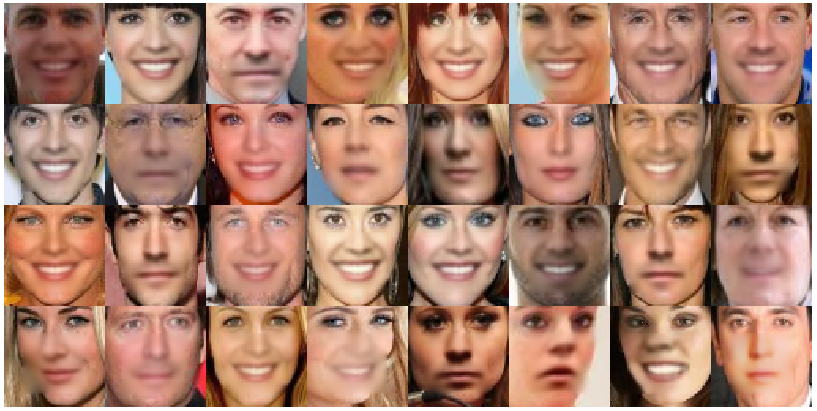}}
    \subfigure[Stage 2 completion, Open Set.]{\includegraphics[width=0.455\linewidth]{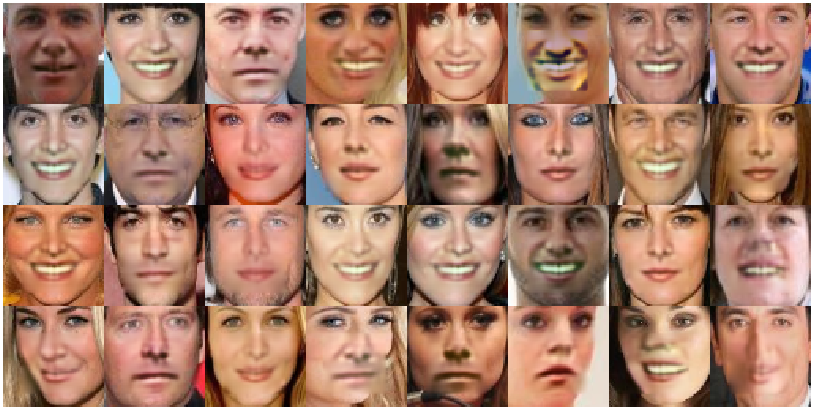}}
    \subfigure[Stage 3 completion, Open Set.]{\includegraphics[width=0.455\linewidth]{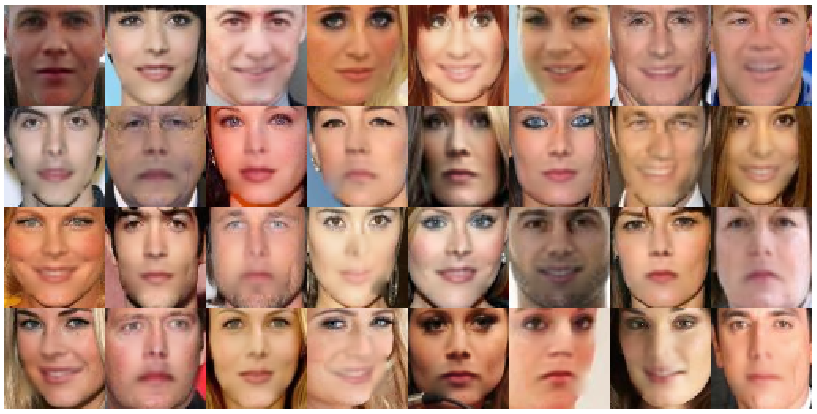}}
\caption{\textcolor{blue}{Best} completion results with RankGAN on CelebA, `Center Large' mask.}\label{fig:montage_celeba_good_center_lg}
\end{figure*}

Due to lack of space, we only show results based on the Center Large mask in the main paper (more qualitative and quantitative results can be found in the supplementary). These results have been summarized in Table~\ref{tab:completion_celeb_centerLg} and can be visualized in Figure~\ref{fig:montage_celeba_good_center_lg}. As can be seen in Table~\ref{tab:completion_celeb_centerLg}, RankGAN Stage-3 outperforms all other baselines in all metrics except FID. The lower FID value for WGAN can be attributed to the fact that FID captures distance between two curves and is, in a way, similar to the Wasserstein distance that is minimized in the case of WGAN. The Stage-3 images appear to be both sharp (as measured by the Inception Score) as well as fidelity-preserving as compared to the original images (as measured by identity matching metrics).
All the four identity-based metrics, PSNR, SSIM, OpenFace scores, and PittPatt scores are higher for Stage-3 of RankGAN. This is due to the fact that our formulation enforces identity-preservation through the encoder and the ranking loss.

\begin{table}[t]
\centering
\scriptsize
\caption{Data: CelebA, Mask: Center Large}
\begin{tabular}{|c|c|c|c|c|c|c|}
\hline
  ~       & FID     & Inception & PSNR & SSIM & OpenFace (AUC) & PittPatt (AUC)\\\hline
  Original& N/A & 2.3286 & N/A & N/A        & 1.0000 (0.9965) & 19.6092 (0.9109) \\ \hline
  Stage-1 & 27.09 & 2.1524 & 22.76 & 0.7405 & 0.6726 (0.9724) & 10.2502 (0.7134) \\
  Stage-2 & 23.69 & 2.1949 & 21.87 & 0.7267 & 0.6771 (0.9573) & 9.9718 (0.8214)  \\
  Stage-3 & 27.31 & \textbf{2.2846} & \textbf{23.30} & \textbf{0.7493} & \textbf{0.6789 (0.9749)} & \textbf{10.4102 (0.7922)} \\ \hline
  WGAN    & \textbf{17.03} & 2.2771 & 23.26 & 0.7362 & 0.5554 (0.9156) & 8.1031 (0.7373) \\ \hline
  LSGAN   & 23.93 & 2.2636 & 23.11 & 0.7361 & 0.6676 (0.9659) &	10.1482 (0.7154) \\
  \hline
\end{tabular}
\label{tab:completion_celeb_centerLg}
\end{table}
\section{Conclusions}\label{sec-gogan:concl}
In this work, we introduced a new loss function to train GANs - the margin loss, that leads to a better discriminator and in turn a better generator. We then extended the margin loss to a margin-based ranking loss and evolved a new multi-stage GAN training paradigm that progressively strengthens both the discriminator and the generator.
We also proposed a new way of measuring GAN quality based on image completion tasks. We have seen both visual and quantitative improvements over the baselines WGAN and LS-GAN on face generation and completion tasks. 

\appendix
\chapter*{Appendix}
We detail some of the face generation experimental results (in Section~\ref{sec:supp-1}) and face completion experimental results (in Section~\ref{sec:supp-2}) in the appendix. These results demonstrate the effectiveness of the proposed RankGAN method. Further, we mention sample aware vs. sample agnostic RankGAN in Section~\ref{sec:supp-3}. Finally, we discuss a connection between margin loss and $f$-divergence in Section~\ref{sec:supp-4}.

\section{Face Generation Experiments}\label{sec:supp-1}
In Figure~\ref{fig:gen_encoder_z}, we show face generation results from all the three stages of the RankGAN and the WGAN and LSGAN baselines. These images were generated by passing the input images through the encoder $\mathcal{E}$ and using the obtained latent vectors $\mathbf{z}$ as input to the generators. Since WGAN and LSGAN were trained without an encoder, image generation experiments don't preserve identity for them. One can observe that, the Stage-3 generated images look much more aesthetically appealing and realistic than both WGAN and LSGAN images.
\begin{figure*}[ht]
    \centering
    \subfigure[Input faces.]{\includegraphics[width=0.49\linewidth]{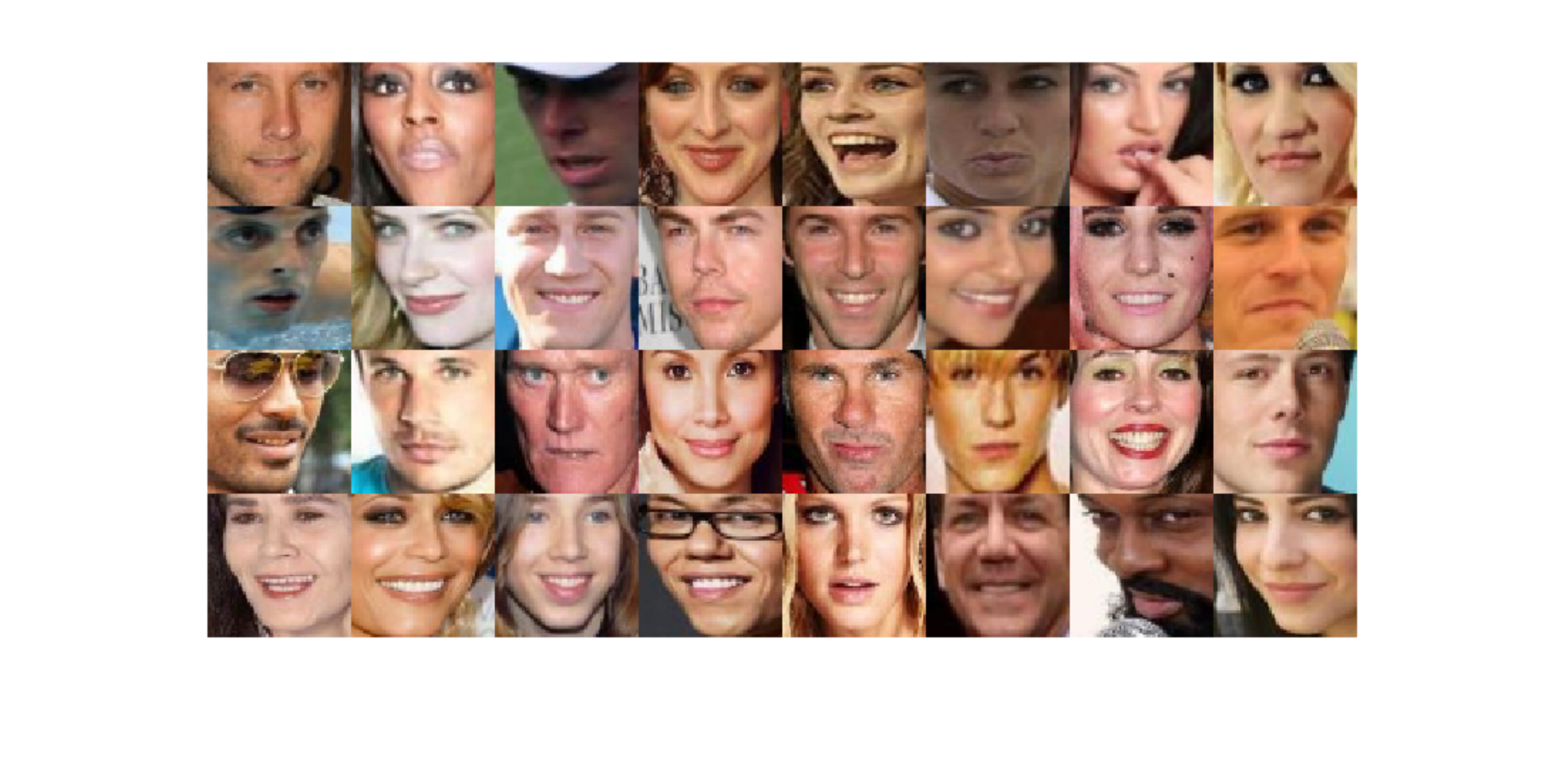}}
    \subfigure[Stage 1 generated faces, Open Set.]{\includegraphics[width=0.49\linewidth]{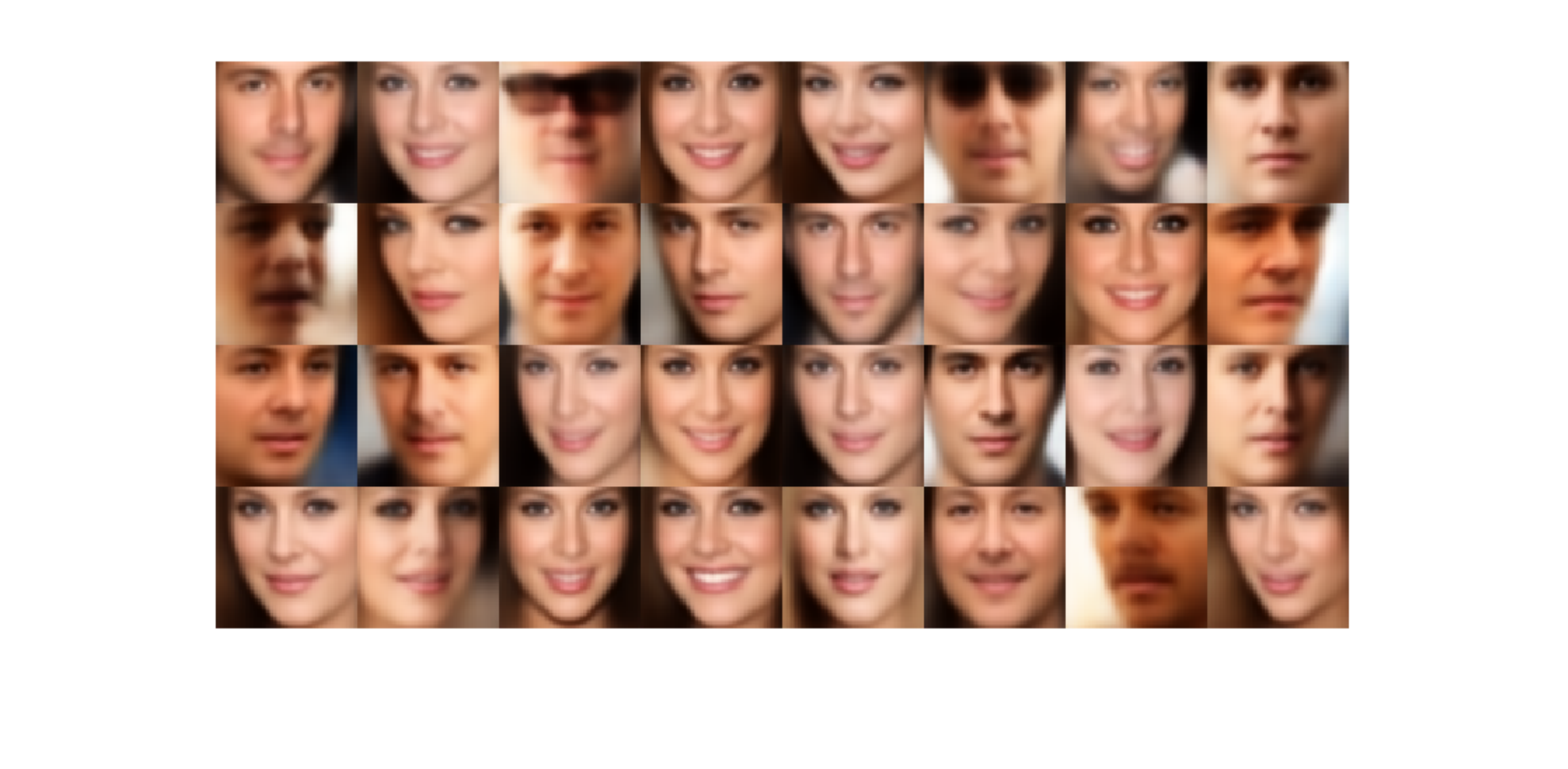}}
    \subfigure[Stage 2 generated faces, Open Set.]{\includegraphics[width=0.49\linewidth]{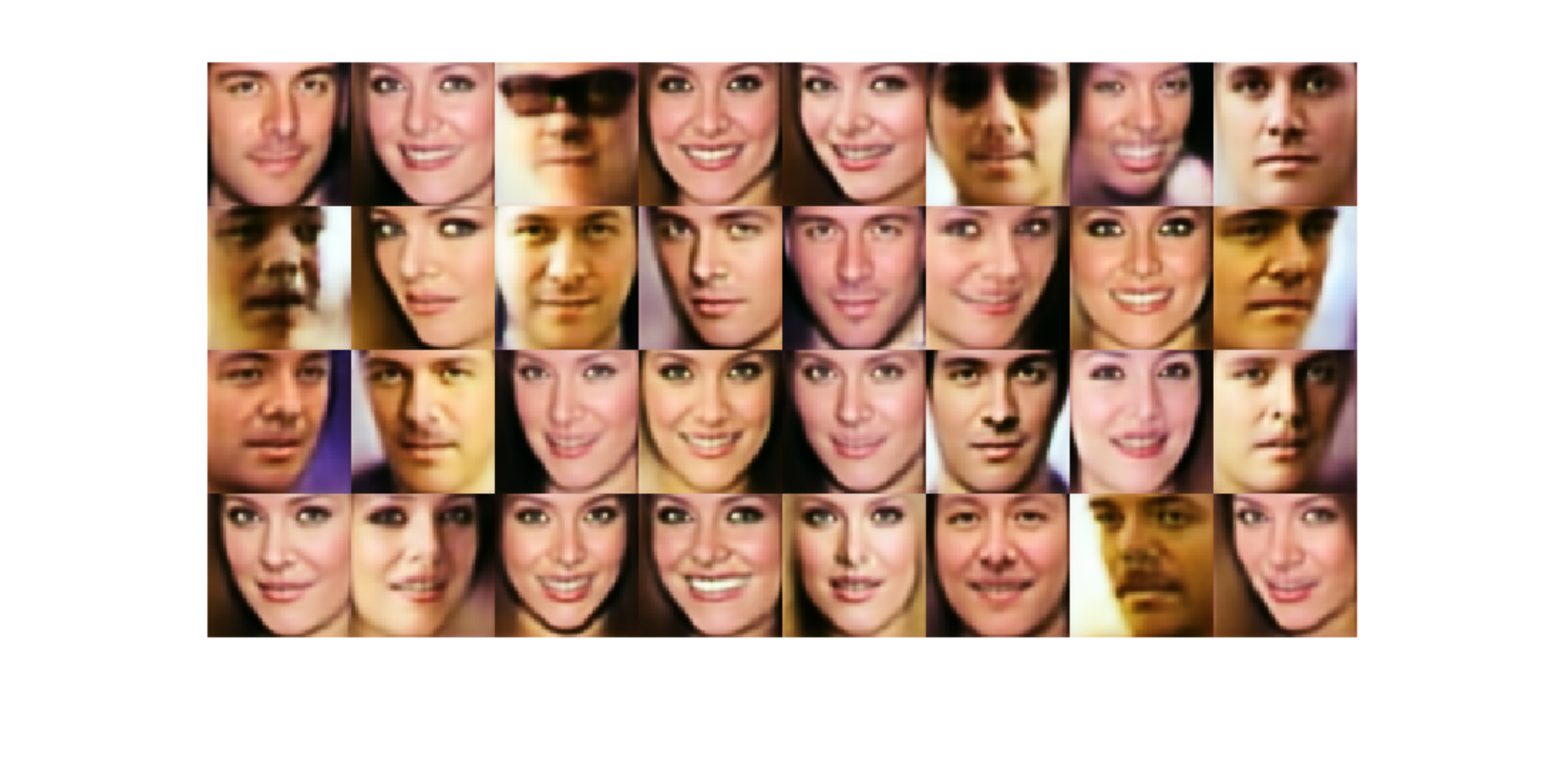}}
    \subfigure[Stage 3 generated faces, Open Set.]{\includegraphics[width=0.49\linewidth]{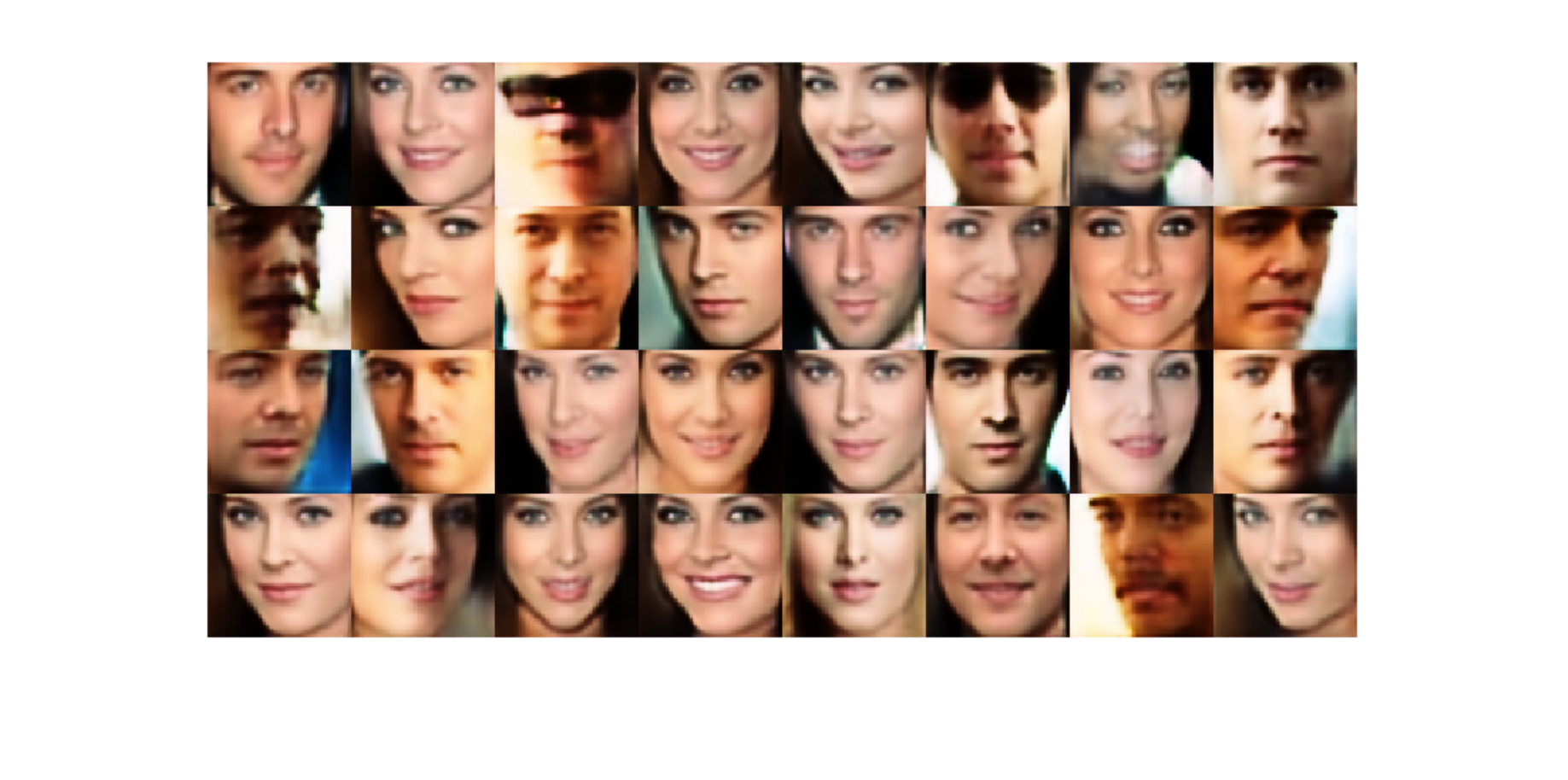}}
    \subfigure[WGAN generated faces, Open Set.]{\includegraphics[width=0.49\linewidth]{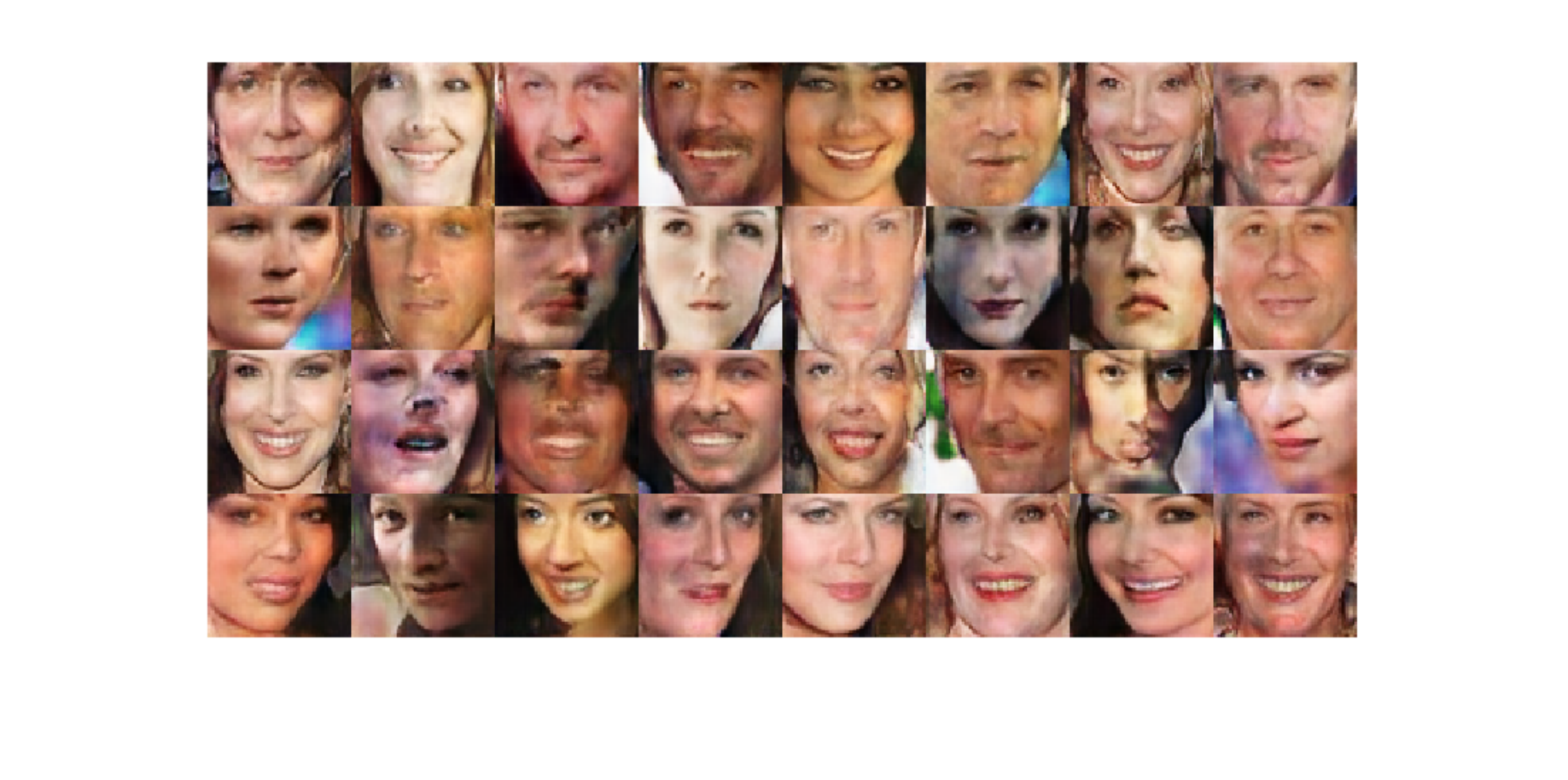}}
    \subfigure[LSGAN generated faces, Open Set.]{\includegraphics[width=0.49\linewidth]{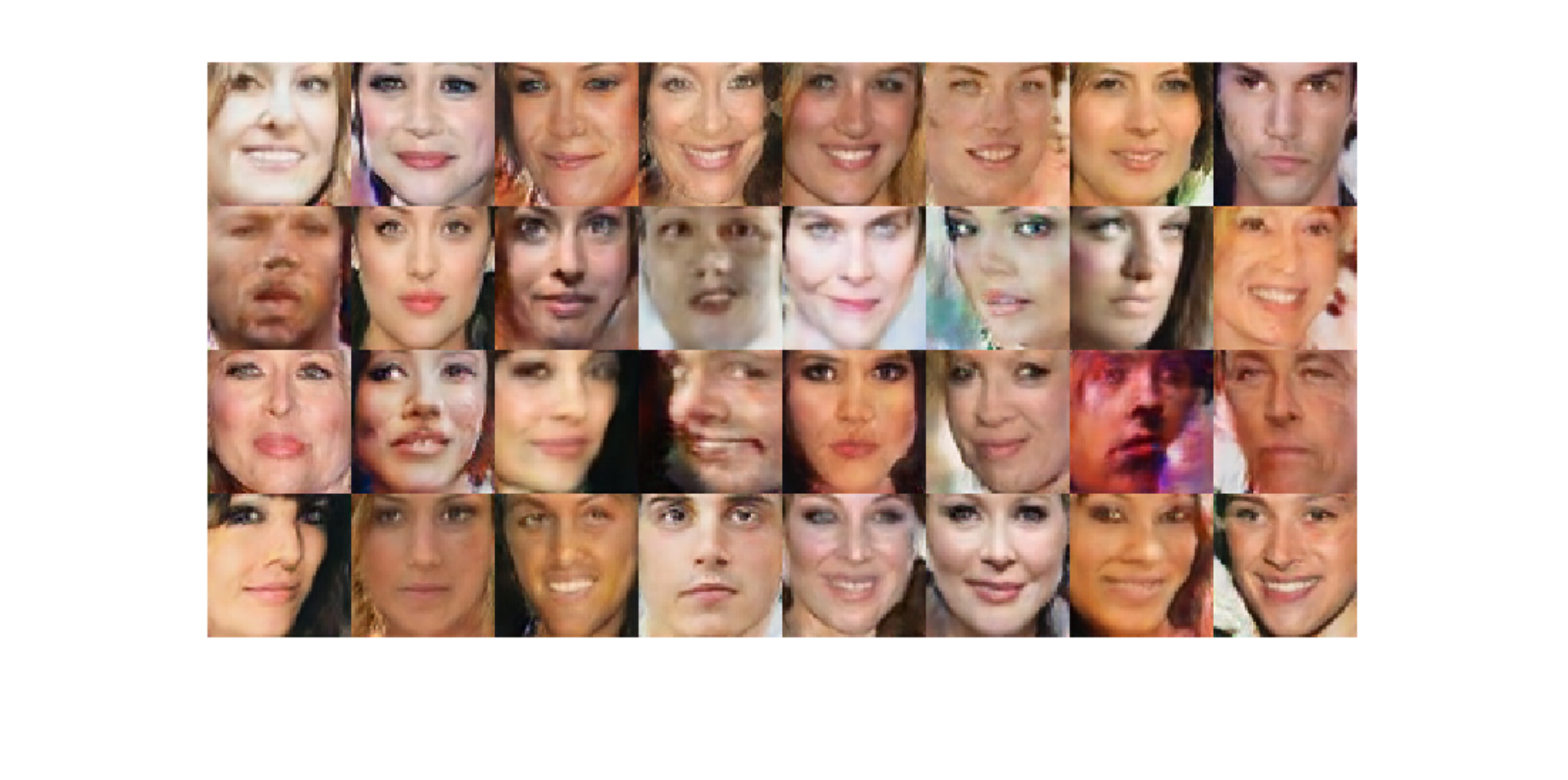}}
\caption{Visualization of face generation results with RankGAN, Wasserstein GAN (WGAN) and Least Squares GAN (LSGAN). Latent vectors $\mathbf{z}$'s are obtained by passing the input faces through the encoder $\mathcal{E}$. Since WGAN and LSGAN have not been trained with an encoder, the face identities are not preserved.}\label{fig:gen_encoder_z}
\end{figure*}

\section{Face Completion Experiments}\label{sec:supp-2}

\subsection{Experimental Setup and Results}
\subsubsection{Database}
In addition to the CelebA dataset we used in the main paper, we also collect a single-sample dataset containing 50K frontal face images from 50K individuals, which we call the SSFF dataset. They are sourced from several frontal face datasets including the FRGC v2.0 dataset \cite{phillips2005overview}, the MPIE dataset \cite{gross2010multi}, the ND-Twin dataset \cite{phillips2011distinguishing}, and mugshot dataset from Pinellas County Sheriff's Office (PCSO). Training and testing split is 9-1, with the image completion results being based only on the open-set. This dataset is single-sample, which means there is only one single image of a particular subject throughout the entire dataset. Images are aligned using two anchor points on the eyes, and cropped to $64\times64$.

\subsubsection{Occlusion Masks}

\begin{figure}[!ht]
    \centering
    \subfigure[`Center Small' mask]{\includegraphics[width=0.45\linewidth]{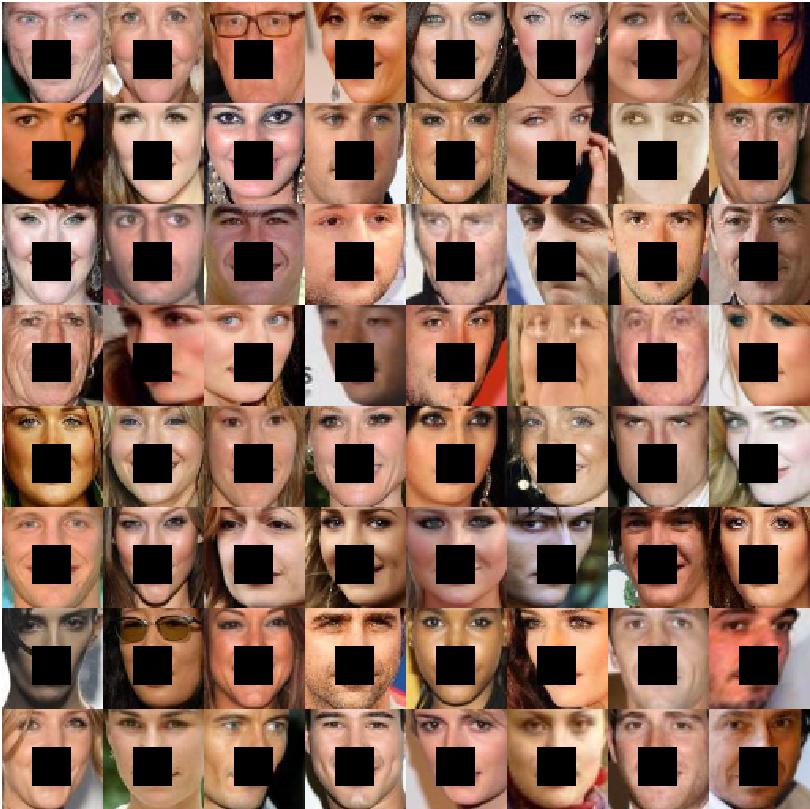}}
    \subfigure[`Center Large' mask]{\includegraphics[width=0.45\linewidth]{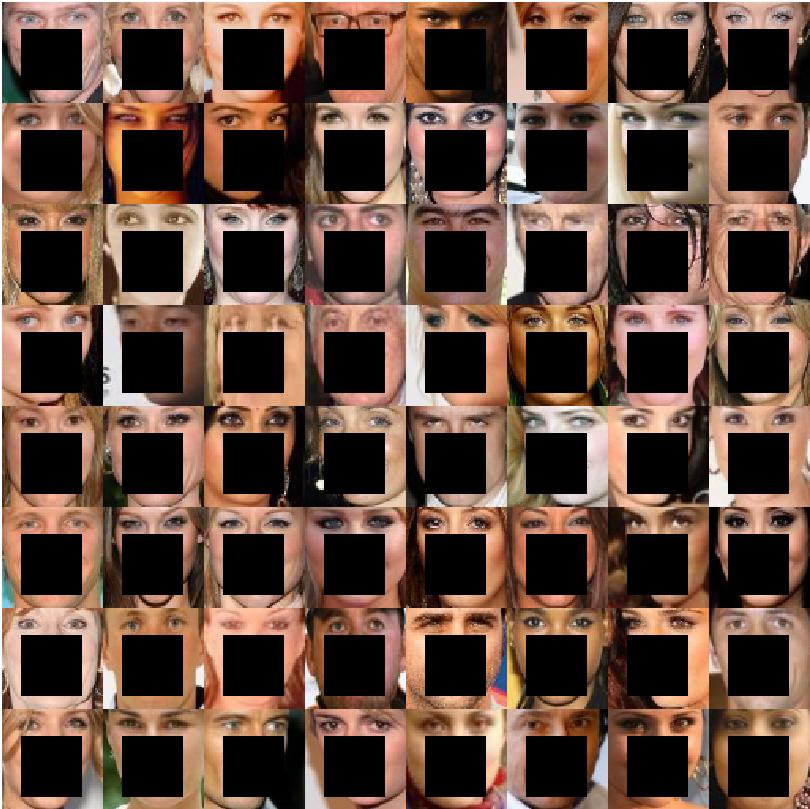}}
    \subfigure[`Periocular Large' mask]{\includegraphics[width=0.45\linewidth]{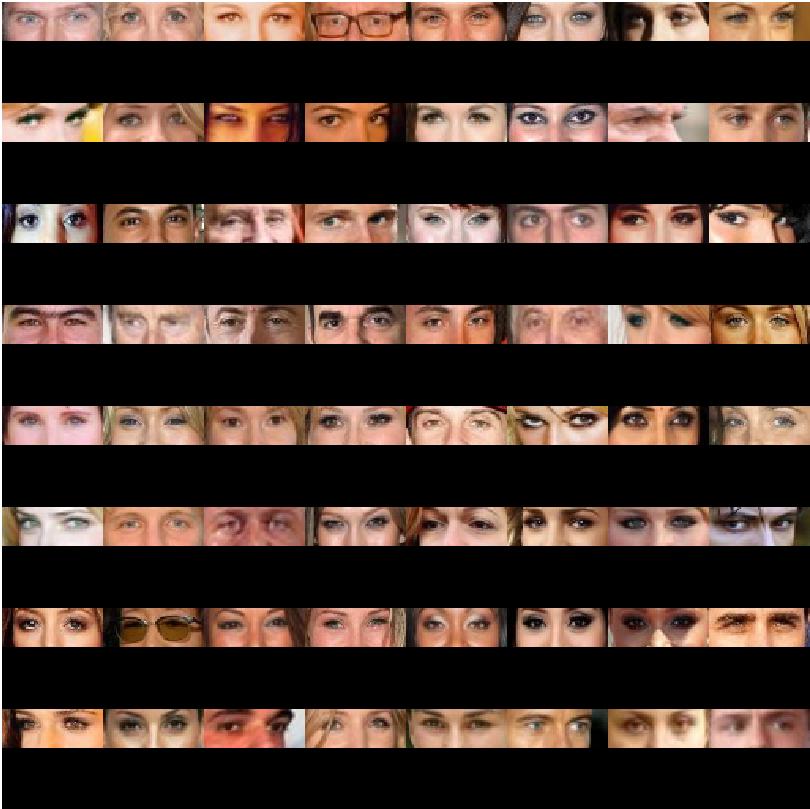}}
    \subfigure[`Periocular Small' mask]{\includegraphics[width=0.45\linewidth]{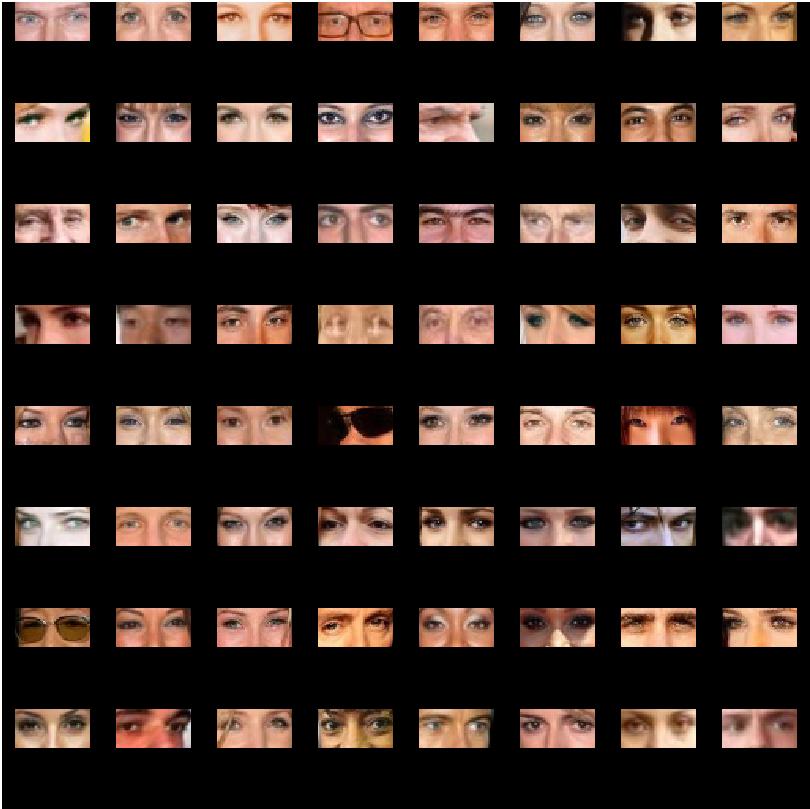}}
\caption{Four masks used in our experiments: `Center Small', `Center Large', `Periocular Large', and `Periocular Small' masks.}\label{fig:fourmasks}
\end{figure}

We carried out face completion experiments on four types of facial masks, which are termed as: `Center Small', `Center Large', `Periocular Small', and `Periocular Large'. Examples can be seen from Figure~\ref{fig:fourmasks}.

\subsubsection{Face Completion Results on Large Number of Iterations}
\begin{figure}[t]
  \centering
  \includegraphics[width=\linewidth]{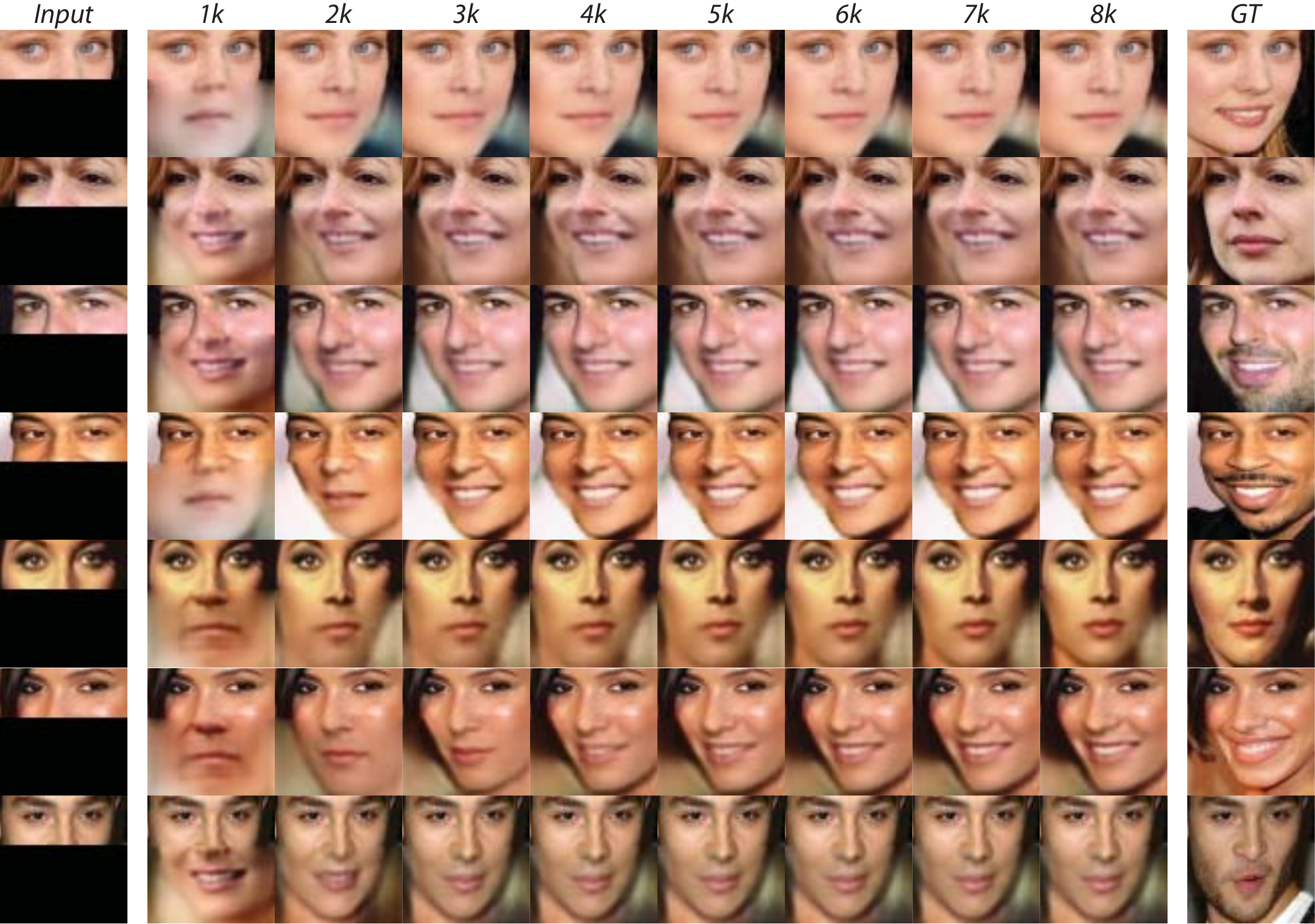}
  \caption{Image completion visual results: progression of 8000 iterations for optimizing for the $\hat{\mathbf{z}}$.}
  \label{fig:progression}
\end{figure}

Some preliminary image completion progression visual results are shown in Figure~\ref{fig:progression} with 8000 iterations for optimizing for the $\hat{\mathbf{z}}$ using Stage-3 RankGAN. As can be seen, after 8000 iterations, the RankGAN is able to achieve decent image completion results. However, such an algorithm is still slow because it requires optimization over $\hat{\mathbf{z}}$ for any query image. Therefore, for the large-scale experiments to be carried out, we limit the algorithm to optimize for only 2000 iterations for the sake of time. Results can be improved if we allow further iterations.


\subsection{Discussions}
\begin{figure*}
    \centering
    \subfigure[Original faces.]{\includegraphics[width=0.432\linewidth]{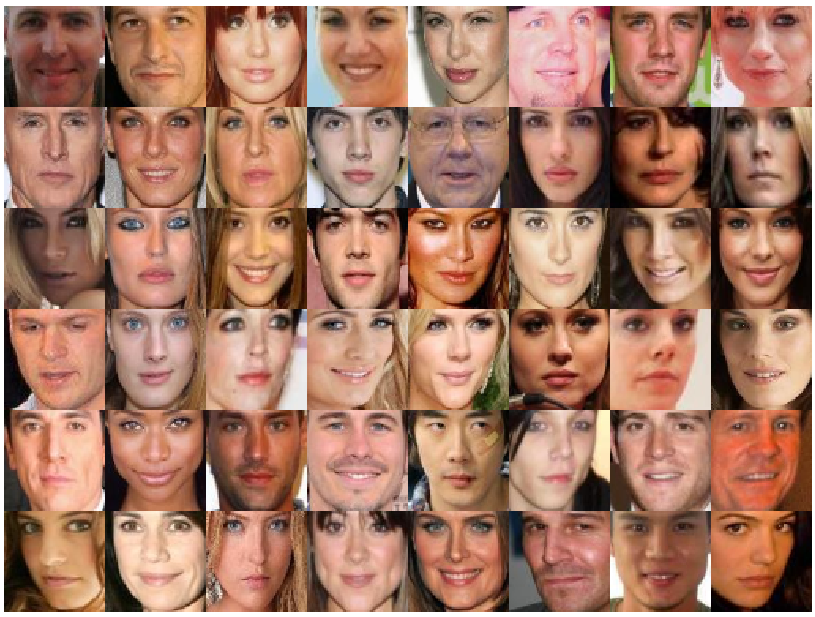}}
    \subfigure[Masked faces. (`Center Small')]{\includegraphics[width=0.432\linewidth]{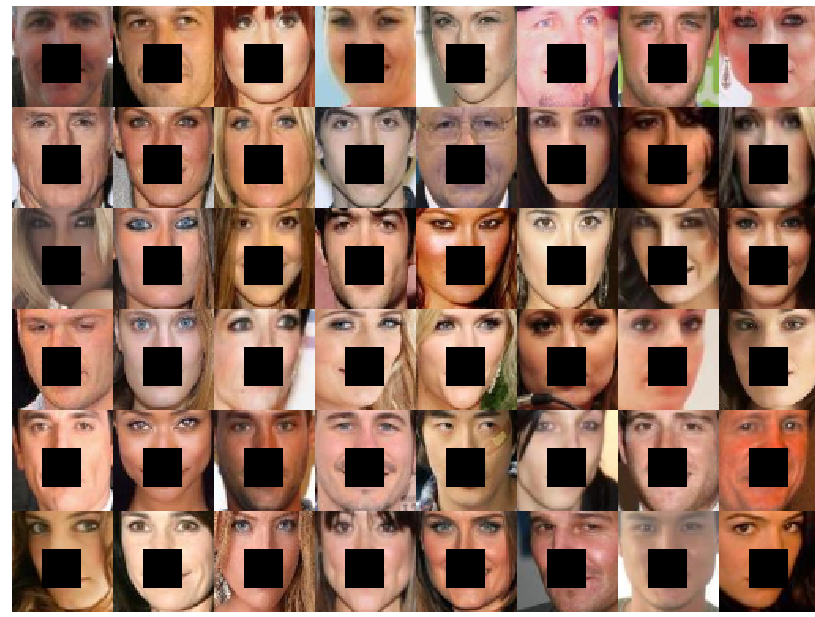}}
    \subfigure[WGAN completion, Open Set.]{\includegraphics[width=0.432\linewidth]{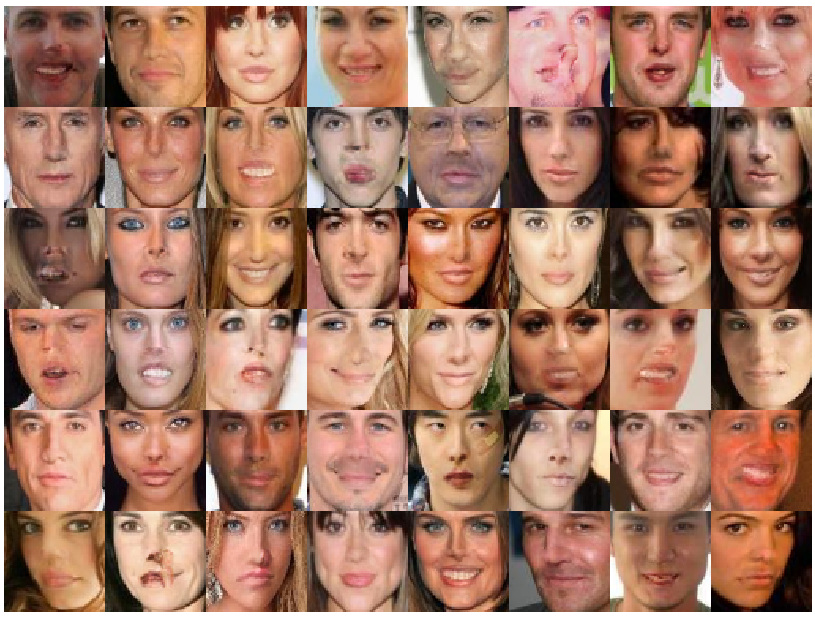}}
    \subfigure[LSGAN completion, Open Set.]{\includegraphics[width=0.432\linewidth]{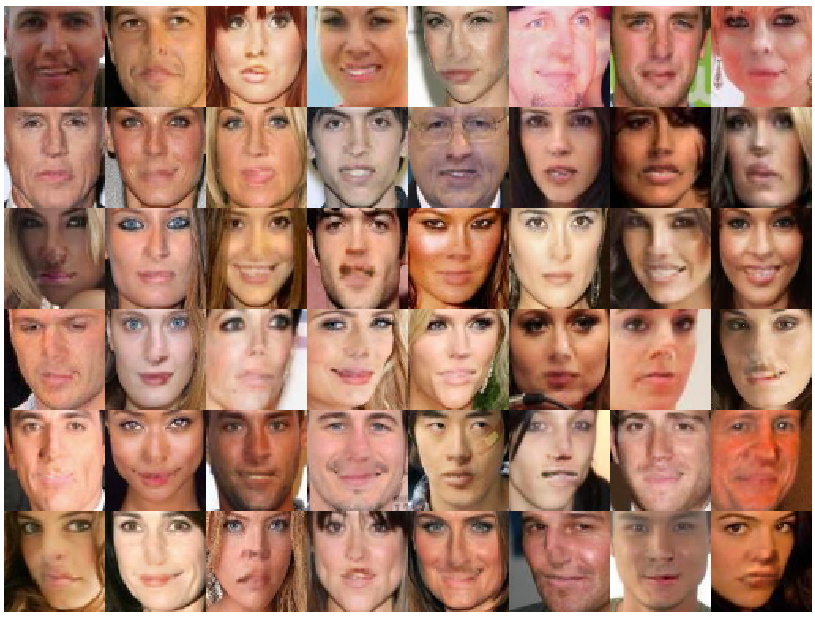}}
    \subfigure[Stage 1 completion, Open Set.]{\includegraphics[width=0.432\linewidth]{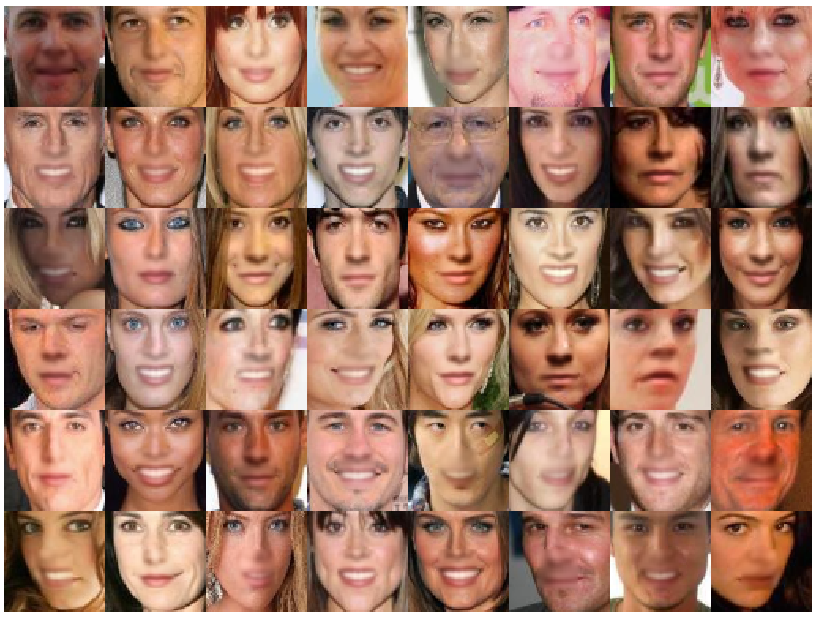}}
    \subfigure[Stage 2 completion, Open Set.]{\includegraphics[width=0.432\linewidth]{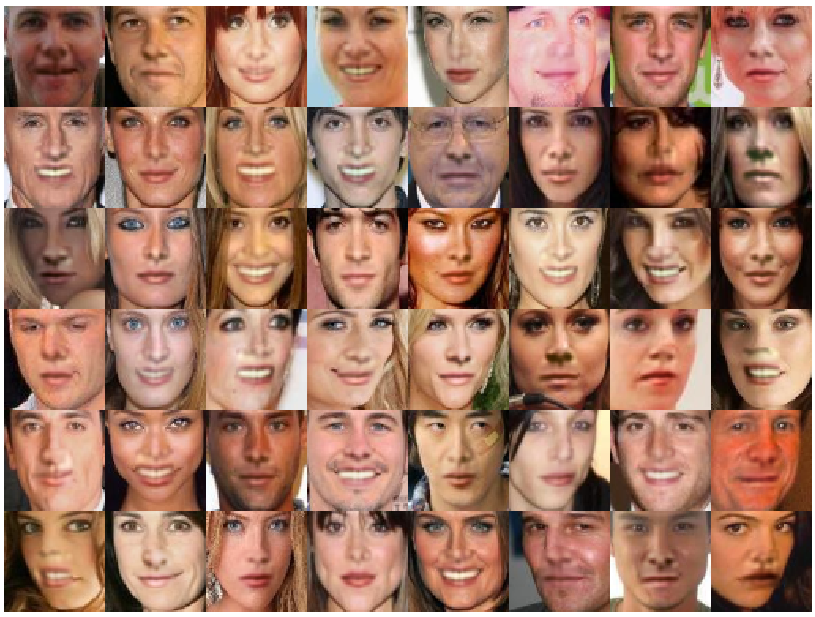}}
    \subfigure[Stage 3 completion, Open Set.]{\includegraphics[width=0.432\linewidth]{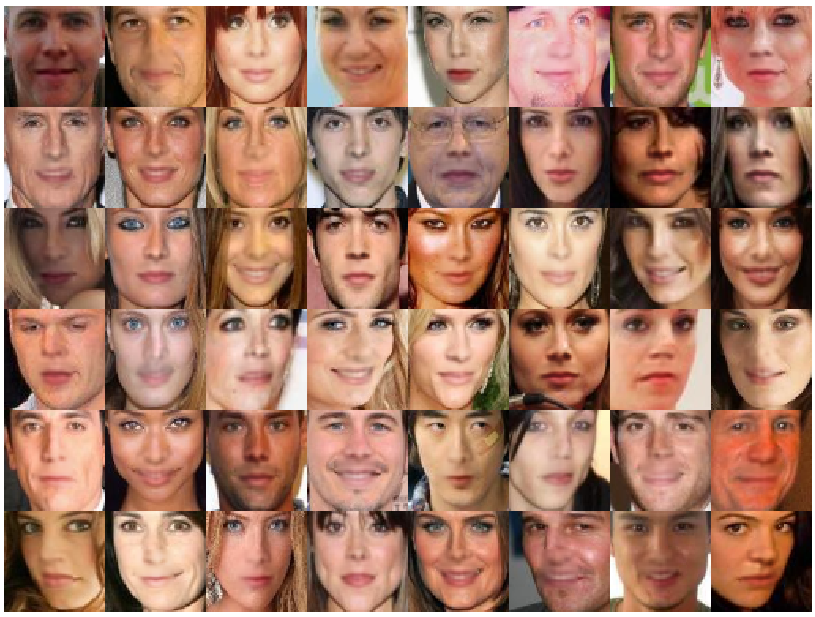}}
\caption{\textcolor{blue}{Best} completion results with RankGAN on CelebA, `Center Small' mask.}\label{fig:supp_celeb_good_center_sm}
\end{figure*}

\begin{figure*}
    \centering
    \subfigure[Original faces.]{\includegraphics[width=0.432\linewidth]{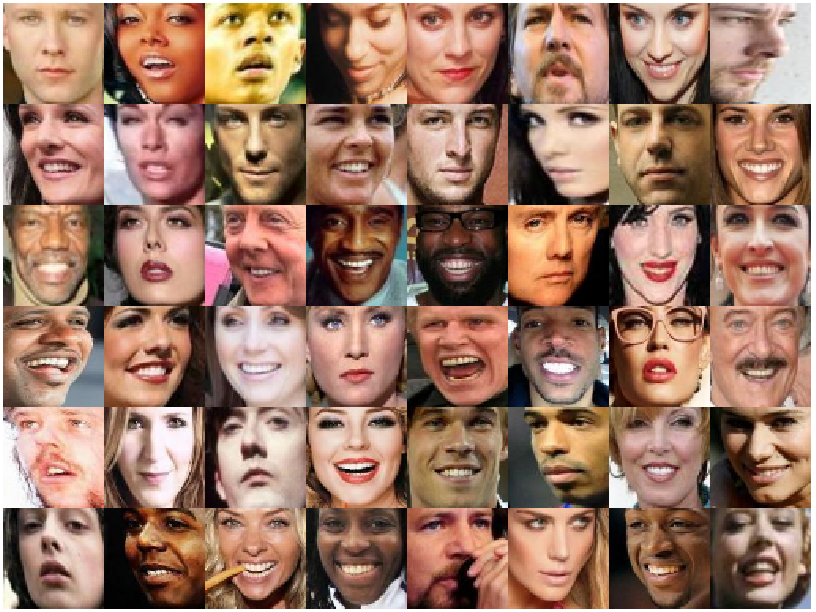}}
    \subfigure[Masked faces. (`Center Small')]{\includegraphics[width=0.432\linewidth]{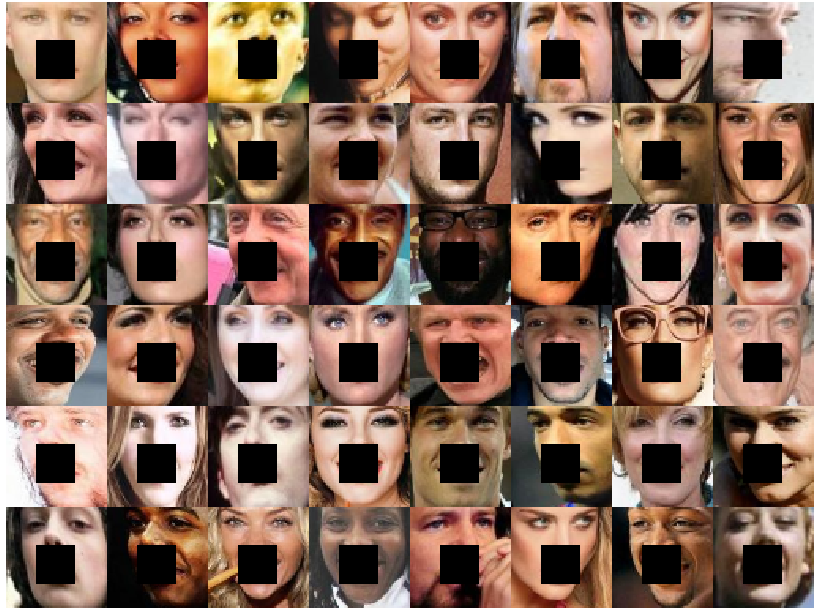}}
    \subfigure[WGAN completion, Open Set.]{\includegraphics[width=0.432\linewidth]{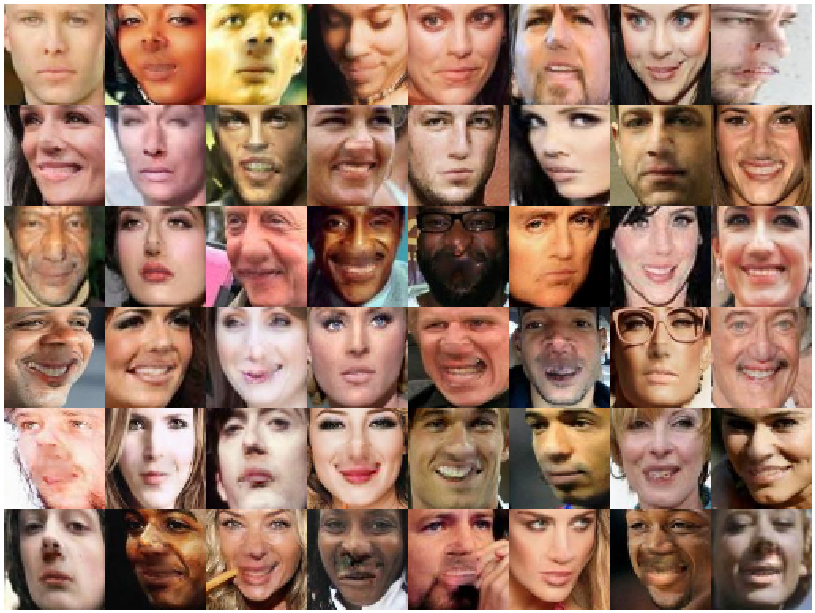}}
    \subfigure[LSGAN completion, Open Set.]{\includegraphics[width=0.432\linewidth]{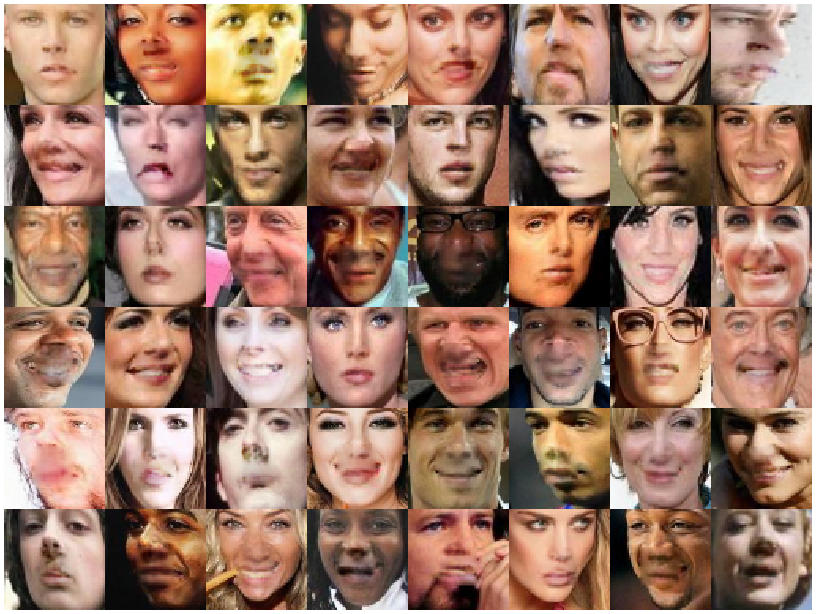}}
    \subfigure[Stage 1 completion, Open Set.]{\includegraphics[width=0.432\linewidth]{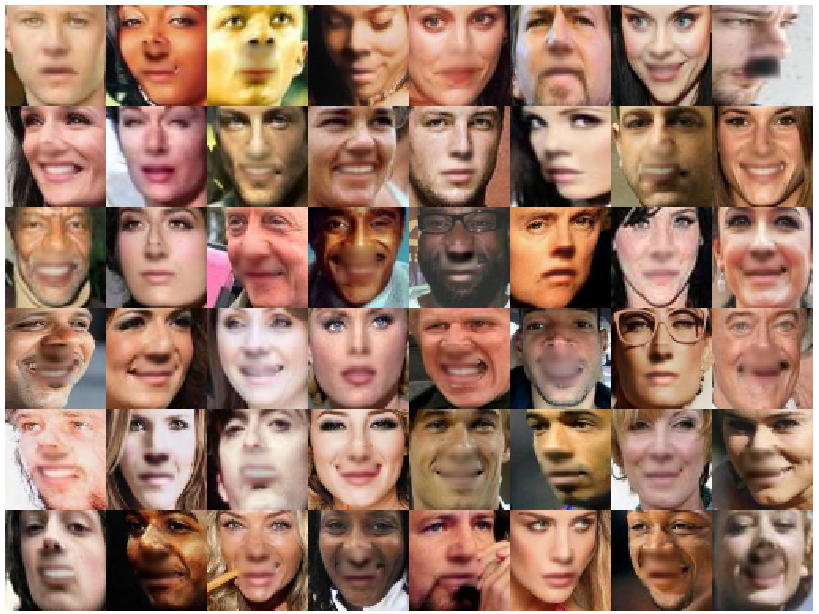}}
    \subfigure[Stage 2 completion, Open Set.]{\includegraphics[width=0.432\linewidth]{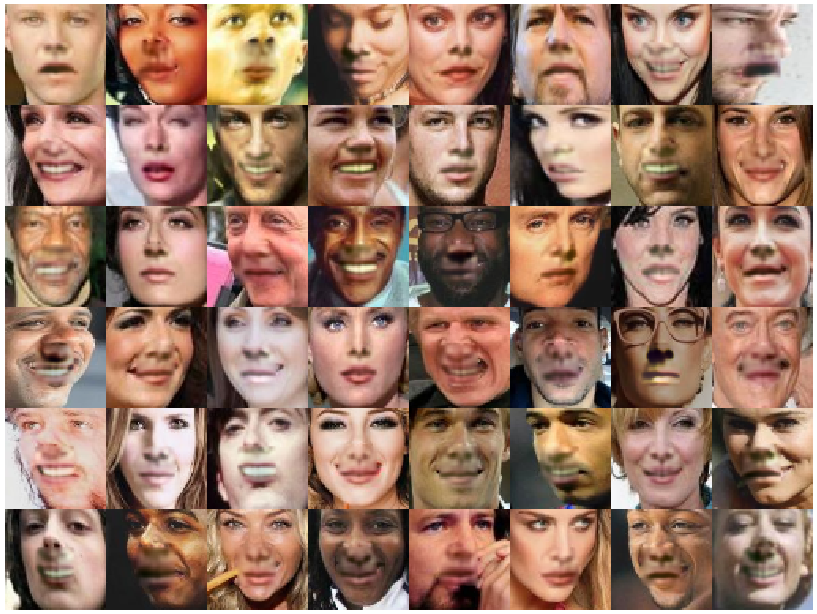}}
    \subfigure[Stage 3 completion, Open Set.]{\includegraphics[width=0.432\linewidth]{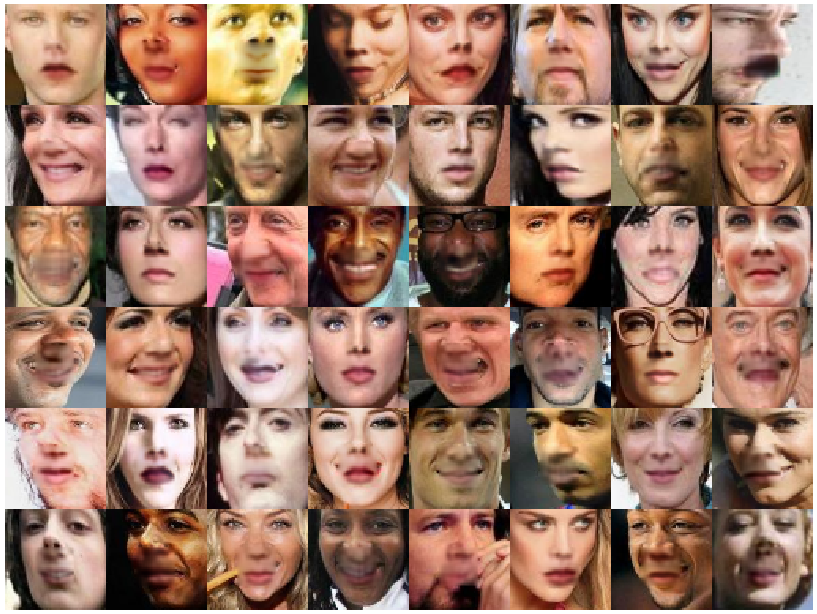}}
\caption{\textcolor{red}{Worst} completion results with RankGAN on CelebA, `Center Small' mask.}\label{fig:supp_celeb_bad_center_sm}
\end{figure*}

\begin{figure*}
    \centering
    \subfigure[Original faces.]{\includegraphics[width=0.432\linewidth]{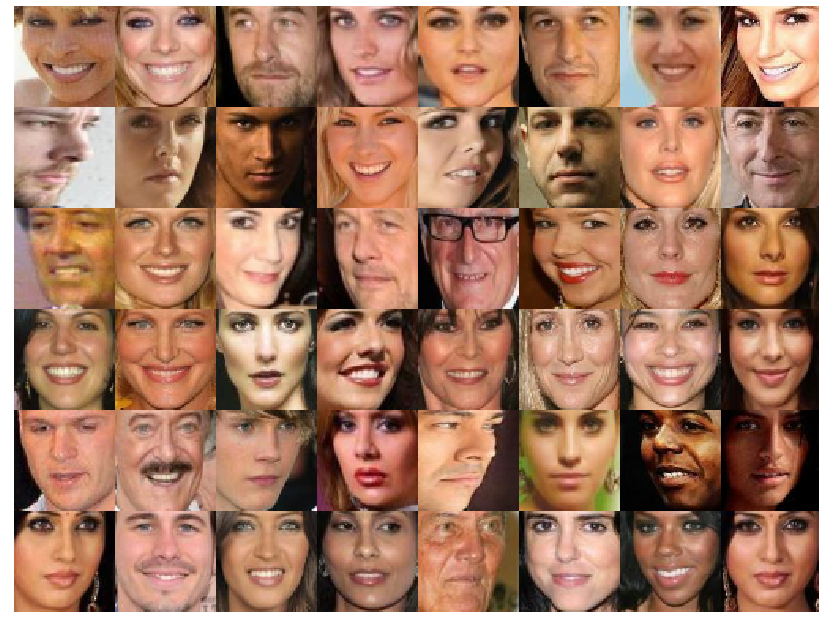}}
    \subfigure[Masked faces. (`Periocular Large')]{\includegraphics[width=0.432\linewidth]{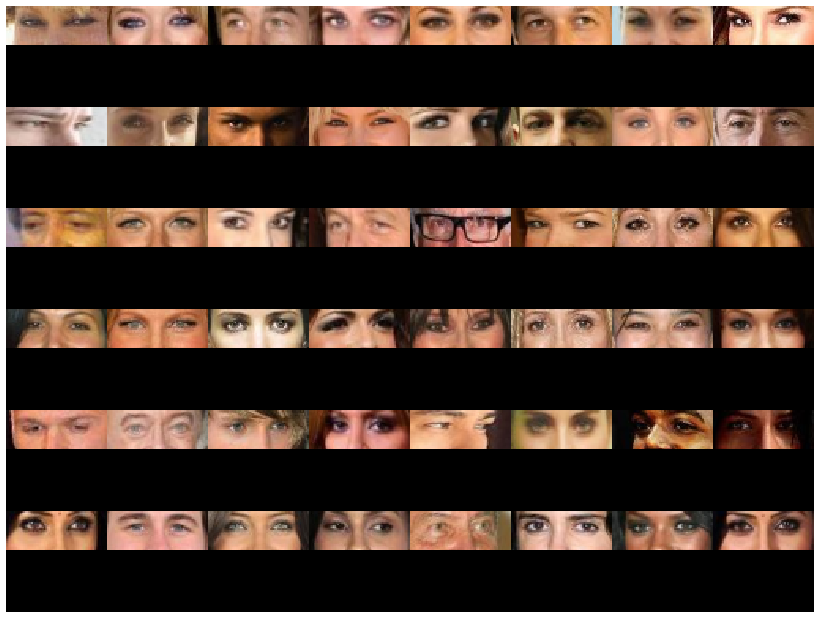}}
    \subfigure[WGAN completion, Open Set.]{\includegraphics[width=0.432\linewidth]{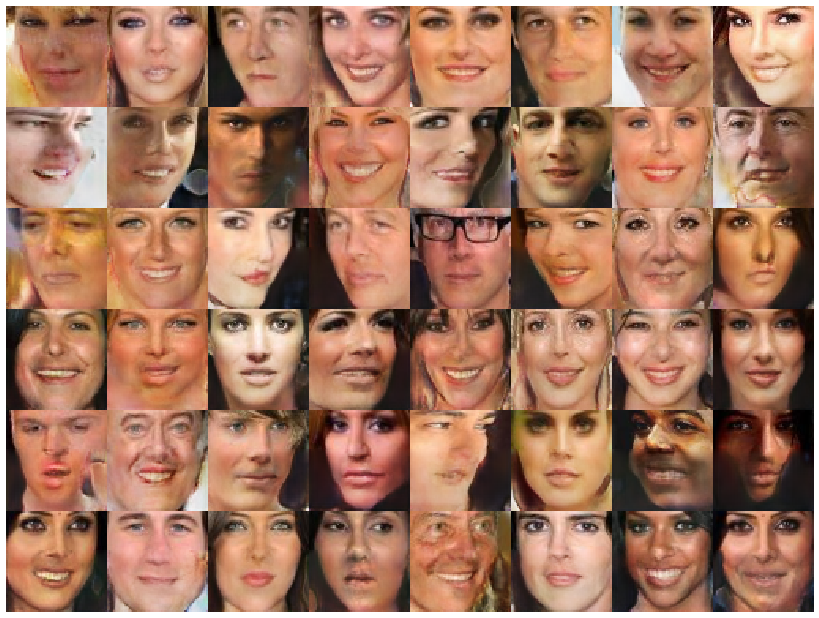}}
    \subfigure[LSGAN completion, Open Set.]{\includegraphics[width=0.432\linewidth]{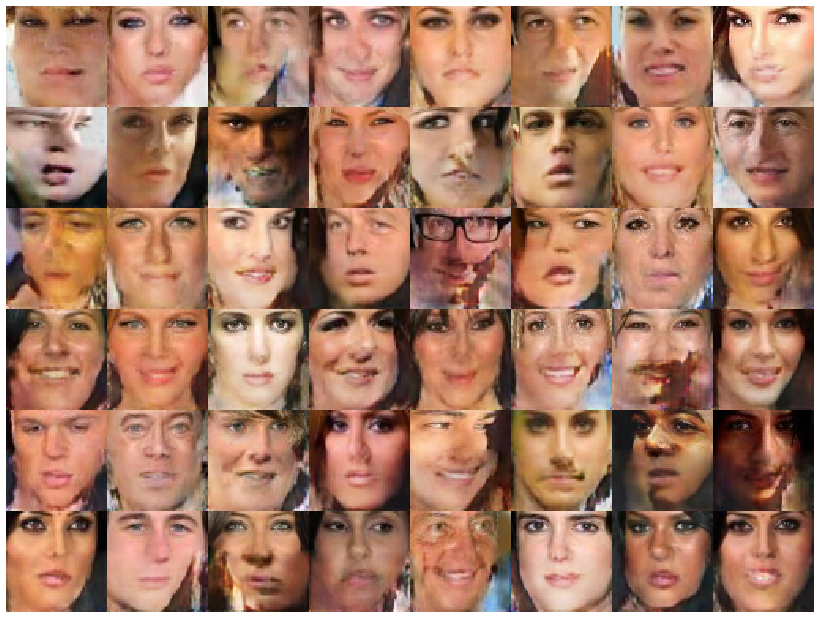}}
    \subfigure[Stage 1 completion, Open Set.]{\includegraphics[width=0.432\linewidth]{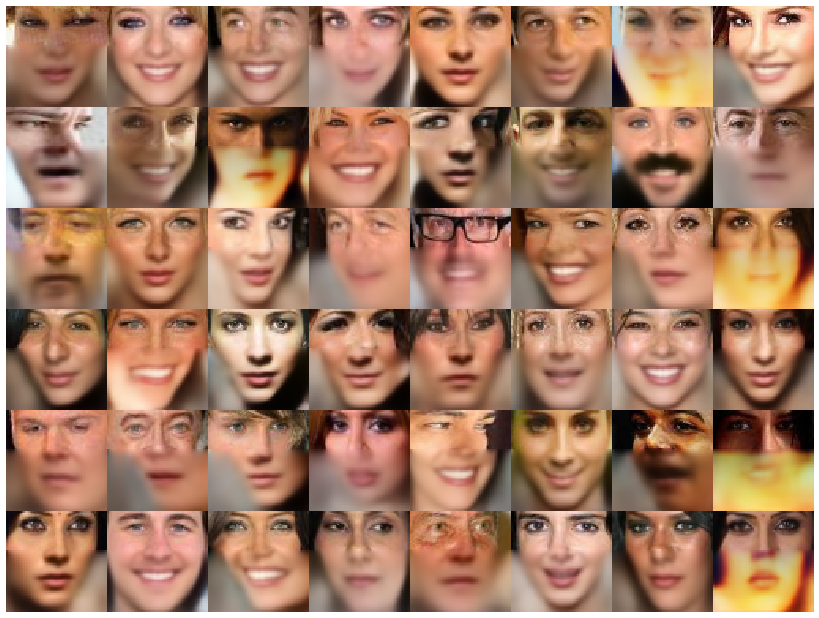}}
    \subfigure[Stage 2 completion, Open Set.]{\includegraphics[width=0.432\linewidth]{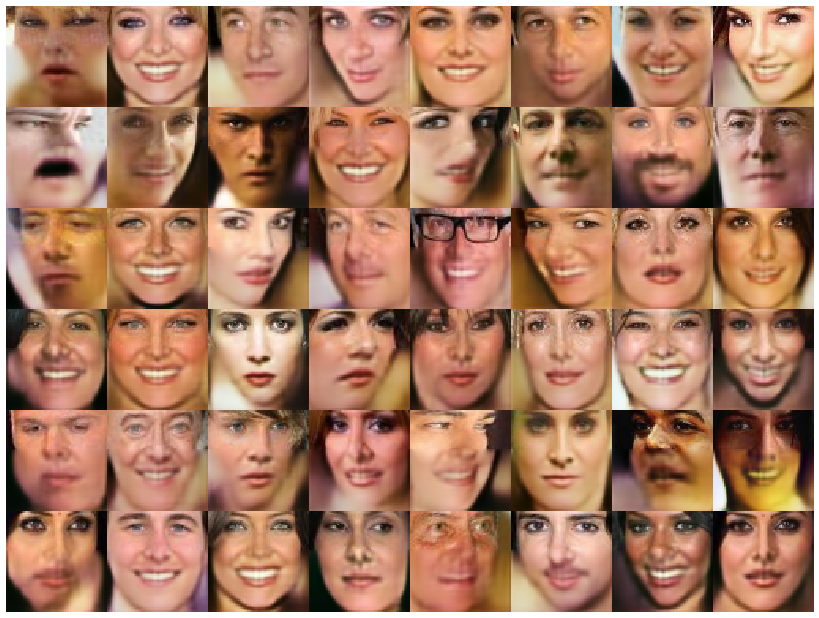}}
    \subfigure[Stage 3 completion, Open Set.]{\includegraphics[width=0.432\linewidth]{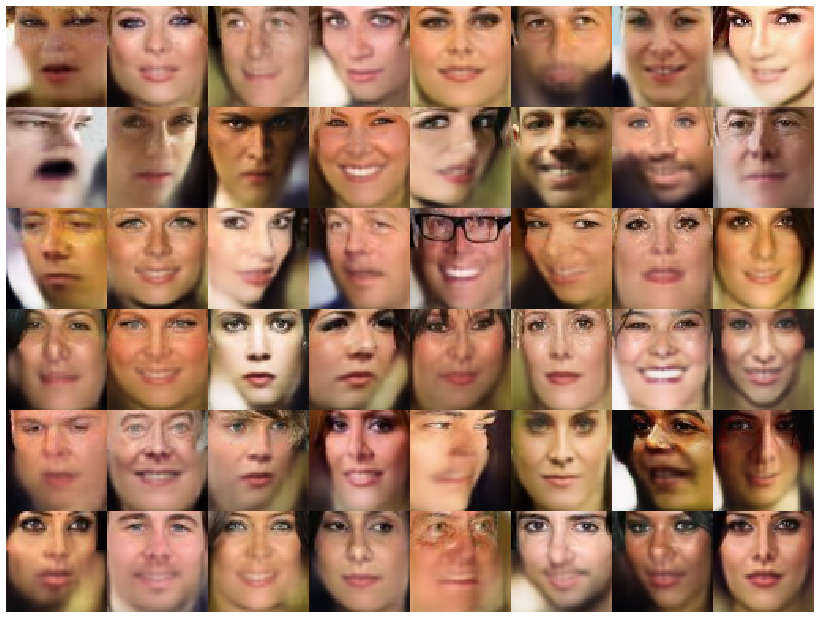}}
\caption{\textcolor{blue}{Best} completion results with RankGAN on CelebA, `Periocular Large' mask.}\label{fig:supp_celeb_good_peri_lg}
\end{figure*}

\begin{figure*}
    \centering
    \subfigure[Original faces.]{\includegraphics[width=0.432\linewidth]{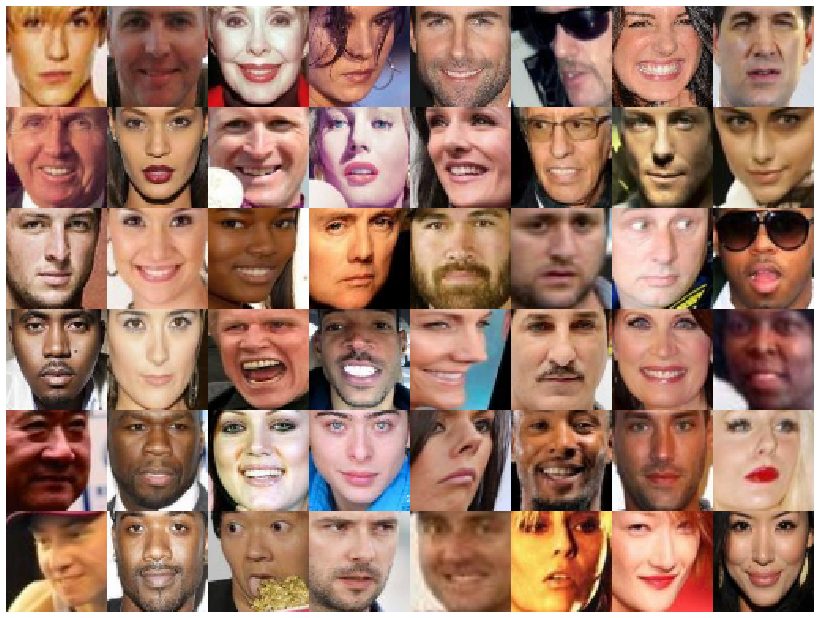}}
    \subfigure[Masked faces. (`Periocular Large')]{\includegraphics[width=0.432\linewidth]{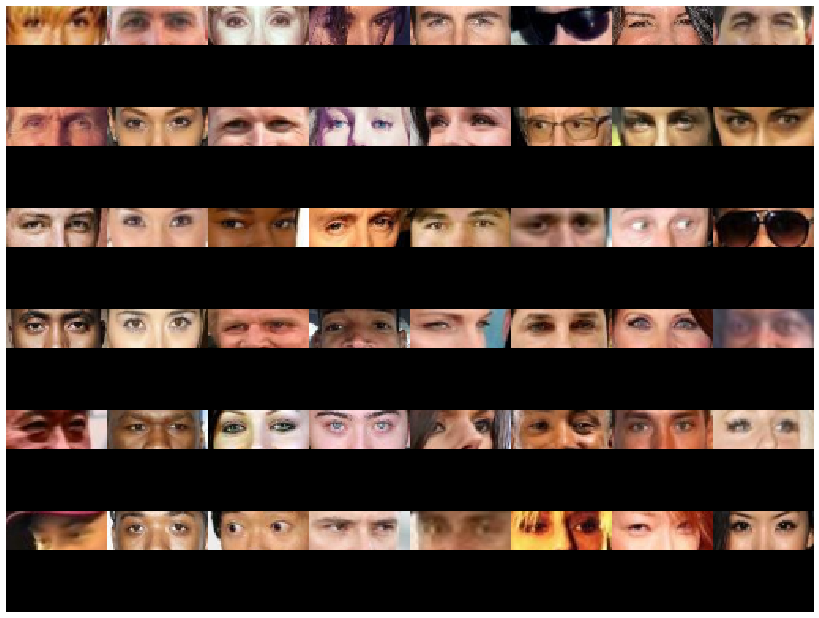}}
    \subfigure[WGAN completion, Open Set.]{\includegraphics[width=0.432\linewidth]{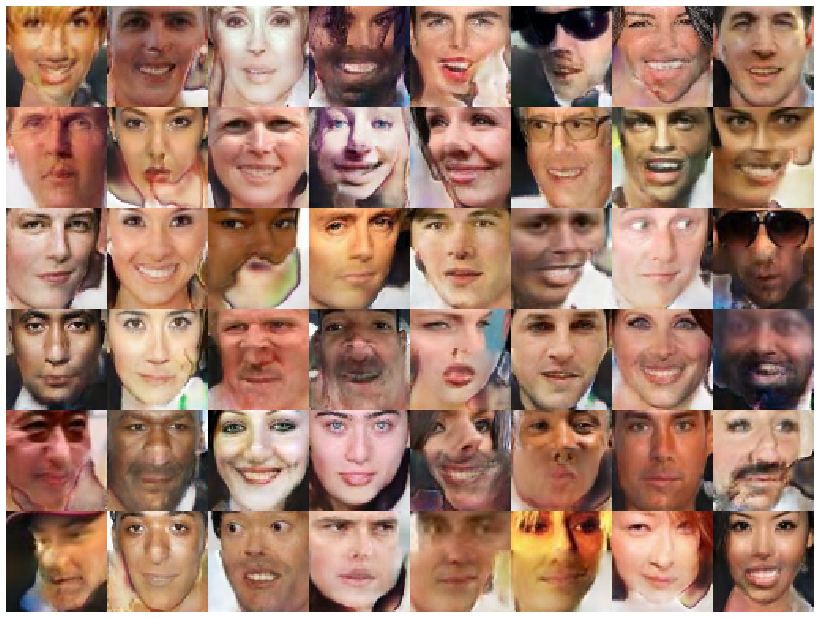}}
    \subfigure[LSGAN completion, Open Set.]{\includegraphics[width=0.432\linewidth]{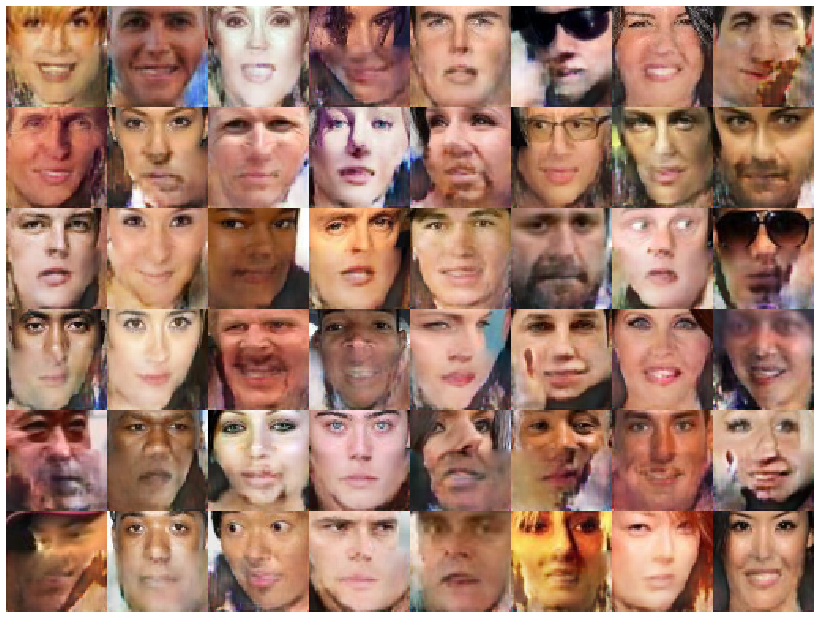}}
    \subfigure[Stage 1 completion, Open Set.]{\includegraphics[width=0.432\linewidth]{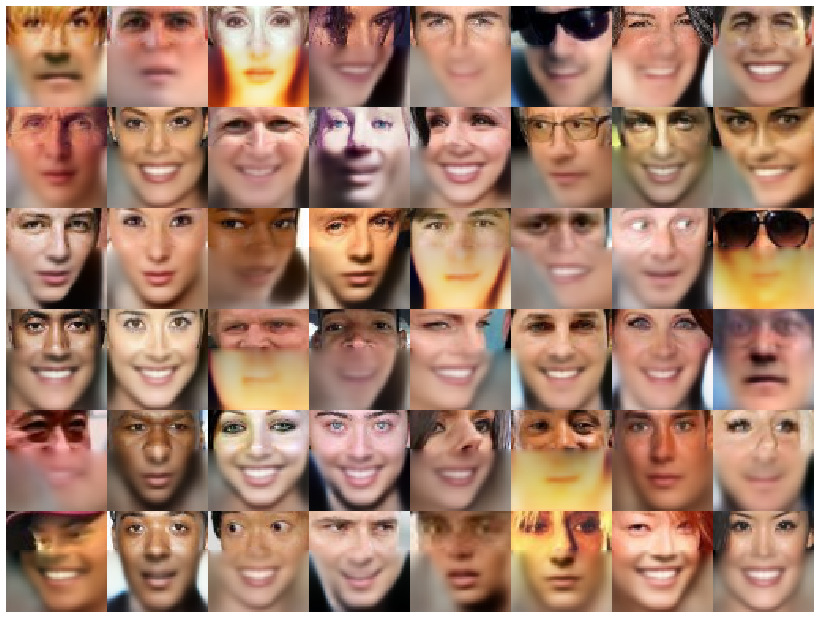}}
    \subfigure[Stage 2 completion, Open Set.]{\includegraphics[width=0.432\linewidth]{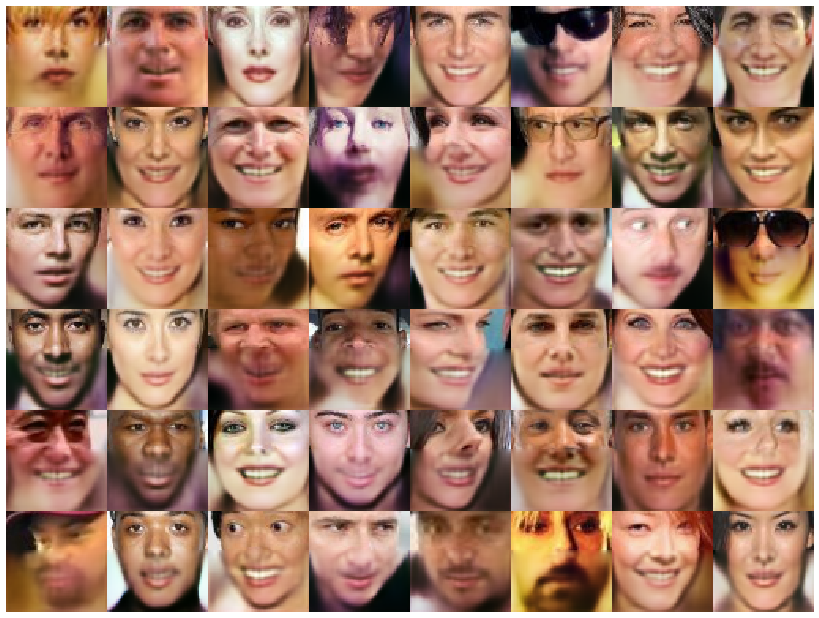}}
    \subfigure[Stage 3 completion, Open Set.]{\includegraphics[width=0.432\linewidth]{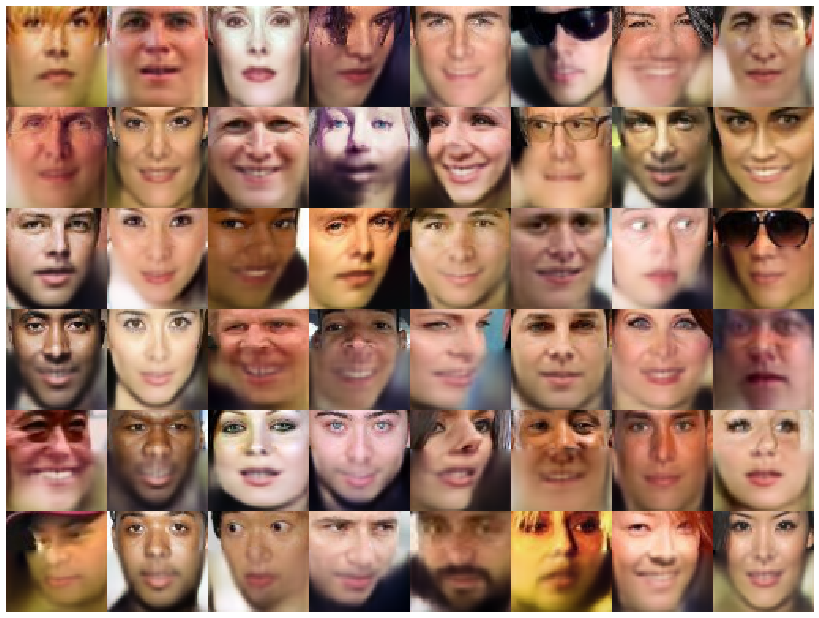}}
\caption{\textcolor{red}{Worst} completion results with RankGAN on CelebA, `Periocular Large' mask.}\label{fig:supp_celeb_bad_peri_lg}
\end{figure*}

\begin{figure*}
    \centering
    \subfigure[Original faces.]{\includegraphics[width=0.432\linewidth]{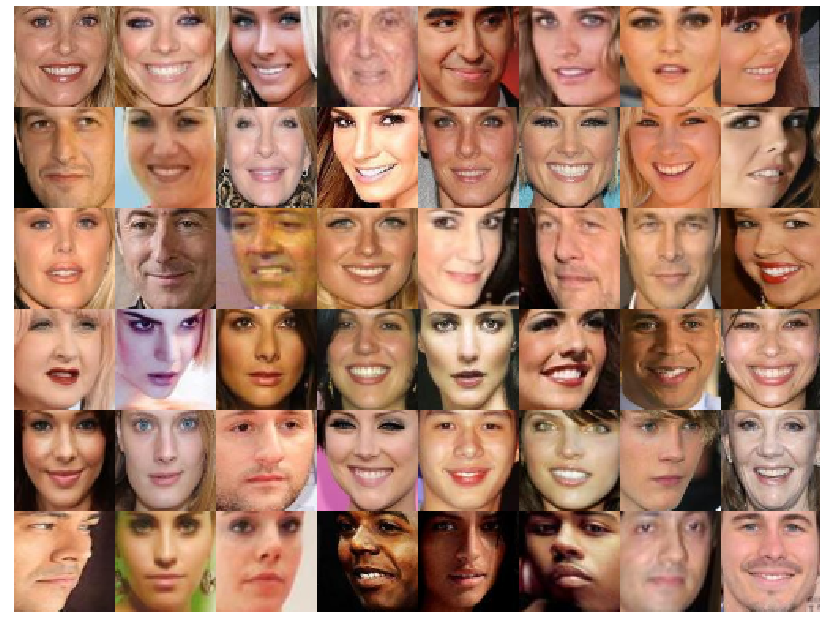}}
    \subfigure[Masked faces. (`Periocular Small')]{\includegraphics[width=0.432\linewidth]{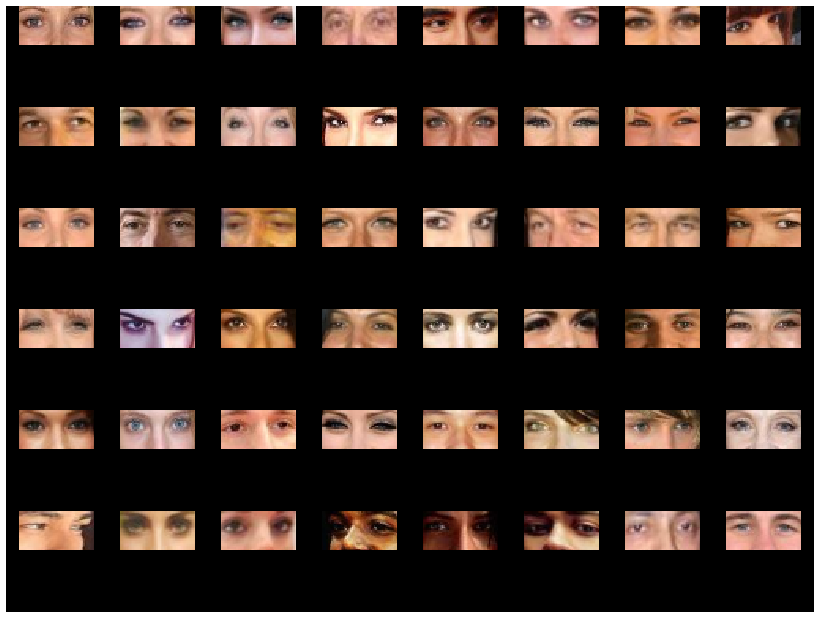}}
    \subfigure[WGAN completion, Open Set.]{\includegraphics[width=0.432\linewidth]{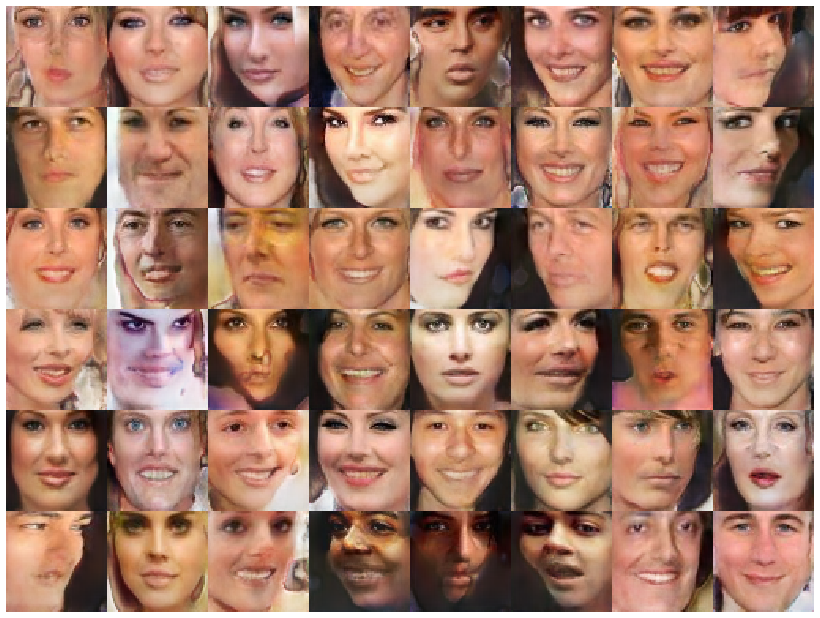}}
    \subfigure[LSGAN completion, Open Set.]{\includegraphics[width=0.432\linewidth]{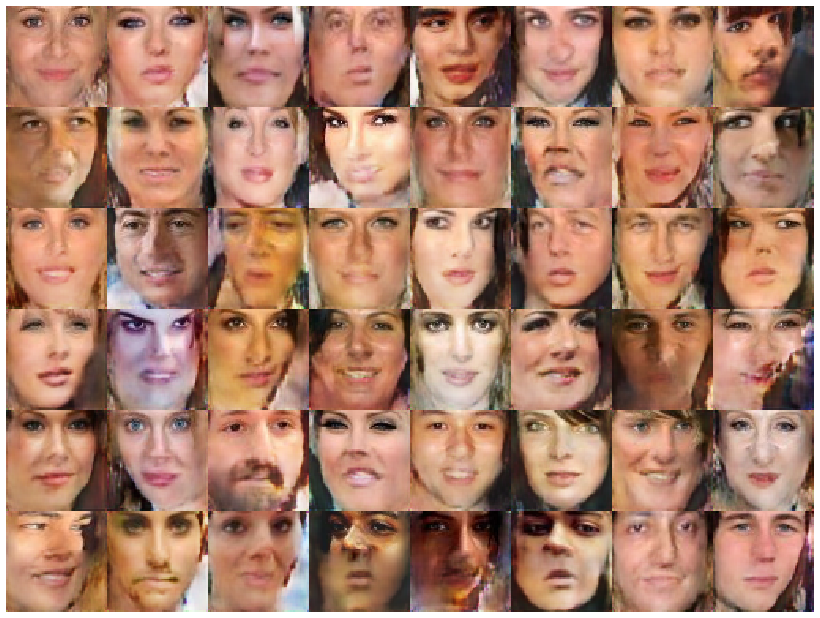}}
    \subfigure[Stage 1 completion, Open Set.]{\includegraphics[width=0.432\linewidth]{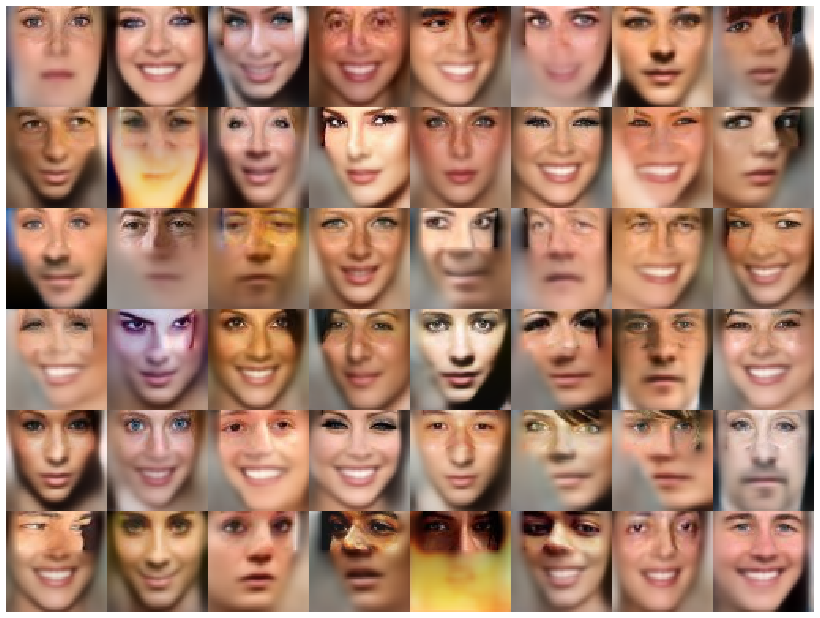}}
    \subfigure[Stage 2 completion, Open Set.]{\includegraphics[width=0.432\linewidth]{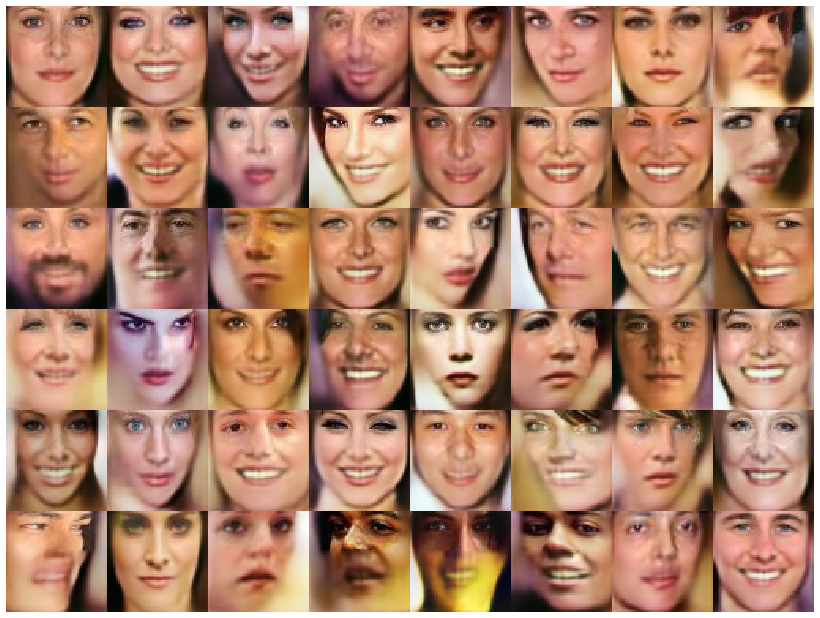}}
    \subfigure[Stage 3 completion, Open Set.]{\includegraphics[width=0.432\linewidth]{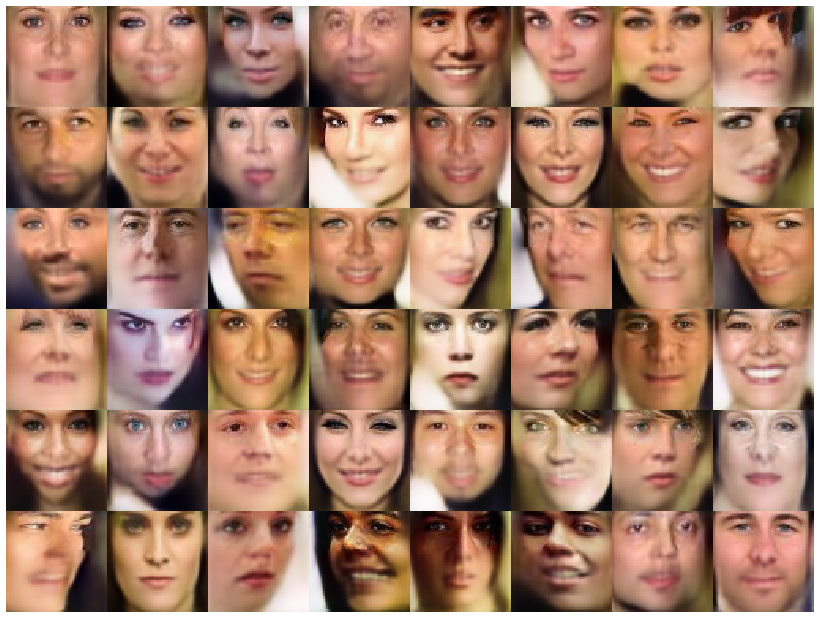}}
\caption{\textcolor{blue}{Best} completion results with RankGAN on CelebA, `Periocular Small' mask.}\label{fig:supp_celeb_good_peri_sm}
\end{figure*}

\begin{figure*}
    \centering
    \subfigure[Original faces.]{\includegraphics[width=0.432\linewidth]{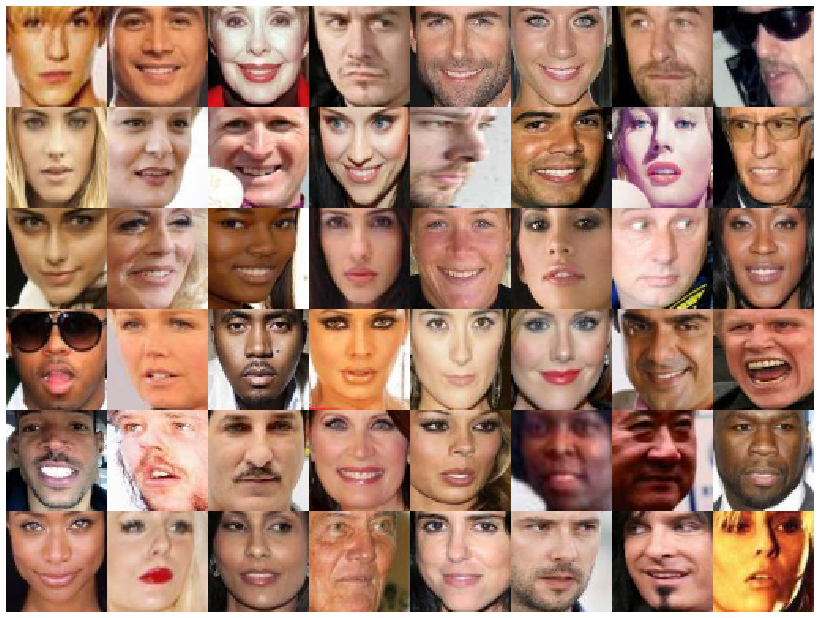}}
    \subfigure[Masked faces. (`Periocular Small')]{\includegraphics[width=0.432\linewidth]{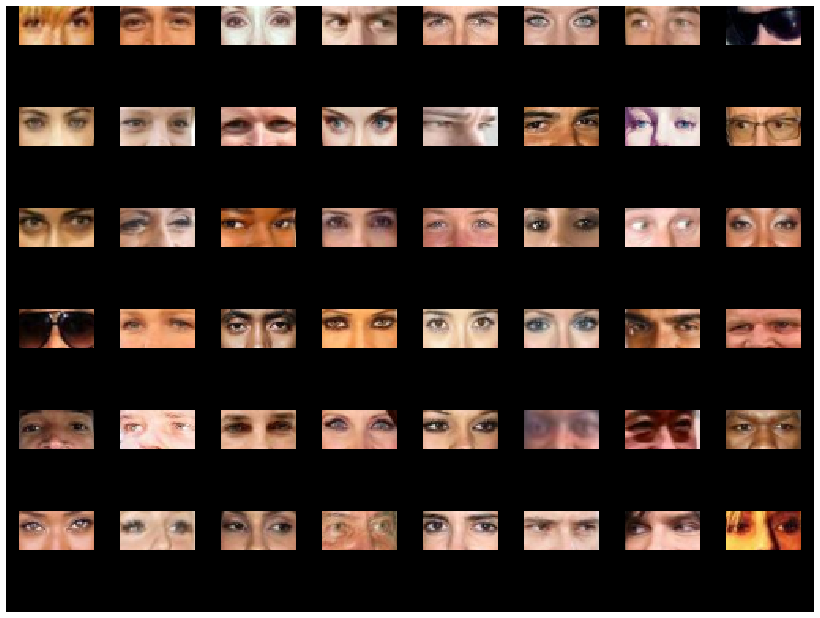}}
    \subfigure[WGAN completion, Open Set.]{\includegraphics[width=0.432\linewidth]{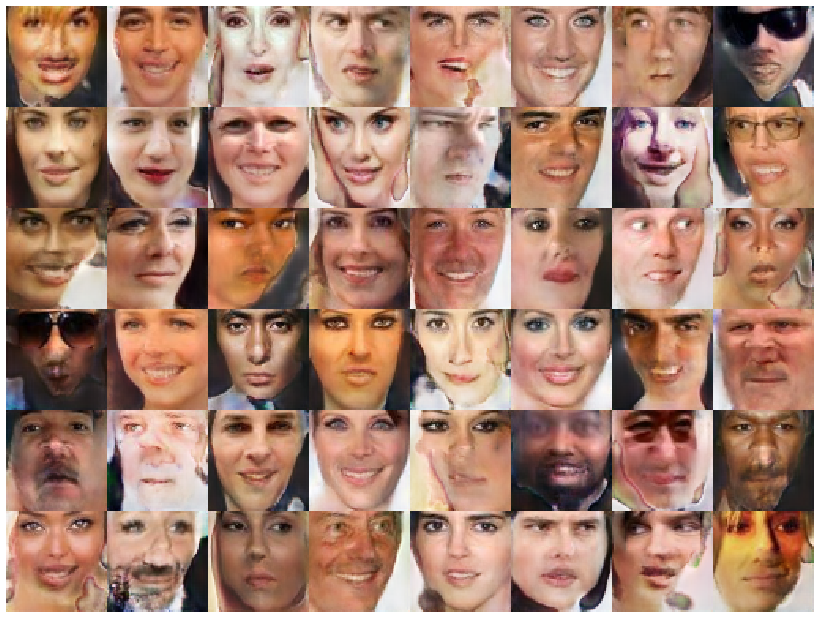}}
    \subfigure[LSGAN completion, Open Set.]{\includegraphics[width=0.432\linewidth]{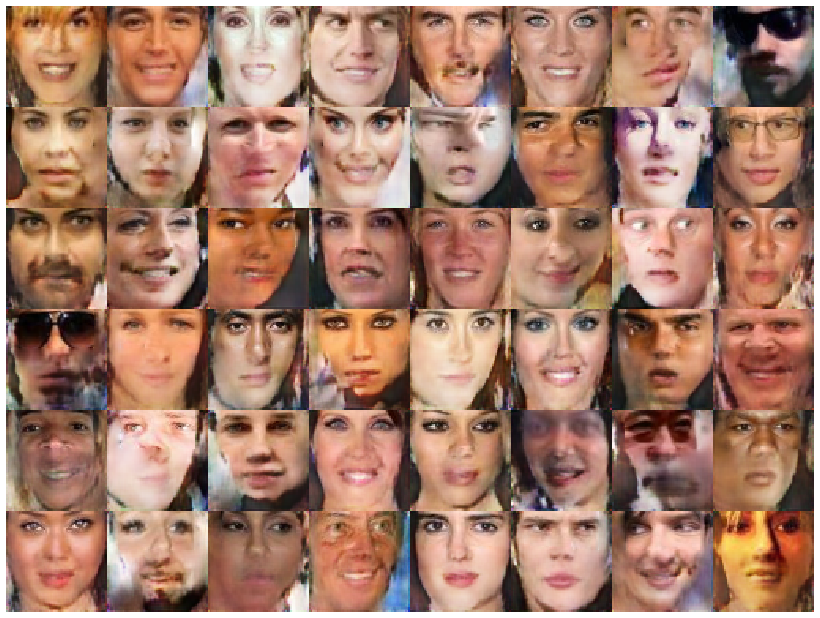}}
    \subfigure[Stage 1 completion, Open Set.]{\includegraphics[width=0.432\linewidth]{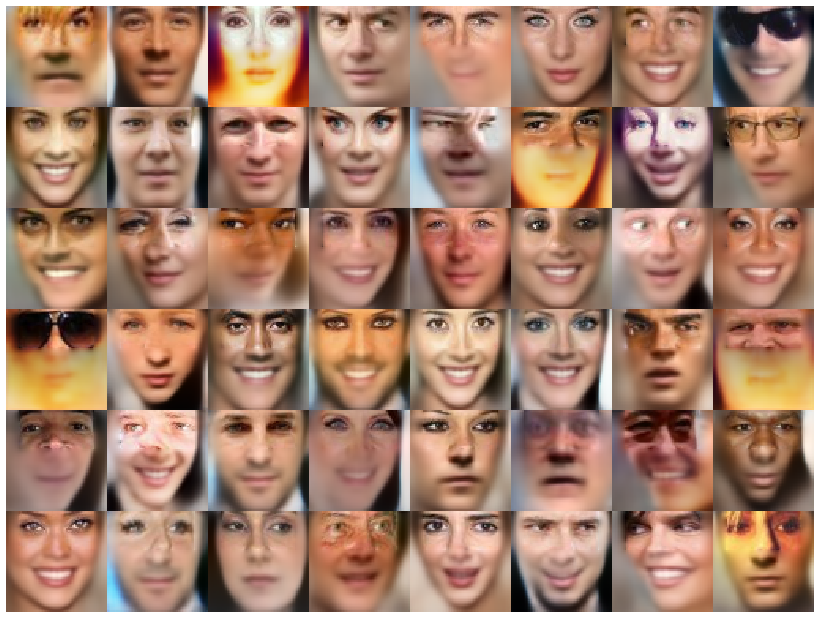}}
    \subfigure[Stage 2 completion, Open Set.]{\includegraphics[width=0.432\linewidth]{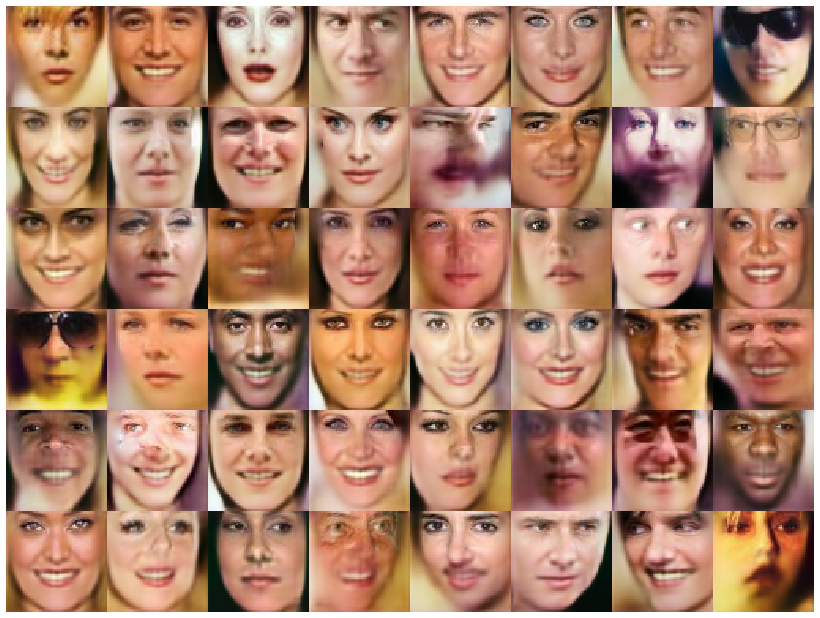}}
    \subfigure[Stage 3 completion, Open Set.]{\includegraphics[width=0.432\linewidth]{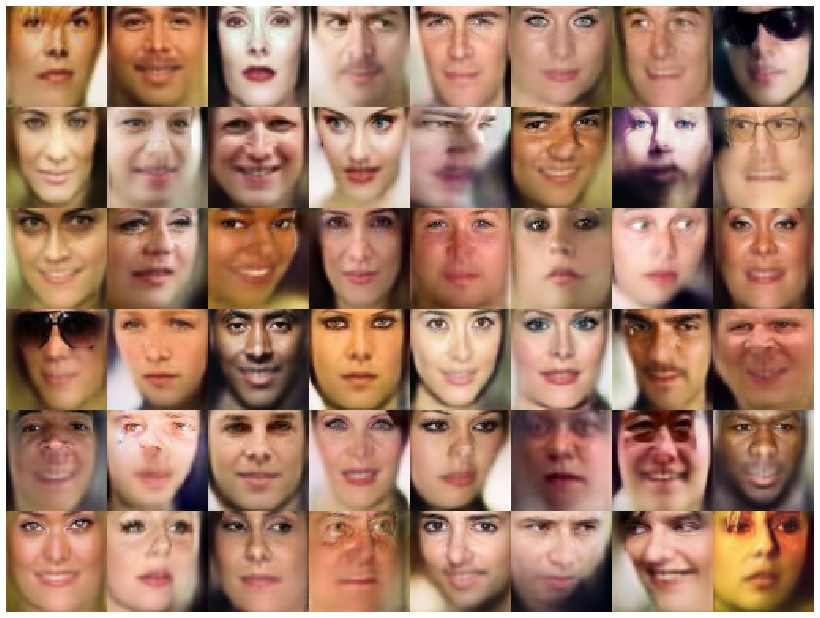}}
\caption{\textcolor{red}{Worst} completion results with RankGAN on CelebA, `Periocular Small' mask.}\label{fig:supp_celeb_bad_peri_sm}
\end{figure*}

As discussed above, we used \emph{Face Completion} task to quantify the performance of our proposed approach. We used four different mask types to perform image completion and showed quantitatively that reconstruction improves with each stage of RankGAN. The four masks represent varying difficulties of image inpainting depending on the amount and type of visible image region. We use a large and a small square mask to occlude the central facial region. Another pair of large and small rectangular masks were used to make visible only the periocular region of the face image. In each of the above cases, we measured the performance of different stages of RankGAN using several metrics that have been detailed in Tables~\ref{tab:completion_celeb_centerSm} to \ref{tab:completion_ssff_perioSm}. In each case, we showed 48 best and worst examples of image inpainting from the open set in Figures~\ref{fig:supp_celeb_good_center_sm} to \ref{fig:supp_celeb_bad_peri_sm}. Note that, the SSFF dataset consists of real mugshot faces and hence for privacy concerns, we only report numerical metrics for the same.

\begin{table}[!ht]
\centering
\caption{Data: CelebA, Mask: Center Large}
\begin{tabular}{|c|c|c|c|c|c|c|}
\hline
  ~       & FID     & Inception & PSNR & SSIM & OpenFace (AUC) & PittPatt (AUC)\\\hline
  Original& N/A & 2.3286 & N/A & N/A        & 1.0000 (0.9965) & 19.6092 (0.9109) \\ \hline
  Stage-1 & 27.09 & 2.1524 & 22.76 & 0.7405 & 0.6726 (0.9724) & 10.2502 (0.7134) \\
  Stage-2 & 23.69 & 2.1949 & 21.87 & 0.7267 & 0.6771 (0.9573) & 9.9718 (0.8214)  \\
  Stage-3 & 27.31 & \textbf{2.2846} & \textbf{23.30} & \textbf{0.7493} & \textbf{0.6789 (0.9749)} & \textbf{10.4102 (0.7922)} \\ \hline
  WGAN    & \textbf{17.03} & 2.2771 & 23.26 & 0.7362 & 0.5554 (0.9156) & 8.1031 (0.7373) \\ \hline
  LSGAN   & 23.93 & 2.2636 & 23.11 & 0.7361 & 0.6676 (0.9659) &	10.1482 (0.7154) \\
  \hline
\end{tabular}
\label{tab:completion_celeb_centerLg}
\end{table}

\begin{table}[!ht]
\centering
\caption{Data: CelebA, Mask: Center Small}
\begin{tabular}{|c|c|c|c|c|c|c|}
  \hline
  ~       & FID     & Inception & PSNR & SSIM & OpenFace (AUC) & PittPatt (AUC)\\\hline
  Original& N/A &  2.2697 & N/A & N/A        & 1.0000 (0.9916) & 19.6108 (0.9135) \\ \hline
  Stage-1 & 16.54 &  2.1331 & 26.64 & 0.8885 & 0.8246 (0.9913) & 12.8730 (0.8201) \\
  Stage-2 & 15.35 & 2.1898 & 25.68 & 0.8808 & 0.8331 (0.9946) & \textbf{13.0946 (0.8601)} \\
  Stage-3 & 15.45 & 2.2903 & 26.65 & \textbf{0.8888} & \textbf{0.8399 (0.9897)} & 13.0318 (0.8260) \\ \hline
  WGAN    & \textbf{12.67} & 2.2498 & \textbf{26.69} & 0.8834 & 0.7650 (0.9956) & 11.3301 (0.8338) \\ \hline
  LSGAN   & 15.30 & \textbf{2.3088} & 26.54 & 0.8833 & 0.8361 (0.9951) & 12.8985 (0.8270) \\
  \hline
\end{tabular}
\label{tab:completion_celeb_centerSm}
\end{table}

\begin{table}[!ht]
\centering
\caption{Data: CelebA, Mask: Periocular Large}
\begin{tabular}{|c|c|c|c|c|c|c|}
  \hline
  ~       & FID     & Inception & PSNR & SSIM & OpenFace (AUC) & PittPatt (AUC)\\\hline
  Original& N/A & 2.257 & N/A & N/A         & 1.0000 (0.9993) & 19.6197 (0.9417) \\ \hline
  Stage-1 & 60.96 & 1.732 & 15.02 & 0.4978 & 0.6793 (0.9643) & 12.8902 (0.7345) \\
  Stage-2 & 48.19 & \textbf{1.839} & 15.86 & 0.5550 & \textbf{0.7080 (0.9812)} & 13.4623 (0.8609) \\
  Stage-3 & 65.83 & 1.818 & 15.80 & \textbf{0.5629} & 0.6892 (0.9788) & \textbf{13.5328 (0.8778)} \\ \hline
  WGAN    & \textbf{25.98} & 1.756 & \textbf{16.50} & 0.5418 & 0.6591 (0.9510) & 12.7374 (0.8543) \\ \hline
  LSGAN   & 40.57 & 1.829 & 15.86 & 0.5216 & 0.6882 (0.9663) & 12.8952 (0.7441) \\
  \hline
\end{tabular}
\label{tab:completion_celeb_perioLg}
\end{table}
\begin{table}[!ht]
\centering
\caption{Data: CelebA, Mask: Periocular Small}
\begin{tabular}{|c|c|c|c|c|c|c|}
  \hline
  ~       & FID     & Inception  & PSNR & SSIM & OpenFace (AUC) & PittPatt (AUC)\\\hline
  Original& N/A & 2.2517 & N/A & N/A         & 1.0000 (0.9992) & 19.4195 (0.9273) \\ \hline
  Stage-1 & 67.95 & 1.6861 & 13.97 & 0.4096 & 0.5810 (0.9506) & 10.3975 (0.7319) \\
  Stage-2 & 50.95 & 1.6034 & 14.47 & 0.4794 & \textbf{0.6283 (0.9659)} & 10.7539 (0.8314) \\
  Stage-3 & 70.68 & 1.6394 & 14.54 & \textbf{0.4884} & 0.5928 (0.9507) & \textbf{11.1033 (0.7982)} \\ \hline
  WGAN    & \textbf{27.15} & \textbf{1.8060} & \textbf{14.69} & 0.4649 & 0.5742 (0.9404) & 10.0387 (0.7990) \\
  \hline
  LSGAN   & 42.59 & 1.7341 & 14.61 & 0.4396 & 0.5861 (0.9501) &	10.4759 (0.7453)  \\
  \hline
\end{tabular}
\label{tab:completion_celeb_perioSm}
\end{table}


\begin{table}[!ht]
\centering
\caption{Data: SSFF, Mask: Center Small}
\begin{tabular}{|c|c|c|c|c|c|c|}
  \hline
  ~       & FID     & Inception  & PSNR & SSIM & OpenFace (AUC) & PittPatt (AUC)\\\hline
  Original& N/A & 1.845 & N/A & N/A        & 1.0000 (0.9987) & 19.1864 (0.9675) \\ \hline
  Stage-1 & 44.57 & 1.766 & \textbf{30.47} & \textbf{0.9087} & 0.7447 (0.9868) & 14.4907 (0.9859) \\
  Stage-2 & 45.34 & 1.766 & 29.87 & 0.9061 & 0.7503 (0.9819) & \textbf{14.5053 (0.9929)} \\
  Stage-3 & \textbf{43.57} & \textbf{1.773} & 30.07 & 0.9070 & \textbf{0.7559 (0.9820)} & 14.0614 (0.9924) \\ \hline
\end{tabular}
\label{tab:completion_ssff_centerSm}
\end{table}
\begin{table}[!ht]
\centering
\caption{Data: SSFF, Mask: Center Large}
\begin{tabular}{|c|c|c|c|c|c|c|}
  \hline
  ~       & FID     & Inception  & PSNR & SSIM & OpenFace (AUC) & PittPatt (AUC)\\\hline
  Original& N/A & 1.803 & N/A & N/A         & 1.0000 (0.9973) & 19.5385 (0.9802) \\ \hline
  Stage-1 & \textbf{81.01} & \textbf{1.883} & \textbf{26.29} & \textbf{0.7906} & 0.5307 (0.9104) & 8.1911 (0.9594) \\
  Stage-2 & 82.02 & 1.840 & 26.26 & 0.7877 & 0.5360 (0.9214) & \textbf{8.2555 (0.9611)} \\
  Stage-3 & 92.75 & 1.853 & 26.16 & 0.7835 & \textbf{0.5394 (0.9115)} & 8.2473 (0.9676) \\ \hline
\end{tabular}
\label{tab:completion_ssff_centerLg}
\end{table}
\begin{table}[!ht]
\centering
\caption{Data: SSFF, Mask: Periocular Large} 
\begin{tabular}{|c|c|c|c|c|c|c|}
  \hline
  ~       & FID     & Inception  & PSNR & SSIM & OpenFace (AUC) & PittPatt (AUC)\\\hline
  Original& N/A & 1.803 & N/A & N/A         & 1.0000 (0.9972) & 19.8658 (0.9866) \\ \hline
  Stage-1 & \textbf{88.96} & 1.717 & \textbf{18.85} & \textbf{0.6041} & 0.6561 (0.9571) & 14.8129 (0.9860) \\
  Stage-2 & 89.21 & \textbf{1.721} & 18.51 & 0.5988 & 0.6415 (0.9419) & 14.7613 (0.9694) \\
  Stage-3 & 105.85 & 1.660 & 18.74 & 0.5861 & \textbf{0.6568 (0.9589)} & \textbf{14.9813 (0.9898)} \\ \hline
\end{tabular}
\label{tab:completion_ssff_perioLg}
\end{table}
\begin{table}[!ht]
\centering
\caption{Data: SSFF, Mask: Periocular Small}
\begin{tabular}{|c|c|c|c|c|c|c|}
  \hline
  ~       & FID     & Inception  & PSNR & SSIM & OpenFace (AUC) & PittPatt (AUC)\\\hline
  Original& N/A & 1.803 & N/A &  N/A        & 1.0000 (0.9991) & 19.1976 (0.9679) \\ \hline
  Stage-1 & \textbf{102.30} & 1.634 & 17.15 & \textbf{0.5218} & \textbf{0.5473 (0.9094)} & 11.2008 (0.9673) \\
  Stage-2 & 104.59 & \textbf{1.665} & 16.84 & 0.5177 & 0.5139 (0.8944) & 11.2372 (0.9808) \\
  Stage-3 & 118.66 & 1.636 & \textbf{17.24} & 0.5012 & 0.5412 (0.8965) & \textbf{11.4808 (0.9800)} \\ \hline
\end{tabular}
\label{tab:completion_ssff_perioSm}
\end{table}
From those figures, it can be seen that the most difficult cases of image completion occur when the faces are not front-facing. This can be related to the relative imbalance in the dataset between front-facing and non front-facing images, specifically the lack of the latter in the dataset. We generally observe better metrics in terms of Inception Score, PSNR, OpenFace NCS and PittPatt Score as we move up the stages. This shows that the higher stages are not just maintaining identity and visual appeal but also adding sharpness and details to the generated images.


\section{Sample Aware vs. Sample Agnostic RankGAN}\label{sec:supp-3}
\begin{figure*}[!ht]
  \centering
  \includegraphics[width=\linewidth]{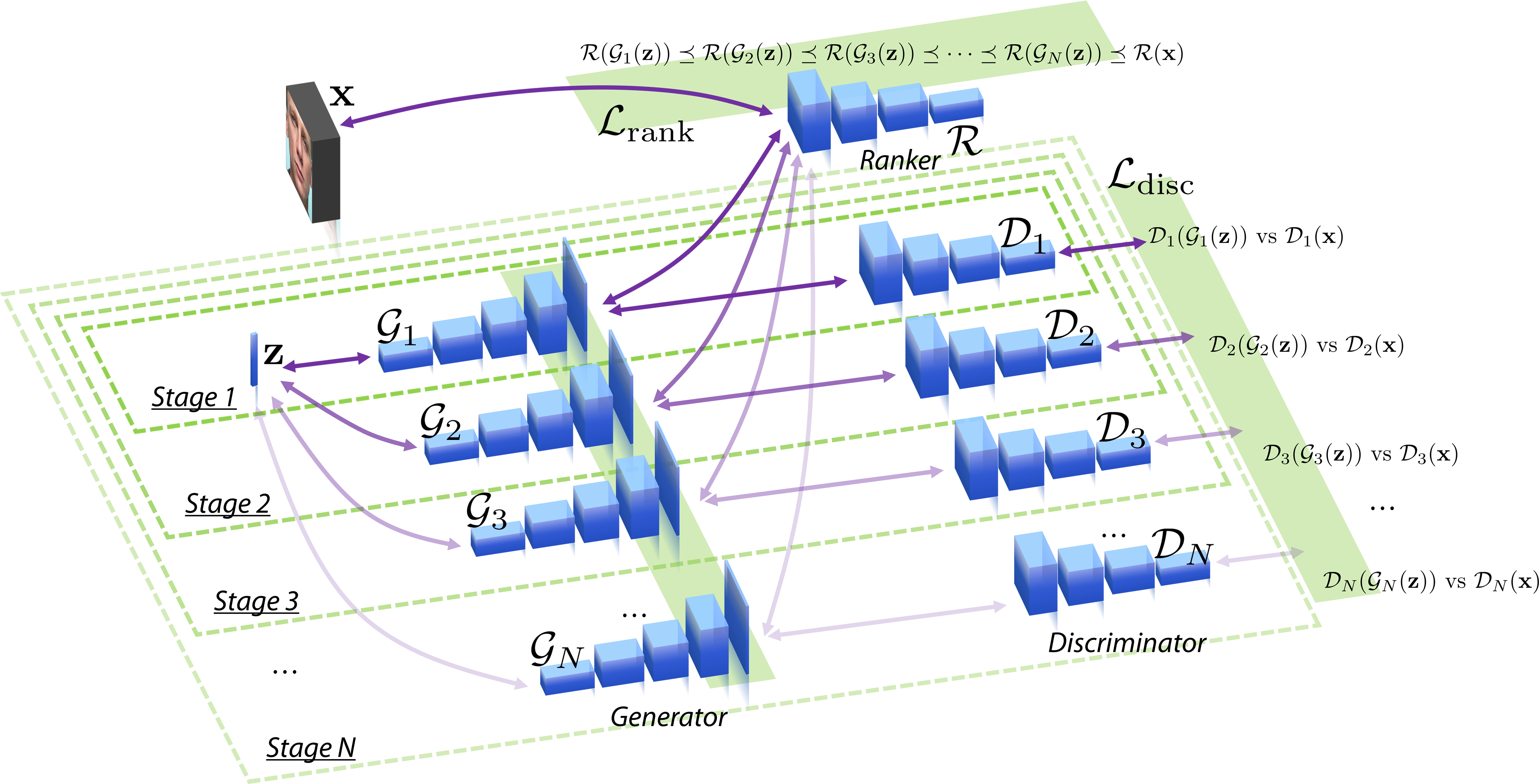}
  \caption{Flowchart of the proposed RankGAN method without the Encoder $\mathcal{E}$.}
  \label{fig:arxiv-stage}
\end{figure*}

In this section, we discuss further on the design principles regarding the Encoder $\mathcal{E}$. The RankGAN framework can be trained both with and without the encoder. We call the former method sample aware training and the latter, sample agnostic. In the case of the latter, as the stage progresses, the discriminator ranks generated samples from various stages while being sample agnostic, \ie, the samples being ranked are arbitrary. This is still a viable solution for the ranker to provide meaningful feedback for the entire network. However, such sample agnostic ranking is `fuzzy' and has been empirically observed to result in slower convergence of the ranking loss. This has been explored in \cite{felix_arxiv17_gogan}. The training pipeline for sample agnostic RankGAN is shown in Figure~\ref{fig:arxiv-stage}.

\section{Connection Between Hinge Loss and $f$-Divergence}\label{sec:supp-4}

A connection can be established between the hinge loss we used for RankGAN, and $f$-divergence. In particular, minimizing the hinge loss is closely related to minimizing one particular instance of the $f$-divergence, which is the variational distance. The goal of the discussion is to establish a correspondence between the family of surrogate loss (convex upper bound on the 0-1 loss, such as hinge loss) functions and the family of $f$-divergences. It can be shown that any surrogate loss induces a corresponding $f$-divergence, and any $f$-divergence satisfying certain conditions corresponds to a family of surrogate loss functions. Readers are encouraged to refer to the work of Ngugen \etal \cite{hinge-div1,hinge-div2,hinge-div3,hinge-div4} for such a connection in detail.

\begin{figure*}
    \centering
    \subfigure[Original faces.]{\includegraphics[width=0.432\linewidth]{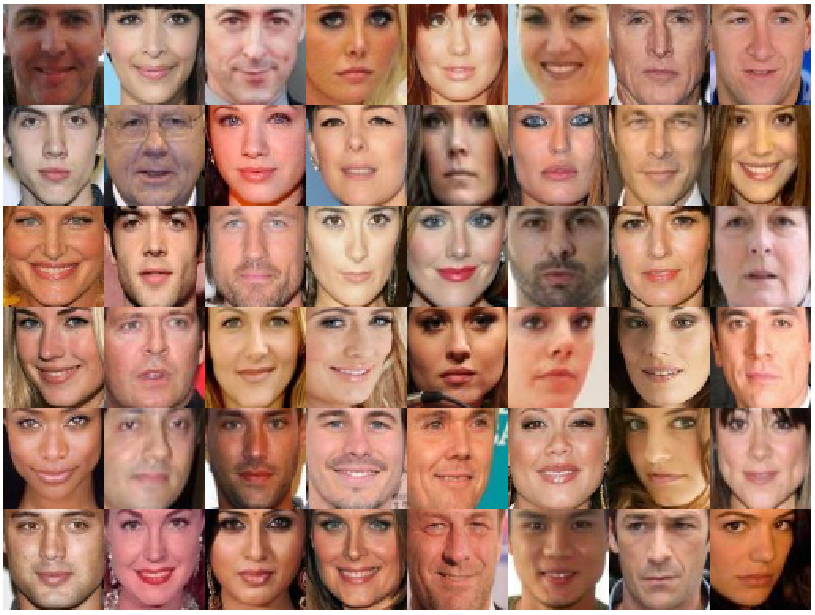}}
    \subfigure[Masked faces. (`Center Large')]{\includegraphics[width=0.432\linewidth]{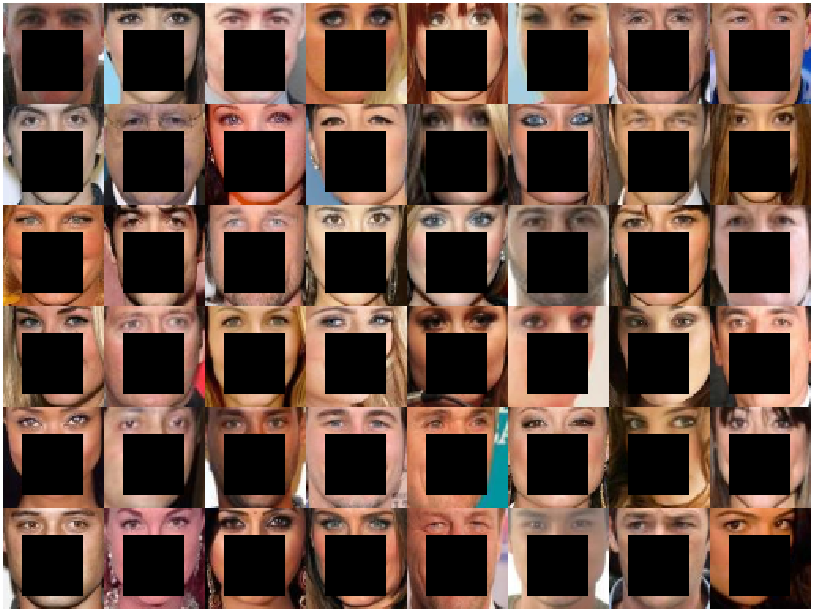}}
    \subfigure[WGAN completion, Open Set.]{\includegraphics[width=0.432\linewidth]{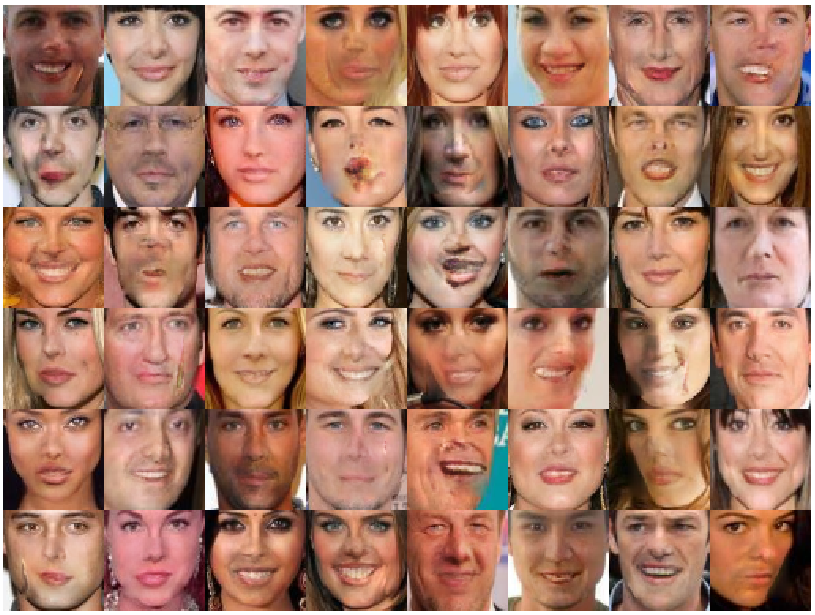}}
    \subfigure[LSGAN completion, Open Set.]{\includegraphics[width=0.432\linewidth]{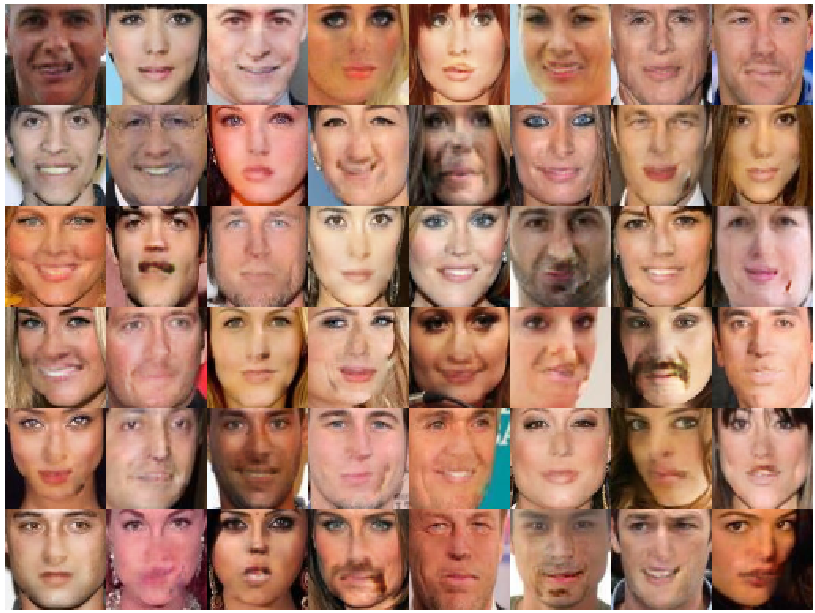}}
    \subfigure[Stage 1 completion, Open Set.]{\includegraphics[width=0.432\linewidth]{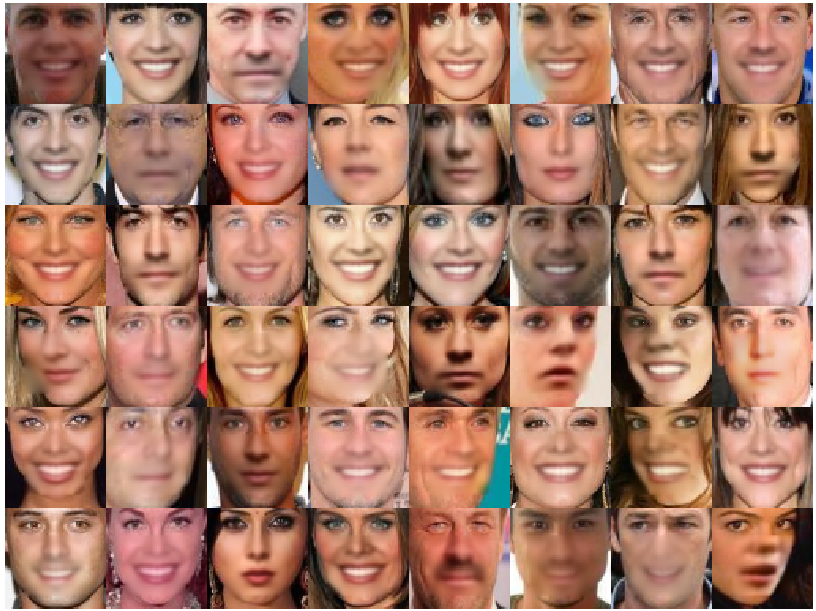}}
    \subfigure[Stage 2 completion, Open Set.]{\includegraphics[width=0.432\linewidth]{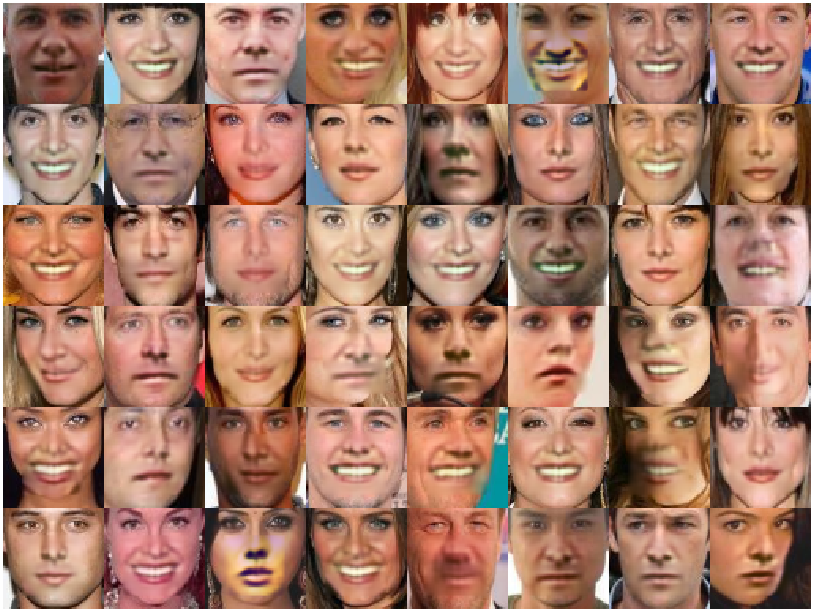}}
    \subfigure[Stage 3 completion, Open Set.]{\includegraphics[width=0.432\linewidth]{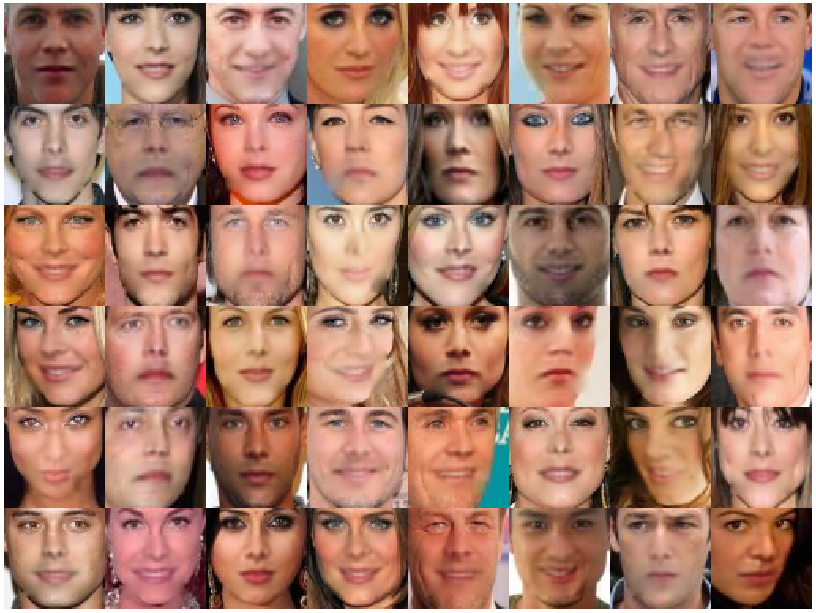}}
\caption{\textcolor{blue}{Best} completion results with RankGAN on CelebA, `Center Large' mask.}\label{fig:sup_celeba_good_center_lg}
\end{figure*}
\begin{figure*}
    \centering
    \subfigure[Original faces.]{\includegraphics[width=0.432\linewidth]{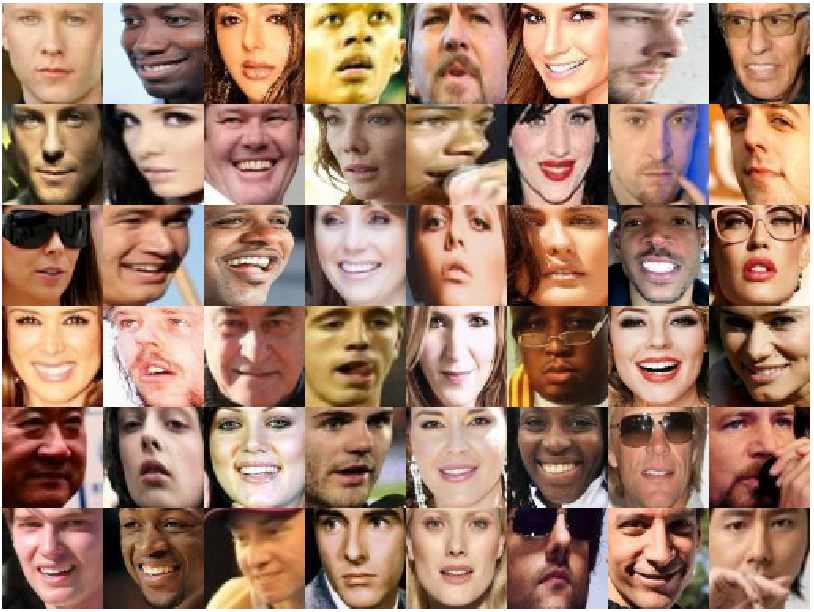}}
    \subfigure[Masked faces. (`Center Large')]{\includegraphics[width=0.432\linewidth]{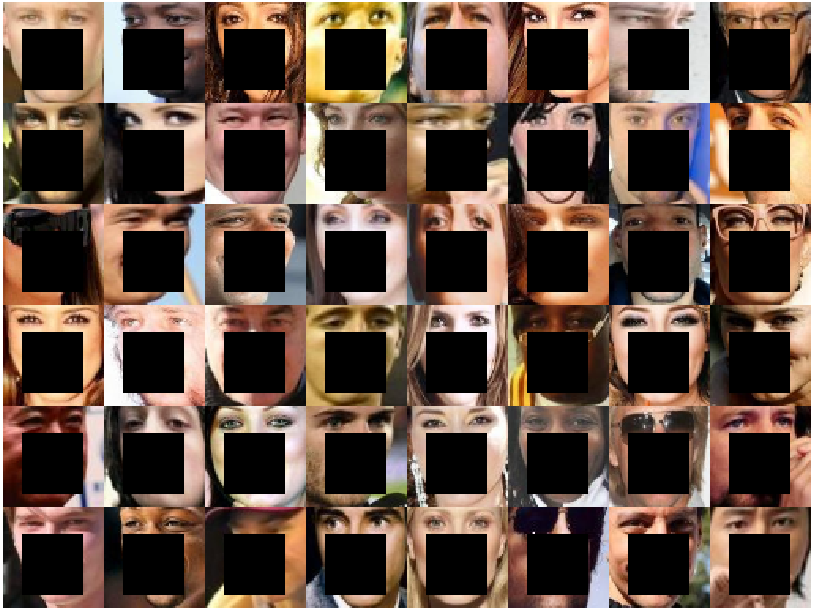}}
    \subfigure[WGAN completion, Open Set.]{\includegraphics[width=0.432\linewidth]{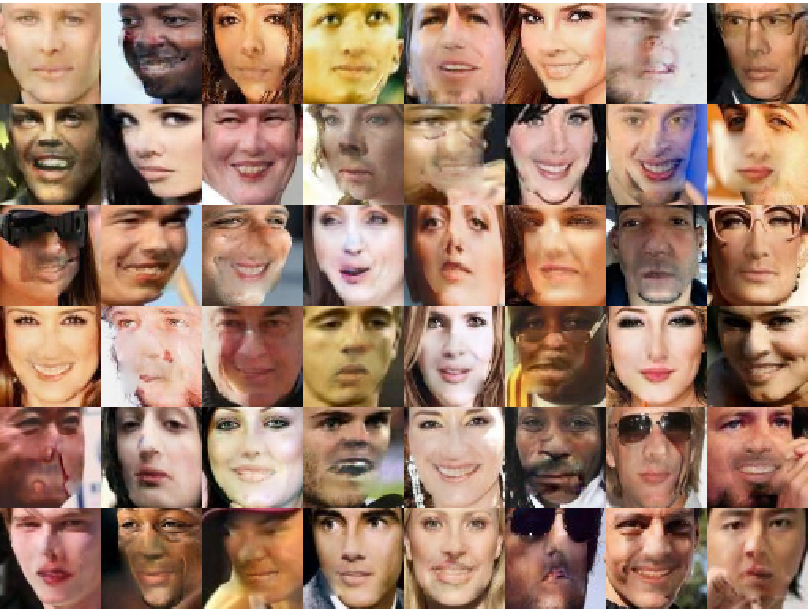}}
    \subfigure[LSGAN completion, Open Set.]{\includegraphics[width=0.432\linewidth]{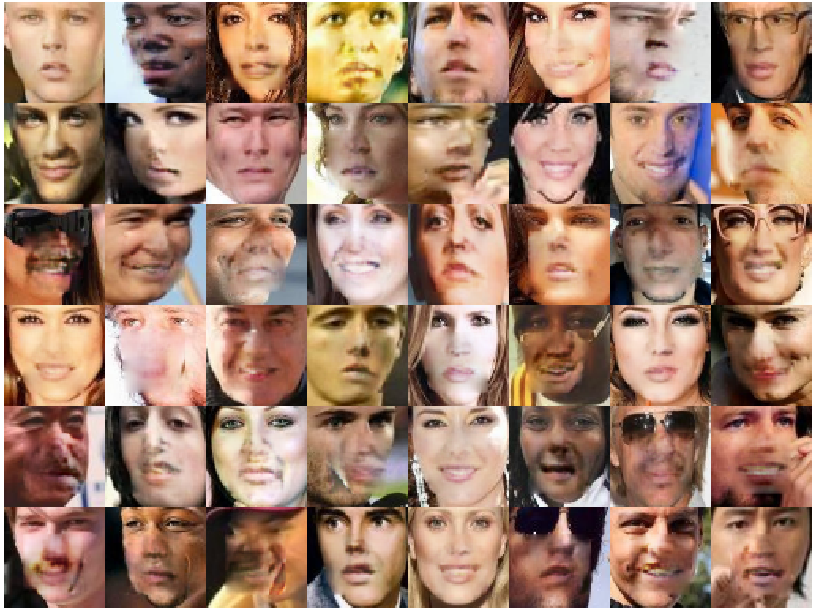}}
    \subfigure[Stage 1 completion, Open Set.]{\includegraphics[width=0.432\linewidth]{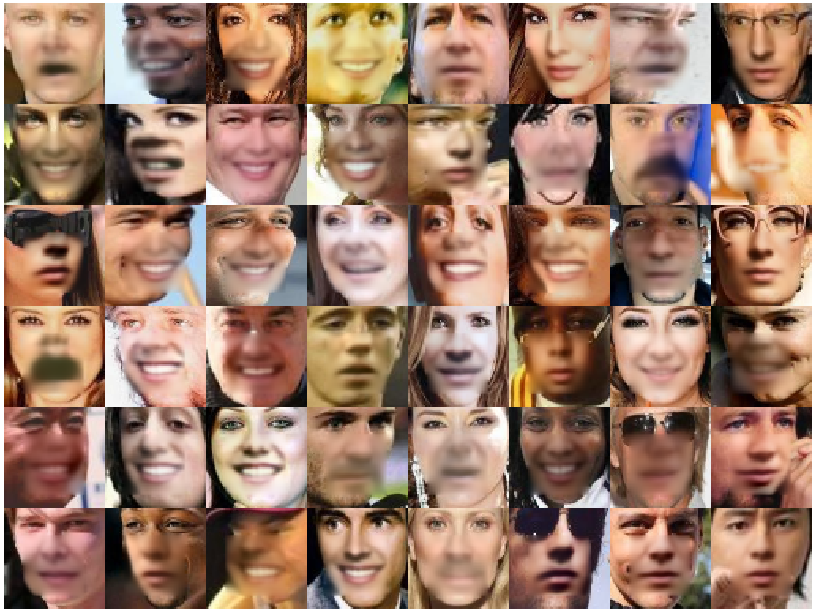}}
    \subfigure[Stage 2 completion, Open Set.]{\includegraphics[width=0.432\linewidth]{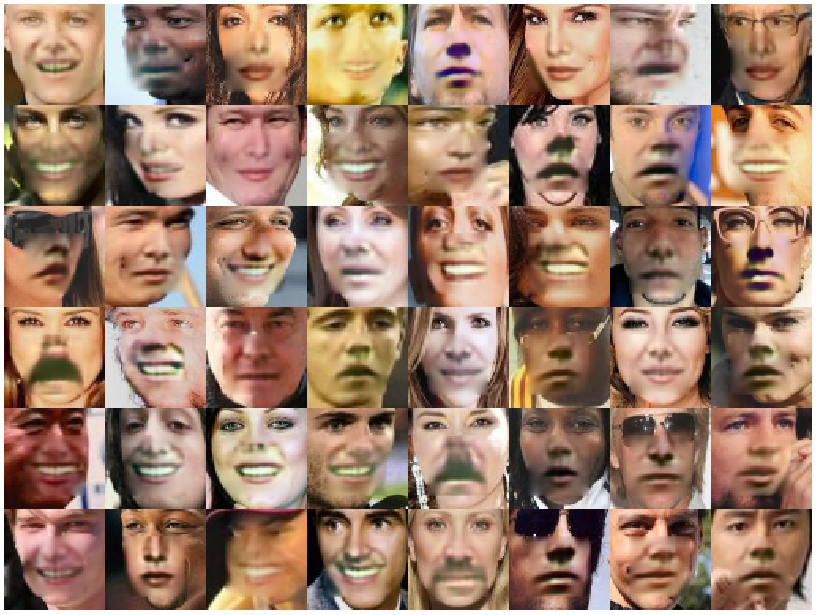}}
    \subfigure[Stage 3 completion, Open Set.]{\includegraphics[width=0.432\linewidth]{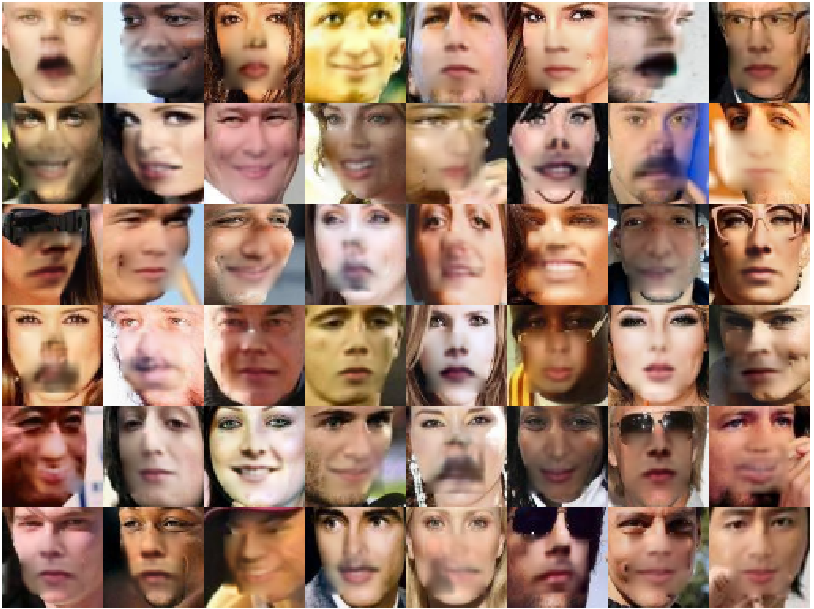}}
\caption{\textcolor{red}{Worst} completion results with RankGAN on CelebA, `Center Large' mask.}\label{fig:sup_celeba_bad_center_lg}
\end{figure*}



%
%
%
\clearpage
\bibliographystyle{splncs04}
\bibliography{ref_rankgan}

\end{document}